\documentclass[10pt,twocolumn,letterpaper]{article}

\usepackage[pagenumbers]{cvpr} 

\usepackage[dvipsnames]{xcolor}

\definecolor{cvprblue}{rgb}{0.21,0.49,0.74}
\usepackage[pagebackref,breaklinks,colorlinks,citecolor=cvprblue]{hyperref}

\def\paperID{12930} 
\def\confName{CVPR}
\def\confYear{2024}

\usepackage{multirow}
\usepackage{subcaption}
\title{ Continuous Pose for Monocular Cameras in Neural Implicit Representation}

\author{ Qi Ma\textsuperscript{1,2}\space\space\space\space Danda Pani Paudel\textsuperscript{1,2}\space\space\space\space Ajad Chhatkuli\textsuperscript{1}\space\space\space\space Luc Van Gool\textsuperscript{1,2}\\
\textsuperscript{1}Computer Vision Lab, ETH Zurich\space\space\space\space \textsuperscript{2}INSAIT, Sofia University  }

\begin{document}
\maketitle
\begin{abstract}

In this paper, we showcase the effectiveness of optimizing monocular camera poses as a continuous function of time. The camera poses are represented using an implicit neural function which maps the given time to the corresponding camera pose. The mapped camera poses are then used for the downstream tasks where joint camera pose optimization is also required. While doing so, the network parameters -- that implicitly represent camera poses --  are optimized. We exploit the proposed method in four diverse experimental settings, namely,
(1) NeRF from noisy poses; (2) NeRF from asynchronous Events; (3) Visual Simultaneous Localization and Mapping (vSLAM); and (4) vSLAM with IMUs. In all four settings, the proposed method performs significantly better than the compared baselines and the state-of-the-art methods. Additionally, using the assumption of continuous motion, changes in pose may actually live in a manifold that has lower than 6 degrees of freedom (DOF) is also realized. We call this low DOF motion representation as the \emph{intrinsic motion} and use the approach in vSLAM settings, showing impressive camera tracking performance.
We release our code at  \href{https://github.com/qimaqi/Continuous-Pose-in-NeRF.git}{https://github.com/qimaqi/Continuous-Pose-in-NeRF} .

\end{abstract}    
\section{Introduction}
\label{sec:intro}

The concept of motion, the change of position and orientation of an object in its surroundings, is fundamentally continuous in nature. This continuity is evident in the ways we achieve, perceive and measure motion, with velocity and acceleration being the most common measures for both linear and angular motion. This idea of continuity is also true for the 3D poses of navigating cameras. Often the camera motion needs to be estimated from its measurements -- also known as the camera localization problem. In most common settings, the inputs are RGB-only frames, depth frames, asynchronous event streams, or a combination thereof.  In some cases, these measurements are augmented by Inertial Measurement Unit (IMU) outputs, which measure a change in pose directly. In all those settings, the camera motion is estimated via some optimization technique that searches $SE(3)$ pose parameters. While doing so, existing techniques choose to optimize a discrete set of $SE(3)$ parameters, ignoring the inter-frame continuity of camera poses. This choice can be primarily attributed to the otherwise difficulty in optimization.


\begin{figure*}
\centering
\includegraphics[width=0.9\linewidth]{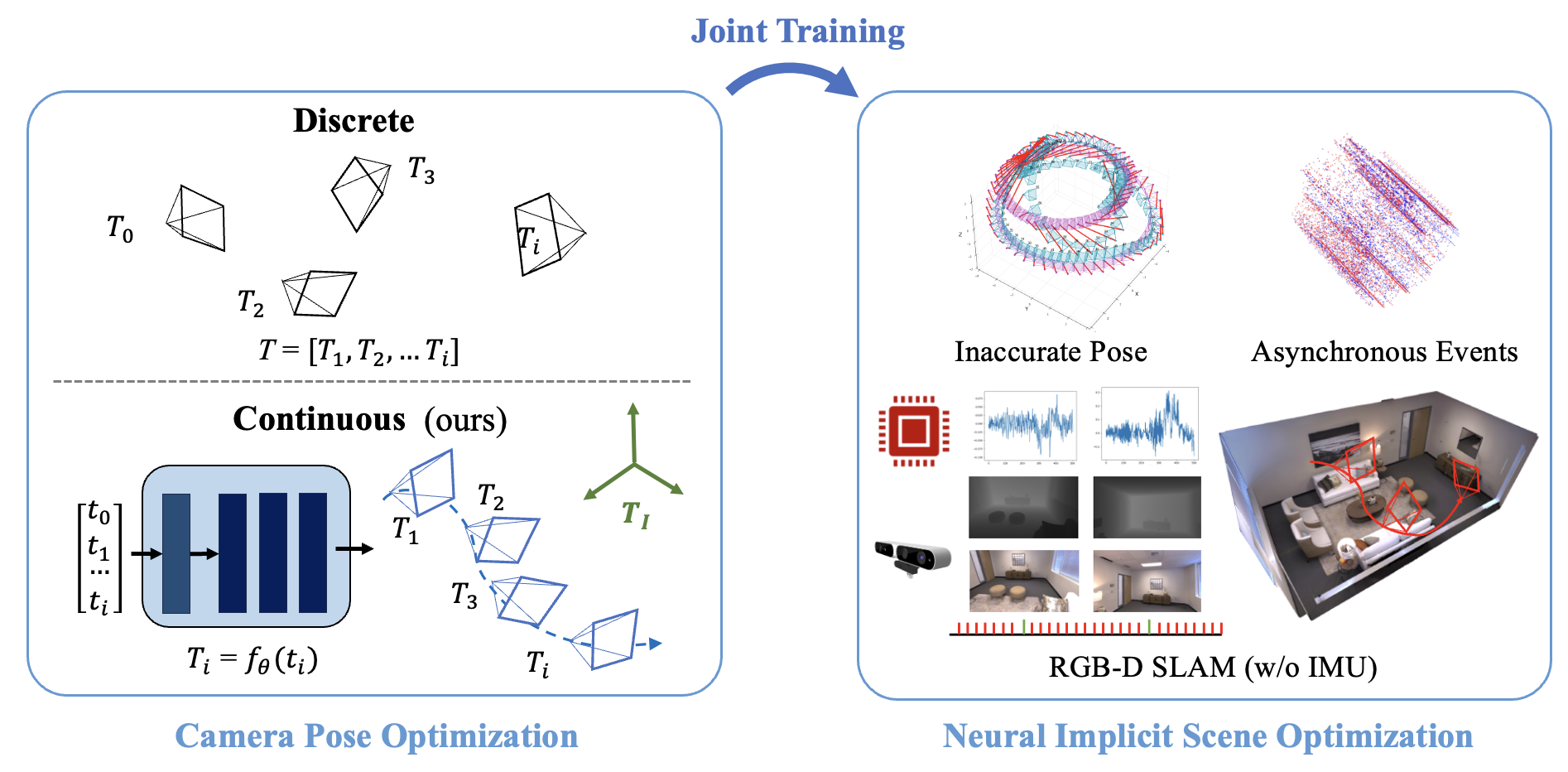}

\caption{We showcase the benefits of optimizing the poses as a continuous function of time in diverse settings. We conduct exhaustive experiments on (a) rectifying inaccurate poses in RGB-only settings; (b) utilizing the asynchronous stream of events, (c) performing vSLAM in RGB-D camera settings;  (d) integrating high-frequency IMUs in vSLAM. All experiments use neural functions for both camera poses and scene representations. Additionaly we exploit low dof motion representation in intrinsic motion frame $T_I$. \label{fig:overview}}
\end{figure*}

While handling high-frequency IMUs or asynchronous events in common practice, pose optimization at every measurement time is avoided, due to its computational cost. Instead, the measurements between two arbitrarily chosen keyframes are accumulated before utilizing them. Then the poses are optimized only for the chosen, sparser keyframes. We argue that this raises three major concerns: (i) inaccurate accumulation of intermediate measurements; (ii) loss of fine-grained motion details; (iii) lack of the continuous motion prior. 

In order to address these concerns, we represent and optimize the pose of a moving camera as a continuous function of time. Unlike classical state estimation method ~\cite{barfoot2014batch, furgale2012continuous, 8432102} which models continuous pose with Gaussian Process or B-spline, our neural pose function can be easily optimized jointly with other task-specific implicit neural representation (INR)~\cite{mildenhall2021nerf, park2019deepsdf, mescheder2019occupancy}. More precisely, for translation $\mathsf{v}\in\mathbb{R}^3$ and rotation $\mathsf{R}(\mathsf{q})\in SO(3)$ parameterized by quaternions $\mathsf{q}\in\mathbb{R}^4$ with $||\mathsf{q}||=1$, the continuous pose of the monocular camera is given by,
\begin{equation}
[\mathsf{q};
\mathsf{v}]
= f_\theta (t),
\end{equation}
where $f_\theta(.)$ is the neural function parameterized by  $\theta$. This is a continuous function that maps the time $t\in\mathbb{R}$ to the pose in $SE(3)$.   
While being simple, this representation has numerous benefits including ease of optimization and its cosmopolitan applicability. Some example applications are illustrated in Figure~\ref{fig:overview}. In the following, we further discuss how our simple approach addresses the previously raised concerns.

\paragraph{No error due to measurement accumulation:} High frequency or asynchronous measurements can be utilized directly without accumulation, integral, or rounding. We infer the camera pose precisely at the measurement time. For example, in the case of an event camera, each asynchronous event's pose is inferred precisely at the event time. Similarly, in the case of IMUs, no motion integration before supervision is required. These abilities protect our approach against error injection due to any form of accumulation.


\paragraph{Fine-grained motion details:} By virtue of the continuous representation, temporally fine-grained details of the pose can be captured. This is particularly interesting with high-frequency IMUs or asynchronous event cameras. Our approach allows for the recovery of the pose at the very moment of measurement, which otherwise often is an ill-posed problem and could only be interpolated with an assumed smoothness and order.

\paragraph{Continuous motion prior:} The inductive bias of continuous monocular camera motion is meaningfully injected by the proposed method. This resulted in very encouraging results in our experiments. In particular, while denoising the inaccurate camera poses and during the vSLAM experiments, the benefits were evident under the standard settings of BARF~\cite{lin2021barf} and NICE-SLAM~\cite{zhu2022nice}, respectively. It is important to note that our representation offers first- and second-order derivatives via auto-differentiation of the neural network. Consequently, quantities such as velocity and acceleration do not require additional care. Thus the fusion of IMU measurements is natural and straightforward. 

In addition to the above, we further show the utility of the neural pose in order to optimize the continuous pose by decomposing each change in pose into a slowly changing reference and a low DOF motion. We define this as the \emph{intrinsic motion}.
In our experiments we observed that our continuous pose representation improves the tracking performance significantly in the vSLAM tasks. This can be primarily attributed to the reasons mentioned above, which serve to facilitate the optimization process. Inspired by the fact that actual motion always possesses a lower degree of freedom, we define the intrinsic motion frame as a coordinate system that can express the camera motion with the lowest dimensional manifold. For example: Rotational motion around a fixed axis can be expressed in the coordinate system aligned with the rotational axis with only one degree of freedom. A natural observation is that the relative motion with respect to intrinsic motion is usually sparse, moreover, the continuous motion tends to share the same intrinsic motion frame which can be well modeled as a continuous function of time. By exploiting it we decompose the camera relative motion with a low-dimensional intrinsic motion $ [\mathsf{R}_I,
\mathsf{v}_I]$ and the rigid transformation from camera frame to the intrinsic motion frame $ [\mathsf{R}_o,
\mathsf{v}_o]$ as follows:
\begin{equation}
[\mathsf{R},
\mathsf{v}]
= [\mathsf{R}_o,
\mathsf{v}_o] [\mathsf{R}_I,
\mathsf{v}_I],
\end{equation}


Our major contributions can be summarized as follows:
\begin{itemize}
    \item We propose a simple yet effective way to represent the monocular camera motion via a neural function of time that can be optimized efficiently together with implicit neural representations.
    \item We demonstrate the utility of the proposed representation in four diverse applications with different camera setups, including IMUs and moving event cameras. 
    \item Through exhaustive experiments, we demonstrate clear benefits of the proposed representation over the existing alternatives and classical method. These benefits include ease of optimization, widespread use for different camera and sensor types, and notable performance gain with no additional effort. 
    \item We further improve the full 6-DOF pose of monocular camera by exploiting the sparsity of the intrinsic motion, which fits neatly into the proposed framework of continuous neural pose. The final pose thus obtained shows remarkable improvement over the conventional baselines.
\end{itemize}

\section{Related work}
\label{sec:related}


\subsection{Camera Poses in NERF}
NERF~\cite{mildenhall2021nerf} consists of joint optimization of the surface density and the rendered color given the images with known camera rays. Consequently NERF models are highly sensitive to camera pose errors~\cite{truong2022sparf,lin2021barf,nerfmm, GARF, GNerF, sinerf}. Recently several works have tackled the pose error by jointly optimizing poses with the radiance field. \cite{lin2021barf, chen2023local, jeong2021selfcalibrating,chen2023dbarf} optimizes camera poses in bundle adjustment fashion in order to solve the same issue. While these methods use the smooth pose prior, the poses are still optimized as discrete variables. On the other spectrum \cite{truong2022sparf} optimizes noisy poses for sparse camera views with the radiance fields opting for a different class of applications. \cite{bian2023nope} enforces the inter-frame consistency by incorporating monocular depth prior.

\subsection{Camera Poses with IMUs}
The inertial measurement unit (IMU) serves as a scene-independent sensor that is the ideal complement to cameras in order to achieve robustness in low texture, high speed, and HDR scenarios. Fusing visual information and IMU tightly ~\cite{scaramuzza2019visual} to estimate pose as discrete states is proposed first by MSCKF
 ~\cite{mourikis2007multi} (an extended Kalman Filter (EKF)), \cite{leutenegger2013keyframe} further improves it with keyframes and bundle adjustment. \cite{huang2018online, qin2017vins, forster2016svo, usenko2019visual} improve in robustness compared to feature matching by using the direct photometric error. \cite{campos2021orb} propose fast and accurate IMU initialization based on MAP estimation. Recent research has also focused on integrating IMU and visual priors with neural network, \eg, the camera pose is implicitly used for image deblurring  \cite{mustaniemi2019gyroscope} or video stabilization \cite{shi2022deep}. \cite{Herath_2022_CVPR} proposes neural inertial localization with IMU alone for indoor scenes.

\subsection{Camera Poses in Dense SLAM}
Visual SLAM~\cite{davison2007monoslam,klein2009parallel,richard2011dense} is a key 3D vision application where an agent camera is localized simultaneously while building the map using visual information. We again focus on methods in the context of radiance fields~\cite{zhu2022nice,zhu2023nicer,imap,neuralrgbd,rosinol2022nerf}. IMAP~\cite{imap} is a recent seminal work which works on RGBD images to optimize an implicit scene representation with a single MLP network. It optimizes the camera pose while representing them as discrete sets of parameters for the keyframes. NICE-SLAM~\cite{zhu2022nice} improves on it by using 3D voxel features along with corresponding 2D image features thus providing a better scene representation. Indeed most approaches~\cite{zhu2023nicer,sandstrom2023point,rosinol2022nerf,li2023dense} focus on improving the scene representation for better localization and mapping or with RGB-only input.

\subsection{Camera Poses in Event Cameras}
Unlike standard frame-based camera imaging, event cameras provide image signals as asynchronous events in microsecond intervals~\cite{kim2016real}. Thus, it forms the perfect use case for a continuous time representation of camera poses. Similar to NERF-less SLAM~\cite{davison2007monoslam}, this is traditionally done using variations of Kalman Filter with motion models~\cite{kim2016real,milford2015towards}. A recent work~\cite{zhou2021event} represents camera tracking as a function of time but uses a Levenberg-Marquardt optimization directly on the sets of poses without intermediate representation. Recently, there have been efforts to use event-based radiance fields in the neural network~\cite{klenk2023nerf,hwang2023ev,rudnev2022eventnerf}. However, camera pose optimization as a function of time is still not explored in the radiance field literature with events.

\subsection{Continuous Pose representation}

While discrete-time representations are commonly employed in Simultaneous Localization and Mapping (SLAM) tasks, they face challenges when integrating data from high-frequency sensors like Inertial Measurement Units (IMUs) and asynchronous events. \cite{furgale2012continuous} address this issue by proposing representing the continuous-time state using temporal basis functions such as B-spline basis.\cite{barfoot2014batch} model the continuous state using Gaussian processes, defining continuous-time priors through covariance functions.\cite{patron2015spline} leverage cumulative cubic B-splines to mitigate rolling-shutter artifacts. Notably, spline-based continuous-time trajectory representations have found application in laser-based SLAM methods \cite{kaul2016continuous, nuchter2017improving}.

\section{Time-to-Pose Mapping Network}
\subsection{Architecture of the Proposed PoseNet}
In order to learn time-to-pose mapping, we use 8-layer MLP parameterized by $f(\theta_p)$ with ReLU activation functions and 256-dimensional hidden units, which we refer to as pose-network (PoseNet). PoseNet first embeds the time variable into high-dimensional space using sinusoidal harmonic functions~\cite{mildenhall2021nerf}. The outputs of this network are $[\mathsf{v}, \mathsf{q}]$: translation vector $\mathsf{v}\in\mathbb{R}^3$ and the rotation represented by a quaternion $\mathsf{q}\in\mathbb{R}^4$. Finally, we use the tanh activation in the last layer to map output to the range $[-1, 1]$, and normalize it as a unit quaternion. 
We study different embedding dimensions and architectures in the context of NeRF from the inaccurate pose, which is reported in Tables~\ref{tab:barf_encoder}. The best-performing embedding and architecture, in these experiments, are then used for the other applications. Additional information concerning network size and computational details is provided in the supplementary material.

\subsection{Implementation Variances across Applications}
The simplicity of PoseNet allows us to use it in diverse applications in a plug-and-play manner. In all applications that we report in the following sections, we optimize the PoseNet parameters $\theta_p$ as a surrogate of the direct pose optimization. We denote the network parameters for the INR of the scene as  $\theta_s$.
In NeRF from inaccurate poses, the objective is to minimize the radiance field loss~\cite{mildenhall2021nerf,lin2021barf} given $N$ images and corresponding timestamp $t_i$ for image $i$:
\begin{equation}
\min_{\theta_s, \theta_p} \sum_{i=1}^{N}\lVert\mathcal{I}_{i} - g \left( {\theta_s,f(\theta_p, t_i) } \right) \rVert.
\label{eq:obj_barf}
\end{equation}

$g(\theta_s, T_i)$ represents the mapping from the camera pose $T_i$ to the RGB value, including ray composition and radiance field model. In case of NeRF with asynchronous events~\cite{rudnev2022eventnerf}, $N$ refers to the number of the sampled events. Note that in both cases, we output the predicted transformation and compose it with the initial pose: $T_i = T_{init_i} \circ T_{refine_i}$. The refined transformation is obtained as  $T_{refine_i} = P(f(\theta_p, t_i) )$, $P(.)$ being the vector to rigid transformation conversion operator.

In the task of Dense-SLAM tracking, for each tracking iteration we optimize PoseNet with the following objective:
\begin{equation}
\begin{split}
& \min_{\theta_p} \sum_{i=1}^{M}(\mathcal{L}_{g} \left(D_i,P(f(\theta_p, t_i)\right)) + \lambda_{p}\mathcal{L}_{p} \left(I_i,P(f(\theta_p, t_i)\right)) )
\end{split}
\label{eq:obj_slam}
\end{equation}
We use the same geometric loss $\mathcal{L}_{g}$ and photometric loss $\mathcal{L}_{p}$ as in NICE-SLAM~\cite{zhu2022nice}. $D_i, I_i$ represent depth and RGB measurements for $M$ sampled pixels respectively, obtained via volume rendering.

\subsection{Intrinsic Motion Frame}
Within the neural dense SLAM application, we additionally introduce intrinsic motion frame in order to improve tracking within a low-dimensional manifold. This is accomplished through motion decomposition and enforcing minimal DOF. More specifically, we use two PoseNet $f_{o}(\theta_{p_{o}}),\ f_{I}(\theta_{p_{I}})$ in order to model the intrinsic motion $T_o = [\mathsf{R}_o,\mathsf{v}_o],\ T_I = [\mathsf{R}_I,\mathsf{v}_I]$, such that $T = T_o \circ T_I$. Here, $T_o$ is the transformation to the \emph{intrinsic frame} or in short, intrinsic transform. $T_I$ then denotes the intrinsic motion. Therefore we can rewrite Eq \eqref{eq:obj_slam} as:
\vspace{-1ex}
\begin{equation}
\begin{split}
& \min_{\theta_{p_o},\theta_{p_I}} \sum_{i=1}^{M}(\mathcal{L}_{g}(D_j,f_o(\theta_{p_o}, t_i) \circ f_I(\theta_{p_I}, t_i)) \\
& + \mathcal{L}_{p} \left(I_j, f_o(\theta_{p_o}, t_i) \circ f_I(\theta_{p_I}, t_i)\right) \\ 
& + \mathcal{L}_{dof}(f_I(\theta_{p_I}, t_i)) +  \mathcal{L}_{o}(f_o(\theta_{p_o}, t_i))
\end{split}
\label{eq:intr}
\end{equation}
Note that the operator $P$ should be included for absolute correctness in the function compositions in Eq~\eqref{eq:intr}. The DOF loss $\mathcal{L}_{dof}$ is computed as follows:
\begin{itemize}
\item \textbf{Step1}: Obtain $ [\mathsf{R}_I,\mathsf{v}_I]$ from intrinsic motion PoseNet $ f_{I}$
\item \textbf{Step2}: Convert rotation matrix $\mathsf{R}_I$ to Euler angles  $\alpha_I \in\mathbb{R}^3$, normalize with angle of view $\gamma$, $\hat{\alpha_I} = 2\alpha_I / \gamma $
\item \textbf{Step3}: Normalize translation vector with $\hat{v_I} = v_I/ \lVert v_I \rVert$.
\item \textbf{Step4}: DOF Loss $\mathcal{L}_{dof}$ =  $\lVert [\hat{\alpha_I}, \hat{v_I}] \rVert_{0}$.
\label{eq:dof_loss}
\end{itemize}

We relax the $\ell_0$ norm to $\ell_1$ norm for optimization. In steps 2 and 3, normalization also serves to balance translation and rotation components during optimization. We employ view angle normalization with the assumption that the angle between two relative views in vSLAM tasks is always smaller than half of the viewing angle. To handle the cases where unconstrained intrinsic motion tends move to infinity in cases of small rotation, we introduce an additional $\mathcal{L}1$ regularization term for the translation $\mathcal{L}_{o} = |v_o|$.

\begin{figure*}
\begin{minipage}[h]{0.37\textwidth}
\rotatebox{90}{(a)}
\begin{subfigure}{\textwidth}
\centering
\includegraphics[width=0.8\linewidth]{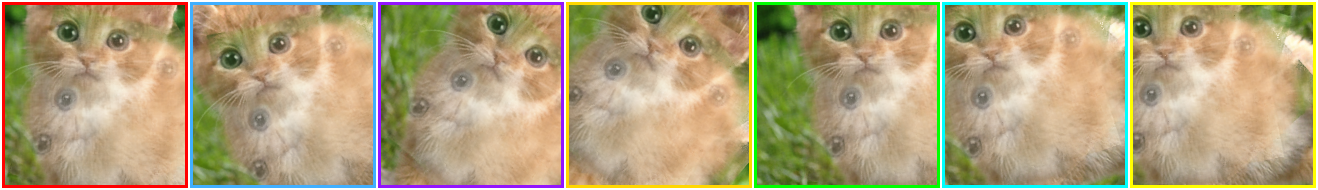}
\end{subfigure}
\rotatebox{90}{(b)}
\begin{subfigure}{\textwidth}
\centering
\includegraphics[width=0.8\linewidth]{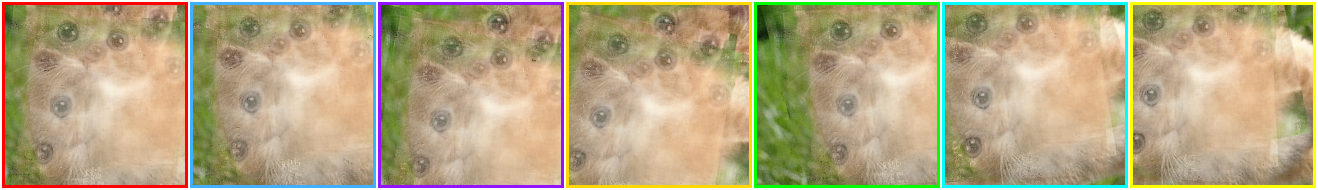}
\end{subfigure}
\rotatebox{90}{(c)}
\begin{subfigure}{\textwidth}
\centering
\includegraphics[width=0.8\linewidth]{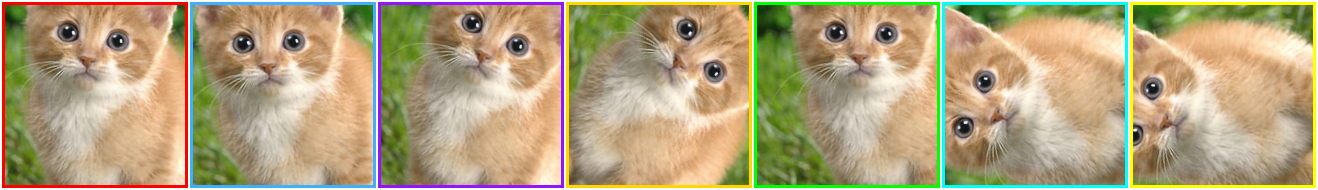}
\end{subfigure}
\rotatebox{90}{(d)}
\begin{subfigure}{\textwidth}
\centering
\includegraphics[width=0.8\linewidth]{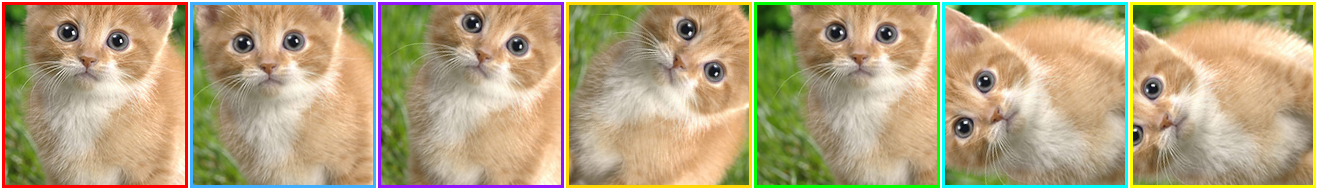}
\end{subfigure}

\vspace{0.1cm}

\rotatebox{90}{(a)}
\begin{subfigure}{\textwidth}
\centering
\includegraphics[width=0.8\linewidth]{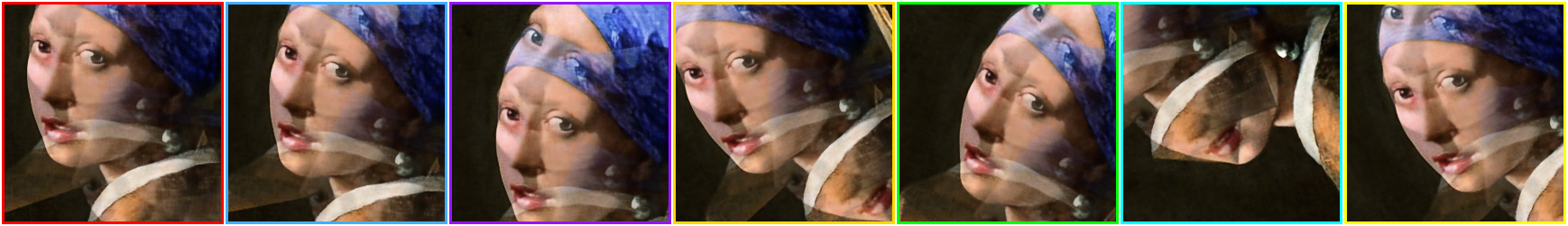}
\end{subfigure}
\rotatebox{90}{(b)}
\begin{subfigure}{\textwidth}
\centering
\includegraphics[width=0.8\linewidth]{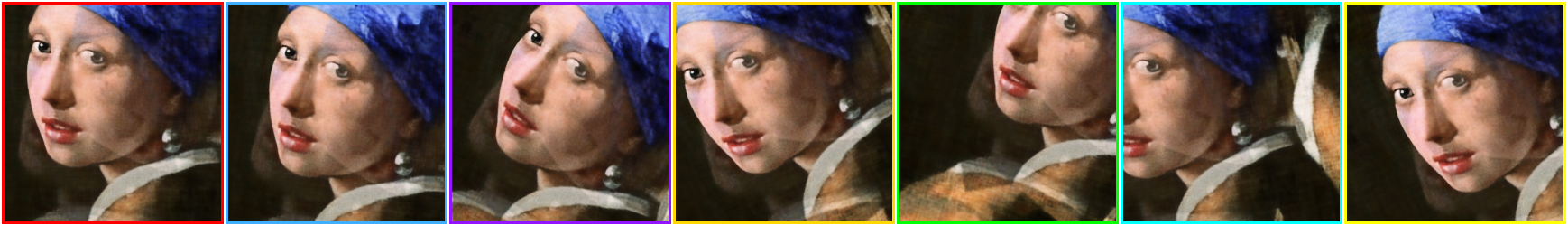}
\end{subfigure}
\rotatebox{90}{(c)}
\begin{subfigure}{\textwidth}
\centering
\includegraphics[width=0.8\linewidth]{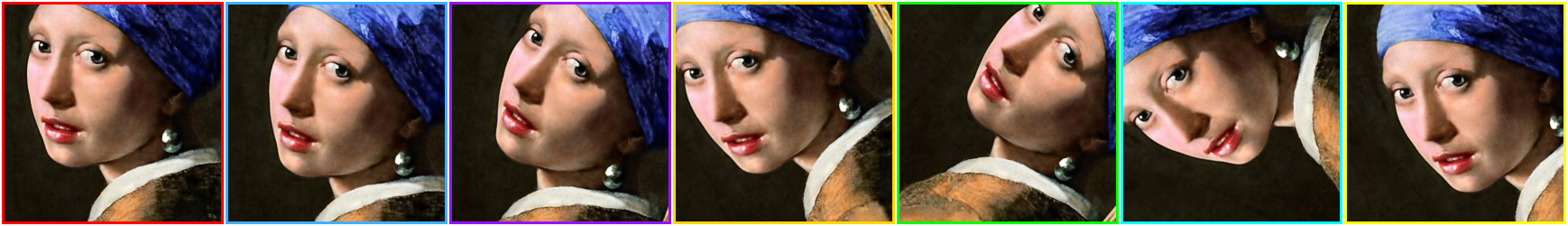}
\end{subfigure}
\rotatebox{90}{(d)}
\begin{subfigure}{\textwidth}
\centering
\includegraphics[width=0.8\linewidth]{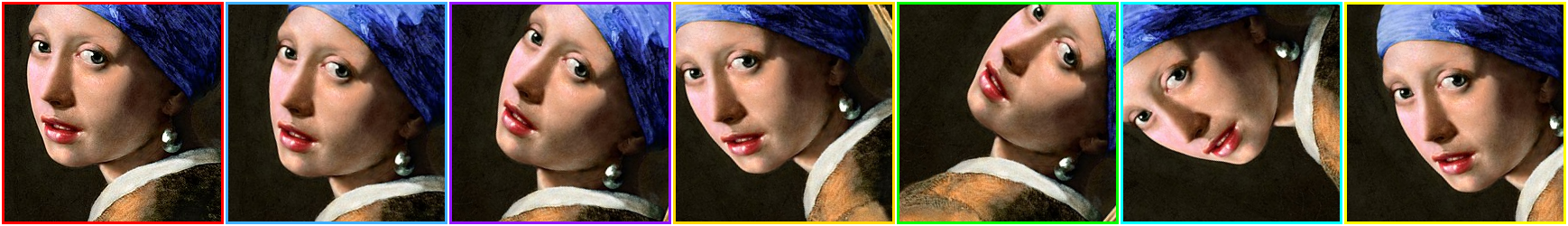}
\end{subfigure}
\caption{\textbf{Patch Reconstruction} Color-coded patch correspond to Fig~\ref{fig:planer_qualtiative}.Note that patch 2D rigid motion exhibits continuity over time (left to right) \label{fig:patch_vis}  }

\vspace{0.1cm}

\centering
\resizebox{0.75\linewidth}{!}{
\begin{tabular}{ccccccc}
\toprule
\multirow{ 2}{*}{Method}  & \multicolumn{3}{|c}{Cat }   \\
\cmidrule{2-4}
 & \multicolumn{1}{|c}{CE(pixel) $\downarrow$}  & PSNR $\uparrow$ & SR $\uparrow$  \\ 
\midrule
BARF\cite{barfoot2014batch}  & \multicolumn{1}{|c}{13.55} & 27.61 & 30\%  \\
B-spline  & \multicolumn{1}{|c}{35.14} & 21.95  & 0\%   \\
Ours  & \multicolumn{1}{|c}{\textbf{0.01}} & \textbf{37.00}  & \textbf{100\%}  \\
\midrule
& \multicolumn{3}{|c}{Girl }  \\
\midrule
BARF\cite{barfoot2014batch}   &  \multicolumn{1}{|c}{29.94}  & 22.09 &  15\%  \\
B-Spline & \multicolumn{1}{|c}{39.08}  & 19.42 & 10\%  \\ 
Ours  &  \multicolumn{1}{|c}{\textbf{6.92}}  & \textbf{32.40} & \textbf{95\%}    \\
\bottomrule
\end{tabular}}
\captionof{table}{\textbf{Image alignment experiment.} Qualitative results of average 20 sampled 2D rigid motion, CE refer to Corner Error and SR refer to successful rate.}
\label{tab:planar_qualitative}
\end{minipage}%
\hspace{1em}
\begin{minipage}[h]{0.63\textwidth}
\centering
\begin{subfigure}{0.24\textwidth}
\centering
\includegraphics[width=0.95\linewidth]{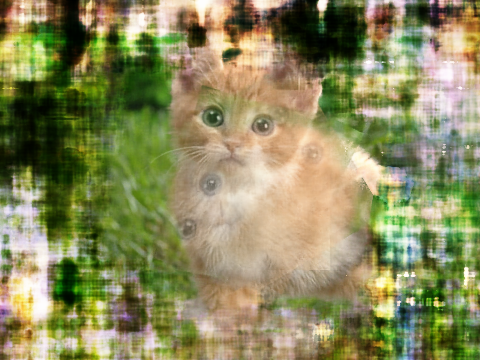}
\end{subfigure}%
\begin{subfigure}{0.24\textwidth}
\centering
\includegraphics[width=0.95\linewidth]{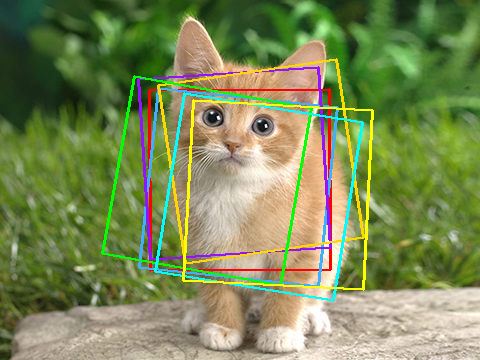}
\end{subfigure}%
\begin{subfigure}{0.24\textwidth}
\centering
\includegraphics[width=0.95\linewidth]{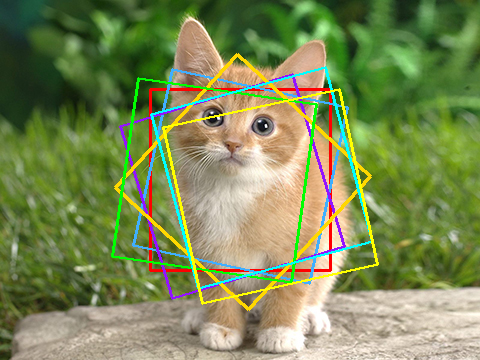}
\end{subfigure}%
\begin{subfigure}{0.24\textwidth}
\centering
\includegraphics[width=0.95\linewidth]{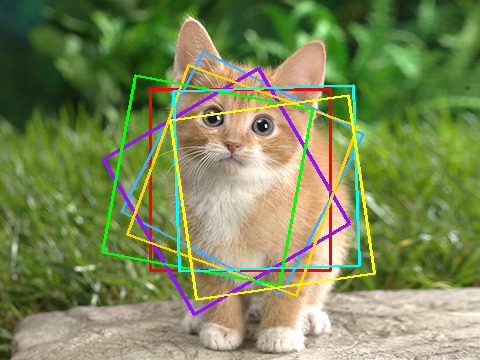}

\end{subfigure}
\begin{subfigure}{0.24\textwidth}
\centering
\includegraphics[width=0.95\linewidth]{figs/planar_exps/cat/barf/image_entire.png}
\end{subfigure}%
\begin{subfigure}{0.24\textwidth}
\centering
\includegraphics[width=0.95\linewidth]{figs/planar_exps/cat/barf/image_entire.png}
\end{subfigure}%
\begin{subfigure}{0.24\textwidth}
\centering
\includegraphics[width=0.95\linewidth]{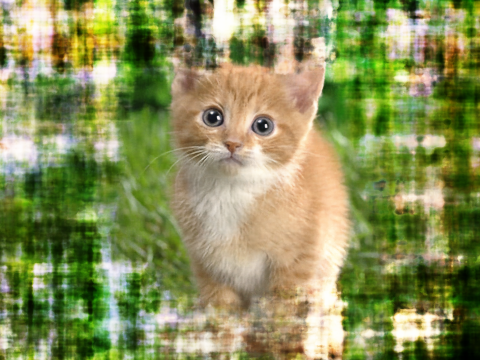}
\end{subfigure}%
\begin{subfigure}{0.24\textwidth}
\centering
\includegraphics[width=0.95\linewidth]{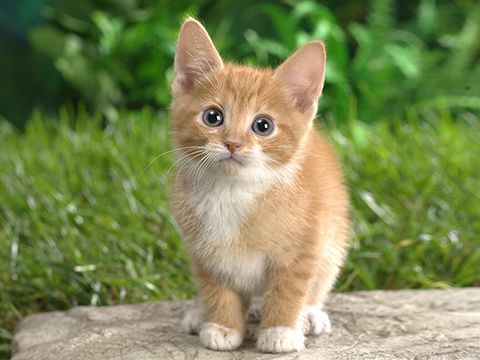}
\end{subfigure}

\begin{subfigure}{0.24\textwidth}
\centering
\includegraphics[width=0.95\linewidth]{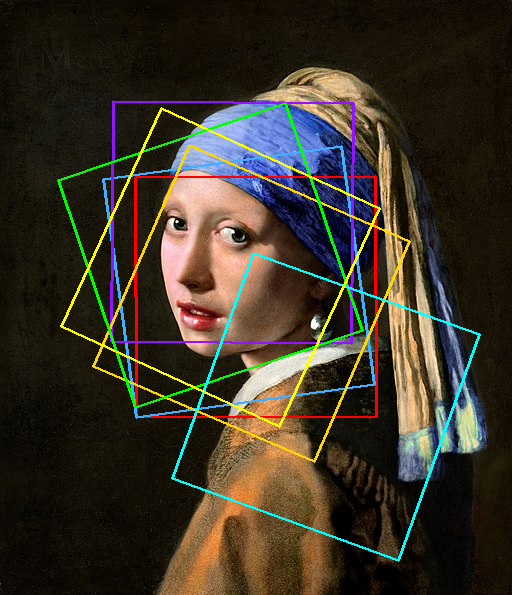}
\end{subfigure}%
\begin{subfigure}{0.24\textwidth}
\centering
\includegraphics[width=0.95\linewidth]{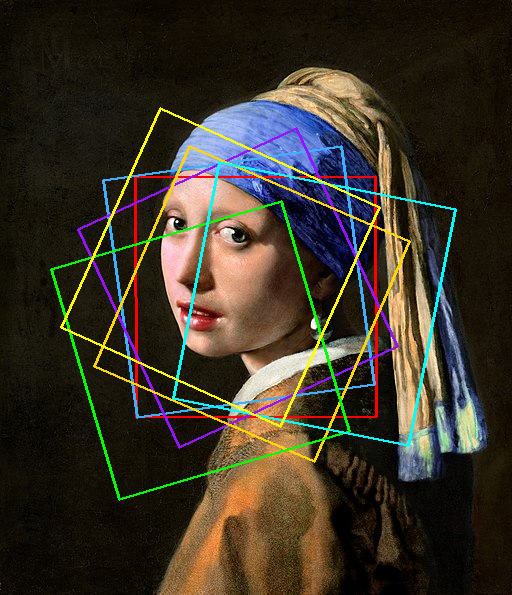}
\end{subfigure}%
\begin{subfigure}{0.24\textwidth}
\centering
\includegraphics[width=0.95\linewidth]{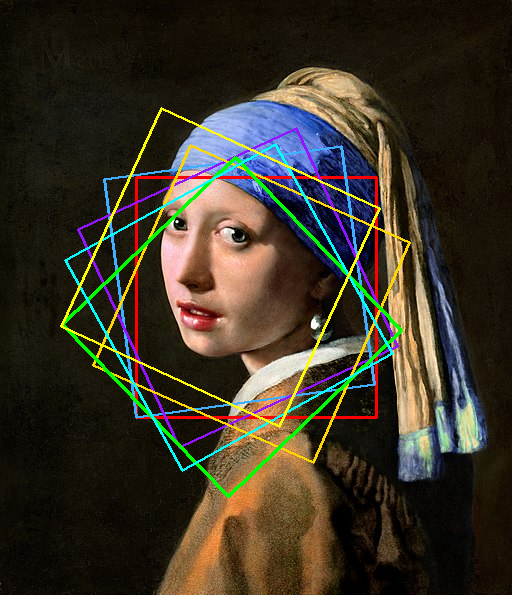}
\end{subfigure}%
\begin{subfigure}{0.24\textwidth}
\centering
\includegraphics[width=0.95\linewidth]{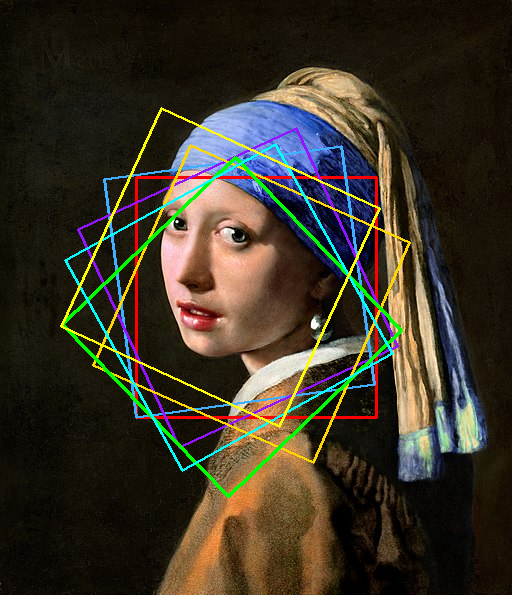}
\end{subfigure}
\vspace{0.1cm}
\begin{subfigure}{0.24\textwidth}
\centering
\includegraphics[width=0.95\linewidth]{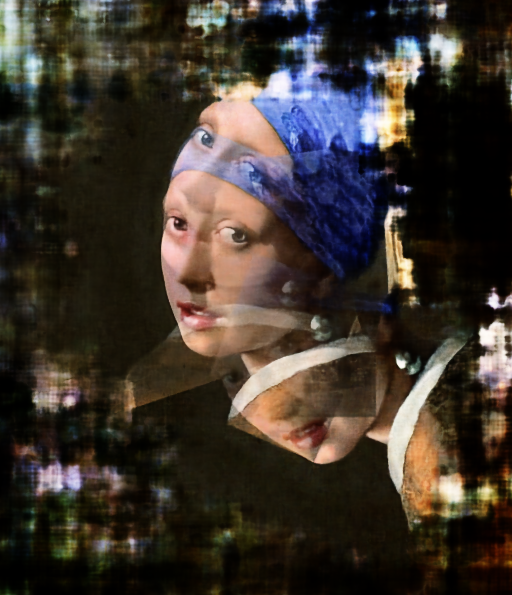}
\caption{BARF\cite{lin2021barf}}
\end{subfigure}%
\begin{subfigure}{0.24\textwidth}
\centering
\includegraphics[width=0.95\linewidth]{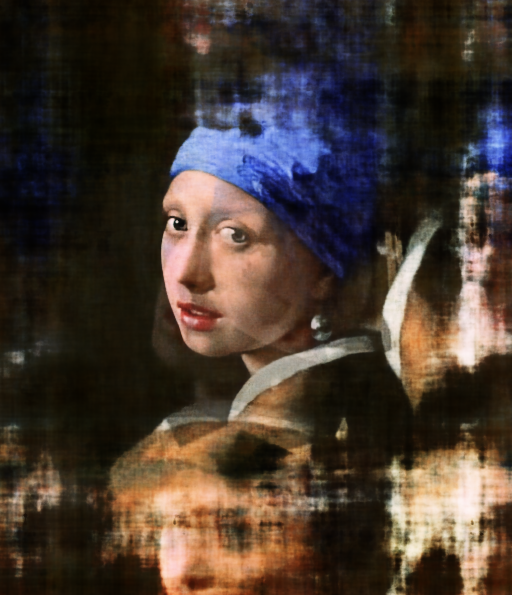}
\caption{B-Spline}
\end{subfigure}%
\begin{subfigure}{0.24\textwidth}
\centering
\includegraphics[width=0.95\linewidth]{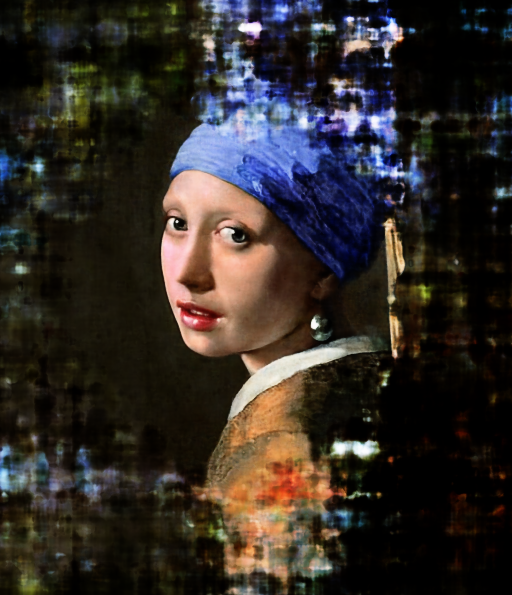}
\caption{Ours}
\end{subfigure}%
\begin{subfigure}{0.24\textwidth}
\centering
\includegraphics[width=0.95\linewidth]{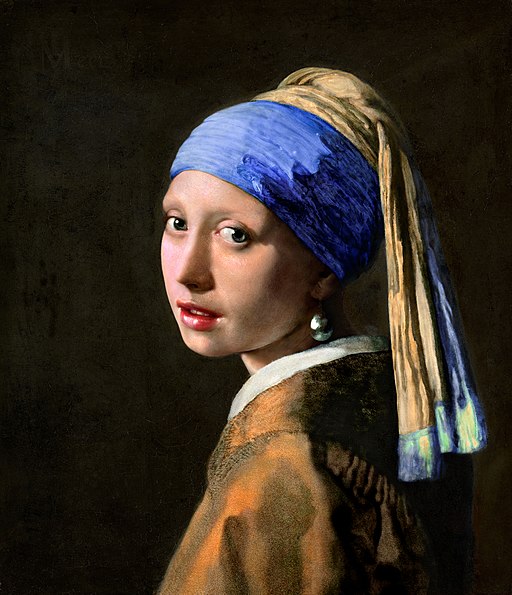}
\caption{GT}
\end{subfigure}
\captionsetup{width=0.9\linewidth}
\caption{\textbf{Qualitative results of 2D planar Alignment.} We report the qualitative results of planar image alignment. Given input as ground truth (d) shown in Fig.~\ref{fig:patch_vis}, the goal is to find the 2D rigid transformation for each patch and optimize the entire neural image. Our method optimizes for accurate alignment and high-fidelity image reconstruction, while baselines fail due to local minima. \label{fig:planer_qualtiative}}
\end{minipage}
\end{figure*}




\subsection{IMU fusion}

Up to our knowledge we are the first to integrate IMU data in NeRF + SLAM setting. The IMU fusion is straightforward in PoseNet taking advantage of the auto-differentiation of the neural network. We propose two different IMU fusion methods with details as follows:

\noindent \textbf{Loose coupling.} Given 3-axis angular velocity measurement from gyroscope $\hat{\omega} = (\hat{\omega}_{x}, \hat{\omega}_{y}, \hat{\omega}_{z} ) $  we get the time step from frequency $\triangle t = \frac{1}{f}$. We can express the rotation angle to be $\triangle t \| \hat{\omega}  \| $ around axis $ \frac{\hat{\omega}}{ \| \hat{\omega}  \|}$ \cite{sola2017quaternion}. This instantaneous rotation from the local sensor between previous and current timestamp can be represented as follows:
\begin{equation}
\mathsf{q}_{\triangle} =  \mathsf{q} \bigg( \triangle t \lVert \hat{\omega}  \rVert,  \frac{\hat{\omega}}{ \| \hat{\omega} \|} \bigg ).
\label{eq:delta_q}
\end{equation}

By continuously integrating the measurements we can get the rotation estimation at time $t_i$ with respect to $t_{j-1}$ from gyroscope: $\mathsf{q'}_{t_i} =  \mathsf{q}^{(t_{j-1})} \mathsf{q}_{\triangle}.$
We add $\ell_1$ loss $\mathcal{L}_{{loose}} = |q_{t_i} - q'_{t_i}|$ into Eq \ref{eq:obj_slam}, where $q_{g_j}$ integrate all gyroscope measurements from timestamps $t_{j-1}$ to $t_{j}$.

\noindent  \textbf{Tight coupling.} However, simply integrating IMU information leads to large drift and noise over time. As an immediate consequence of our continuous pose representation over time, we can directly fuse the angular velocity using the quaternion derivative~\cite{sola2017quaternion}: 
\vspace{-1ex}
\begin{equation}
\dot{\mathsf{q}} =  \frac{1}{2} \Omega(\hat{\omega}).
\label{eq:integrate}
\end{equation}

Thus we can supervise PoseNet by constraining the jacobian with the measured angular velocity. We use  $\ell_1$ loss as $ \mathcal{L}_{{tight}} = {| {\dot{\mathsf{q}} - \frac{1}{2} \Omega(\hat{\omega})} |}  $ and jointly optimize it with the tracking target function in Eq \ref{eq:obj_slam}.

It is noteworthy that, in the aforementioned equation, our PoseNet outputs pose with respect to the body frame rather than the camera frame. Further details regarding the coordinate change can be found in the supplementary materials, along with an explanation of how acceleration is utilized.

\begin{figure*}[t]
\begin{minipage}[t]{0.35\textwidth}
\begin{subfigure}{0.6\textwidth}
\centering
\includegraphics[width=0.8\linewidth, trim={0  0  0  2cm},clip]{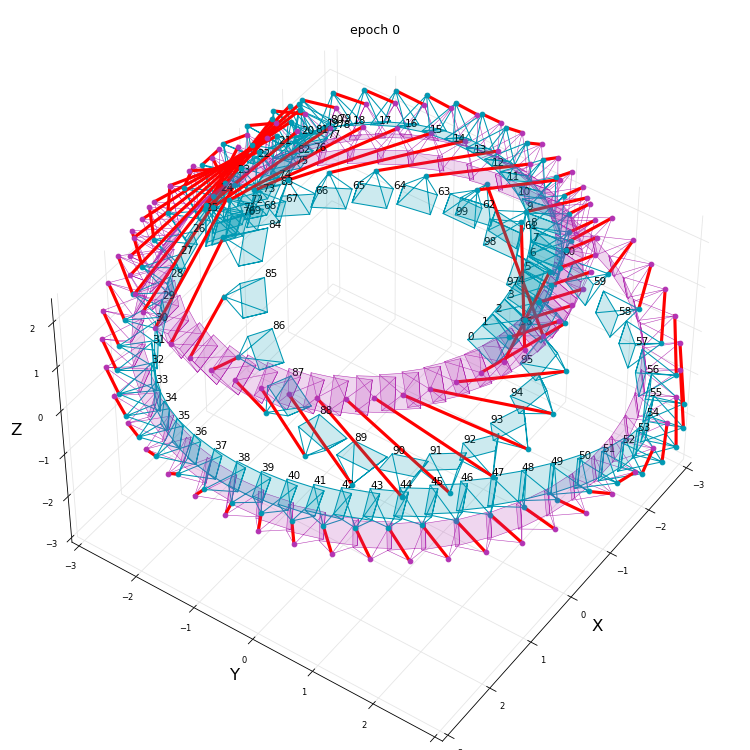}
\caption{initial camera pose}
\end{subfigure}%
\begin{subfigure}{0.4\textwidth}
\centering
\includegraphics[width=\linewidth]{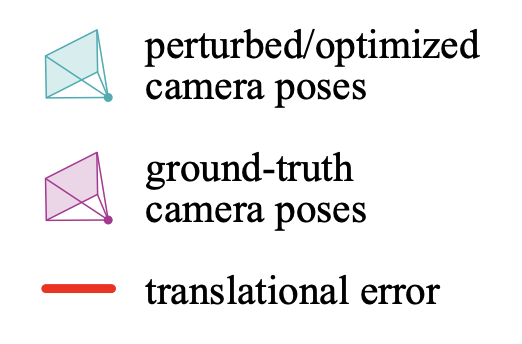}
\end{subfigure}%

\begin{subfigure}{0.5\textwidth}
\centering
\includegraphics[width=0.8\linewidth, trim={0  0  0  2cm},clip]{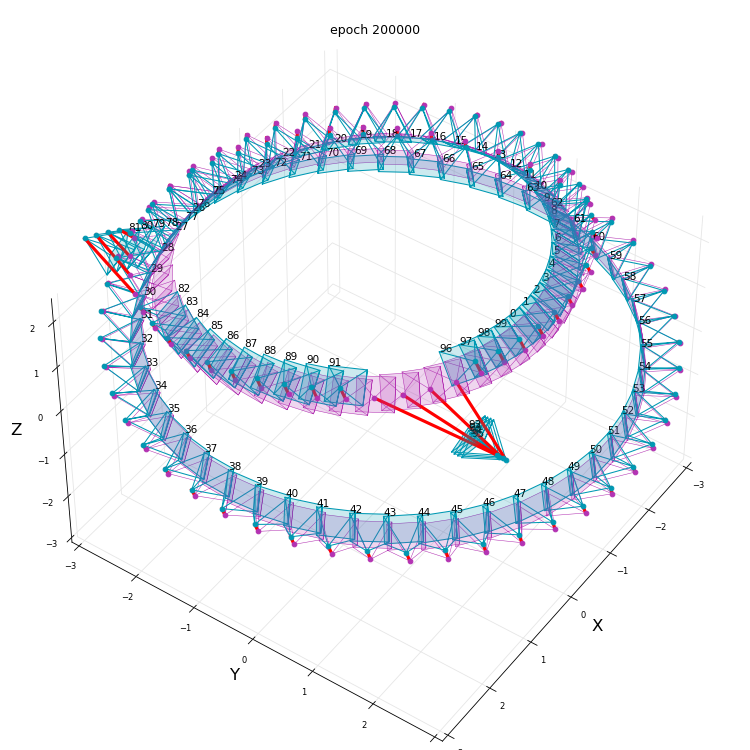}
\caption{BARF~\cite{lin2021barf}}
\end{subfigure}%
\begin{subfigure}{0.5\textwidth}
\centering
\includegraphics[width=0.8\linewidth, trim={0  0  0  2cm},clip]{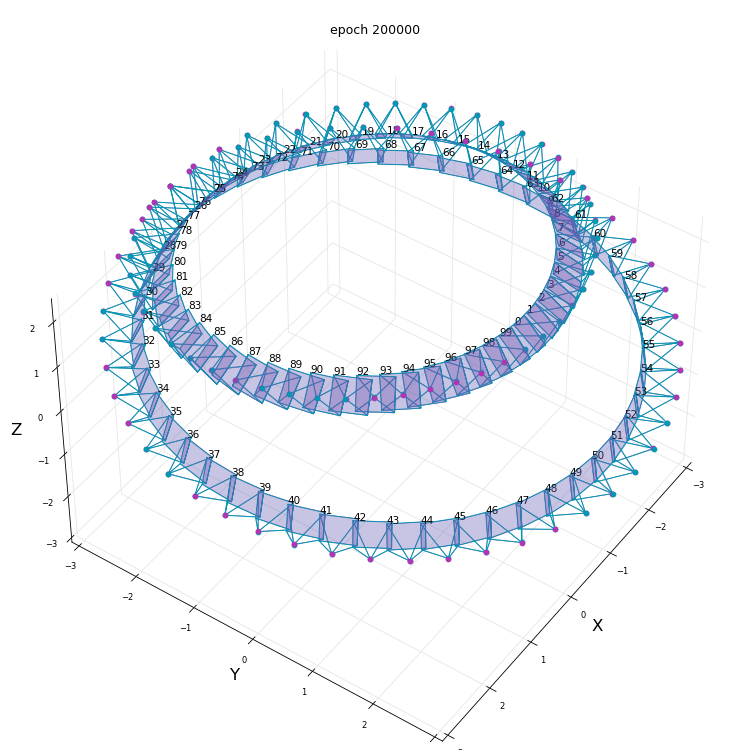}
\caption{Ours}
\end{subfigure}
\captionsetup{width=0.9\linewidth}
\caption{ We introduce continuous errors on the camera trajectories and perform pose refinement in the NeRF setting. (a) Initial pose error; (b) results obtained using BARF~\cite{lin2021barf} that uses a discrete set of poses; (c) results obtained using our continuous pose representation.}
\label{fig:lego_qualtivve}

\end{minipage}%
\begin{minipage}[t]{0.65\textwidth}
\resizebox{\linewidth}{!}{
\begin{tabular}{lcccccccccc}
\toprule

\multirow{ 2}{*}{Scene}  & \multicolumn{2}{|c|}{Rotation$\textdegree{}$ $\downarrow$} & \multicolumn{2}{|c|}{Translation $\downarrow$} & \multicolumn{2}{|c|}{PSNR $\uparrow$} & \multicolumn{2}{|c|}{SSIM $\uparrow$}  & \multicolumn{2}{|c}{LPIPS $\downarrow$}   \\
\cmidrule{2-11}
 & \multicolumn{1}{|c}{BARF}  & ours  & \multicolumn{1}{|c}{BARF}  & ours & \multicolumn{1}{|c}{BARF}  & ours &  \multicolumn{1}{|c}{BARF}  & ours & \multicolumn{1}{|c}{BARF}  & ours \\
\midrule
Fern& \multicolumn{1}{|c}{0.199}  & 0.181  & \multicolumn{1}{|c}{0.196} & 0.181  &  \multicolumn{1}{|c}{21.01}  & 21.08   & \multicolumn{1}{|c}{0.62}  & 0.63 & \multicolumn{1}{|c}{0.33}  & 0.31 \\
Fern$/2$& \multicolumn{1}{|c}{0.344}  & 0.331  & \multicolumn{1}{|c}{0.195} & 0.173  &   \multicolumn{1}{|c}{19.72} &  19.74 &  \multicolumn{1}{|c}{0.53}  &  0.53 &  \multicolumn{1}{|c}{0.33}  & 0.32  \\
Fern$/4$& \multicolumn{1}{|c}{0.289}  &  0.264 & \multicolumn{1}{|c}{0.212} & 0.215  &   \multicolumn{1}{|c}{19.65} &  21.33 &  \multicolumn{1}{|c}{0.54}  & 0.63 &  \multicolumn{1}{|c}{0.33}  & 0.32 \\

\midrule
Fortress  &  \multicolumn{1}{|c}{0.444}  & \multicolumn{1}{c|}{0.360}  &  0.369 & 0.283  & \multicolumn{1}{|c}{23.17}  & 22.17  &\multicolumn{1}{|c}{0.48}  & 0.41 & \multicolumn{1}{|c}{0.12}   & 0.12 \\
Fortress$/2$ \textbf{*}  &  \multicolumn{1}{|c}{6.507}  & 0.574 &  \multicolumn{1}{|c}{3.545}  & 0.418  & \multicolumn{1}{|c}{14.86}   & 20.00  &  \multicolumn{1}{|c}{0.35}  & 0.33  & \multicolumn{1}{|c}{0.40}   & 0.17 \\
Fortress$/4$  &  \multicolumn{1}{|c}{0.607}  &  0.630 &  \multicolumn{1}{|c}{0.583}  & 0.629 & \multicolumn{1}{|c}{20.71}   &  20.72 &  \multicolumn{1}{|c}{0.38}  & 0.40  & \multicolumn{1}{|c}{0.17}   & 0.20 \\

\midrule
Orchids   &  \multicolumn{1}{|c}{0.719}  & 0.645 &  \multicolumn{1}{|c}{0.390}  & 0.364  & \multicolumn{1}{|c}{13.22}   & 14.46  &  \multicolumn{1}{|c}{0.17}  &  0.24 & \multicolumn{1}{|c}{0.35}   & 0.30  \\
Orchids$/2$  &  \multicolumn{1}{|c}{0.809}  & 0.730 &  \multicolumn{1}{|c}{0.387}  & 0.375 & \multicolumn{1}{|c}{12.60}   &  13.50 &  \multicolumn{1}{|c}{0.15}  & 0.19  & \multicolumn{1}{|c}{0.37}   & 0.35  \\
Orchids$/4$ \textbf{*} &  \multicolumn{1}{|c}{92.176}  & 0.865 &  \multicolumn{1}{|c}{46.772}  &  0.388 & \multicolumn{1}{|c}{11.07}   &  12.64 &  \multicolumn{1}{|c}{0.18}  & 0.16  & \multicolumn{1}{|c}{0.97}   & 0.49 \\
\midrule

Room  &  \multicolumn{1}{|c}{0.288}  & 0.106 &  \multicolumn{1}{|c}{0.245}  &  0.101 & \multicolumn{1}{|c}{21.78}   &  25.32  &  \multicolumn{1}{|c}{0.79}  &  0.88 & \multicolumn{1}{|c}{0.14}   &  0.10  \\

Room$/2$ &  \multicolumn{1}{|c}{0.329}  & 0.274 &  \multicolumn{1}{|c}{0.284}  &  0.172  & \multicolumn{1}{|c}{21.20}   & 21.27  &  \multicolumn{1}{|c}{0.77}  & 0.78  & \multicolumn{1}{|c}{0.13}   & 0.13  \\

Room$/4$ \textbf{*}  &  \multicolumn{1}{|c}{118.58}  & 0.403 &  \multicolumn{1}{|c}{76.14}  & 0.550  & \multicolumn{1}{|c}{11,00}   &  22.76 &  \multicolumn{1}{|c}{0.42}  &  0.80 & \multicolumn{1}{|c}{0.89}   & 0.17 \\

\midrule
\multirow{2}{*}{Average} &  \multicolumn{1}{|c}{18.44}  & \multirow{1}{*}{\textbf{0.446}} &  \multicolumn{1}{|c}{10.777}  &  \multirow{1}{*}{\textbf{0.320}} & \multicolumn{1}{|c}{17.50}   &  \multirow{1}{*}{\textbf{19.58}} &  \multicolumn{1}{|c}{0.448}  &  \multirow{1}{*}{\textbf{0.498}} & \multicolumn{1}{|c}{0.378}   & \multirow{1}{*}{\textbf{0.248}} \\

&  \multicolumn{1}{|c}{(0.448)}  & \textbf{(0.391)}  &  \multicolumn{1}{|c}{(0.318)}  & \textbf{(0.276)}  & \multicolumn{1}{|c}{(19.23)}   & \textbf{(19.95)}  &  \multicolumn{1}{|c}{(0.492)}  & \textbf{(0.520)}  & \multicolumn{1}{|c}{(0.252)}   & \textbf{(0.238)}  \\

\bottomrule
\end{tabular}}

\vspace{1mm}
\captionof{table}{\textbf{Real data  with \underline{unknown pose}.} Our PoseNet compared to BARF~\cite{lin2021barf} for the real dataset, simulating different camera moving speeds. Whenever BARF diverges and provides very inaccurate results, we consider them failures and denote them as *. The average across all experiments is provided for all (and averaged only when BARF succeeds). In addition to the 12/12 (Ours) vs.\ 9/12 (BARF) success rate, PoseNet performs better than BARF also in cases when BARF succeeds.
\label{tab:barf_real_overall}}
\end{minipage}
\end{figure*}

\section{Experiments}
\subsection{NeRF from Inaccurate Poses}\label{subsec:inaccurate_pose}

We validate the effectiveness of our proposed method through 2D planar image alignment experiments and 3D scene experiments similar to BARF~\cite{lin2021barf}. During this process, BARF refines a discrete set of inaccurate camera poses while our method leverages the continuous pose information and is therefore less prone to local minima.
\vspace{-1em}
\subsubsection{Planar Image Alignment (2D)}
We choose the same images as \cite{lin2021barf, chen2023local} as shown in Fig \ref{fig:planer_qualtiative}. To obtain a continuous rigid transformation, we initially randomly sample 10 data points from  $T \in SE(2)$, we then interpolate a cubic spline along each dimension. Finally, we interpolate on $7$ uniformly spaced points at previous time instants. As a result, the rigid transformation demonstrates temporal correlation, as illustrated in Fig \ref{fig:patch_vis}. The initialized pose is identity with respect to center crop. 

\noindent \textbf{Experimental settings.} We compare our method against BARF \cite{lin2021barf} and BaRF with B-spline. For the latter, we introduce continuity by resetting the learned $T \in SE(2)$ for every $100$ steps using B-spline interpolation. The learning rate is $1e-3$ for translation and $2e-4$ for rotation. For the B-Spline method we report with 5 knots placed time-wise uniformly with $\text{degree} = 3$.

\noindent \textbf{Results.} The results are visualized in Fig \ref{fig:patch_vis},\ref{fig:planer_qualtiative}. The alignment performance of BARF suffers from local minima, resulting in sub-optimal performance. Experiments are deemed successful if the corner error is below 1 pixel. Although some patches correctly learn the transformation, they do not effectively contribute to neighbouring patches. Merely introducing B-spline directly does not work, as it can over-smooth or under-smooth the poses, whereas our proposed method successfully captures all rigid transformations resulting in high-fidelity neural image. Furthermore as demonstrated in Table~\ref{tab:planar_qualitative} our method performs consistently well across $20$ different trajectories.

\begin{table}
\resizebox{\linewidth}{!}{
\begin{tabular}{lccccc}
\midrule 
Method & RE $\downarrow$ & TE $\downarrow$  & PSNR $\uparrow$ & SSIM $\uparrow$ & LPIPS $\downarrow$  \\

\midrule
LE, C &  13.62 & 48.05  & 9.79  & 0.59  & 0.56 \\
Sinusoidal(2), C &  3.70 & 15.84  & 14.05  & 0.66  & 0.20 \\
Sinusoidal(5), C, &  \textbf{0.07} & \underline{0.32}  & \underline{27.25}  & \underline{0.91}  & \textbf{0.05} \\
Sinusoida(10), C &  \underline{0.18} & 0.88  & 24.88  & 0.88  & \underline{0.06} \\
\midrule
Sinusoidal(2), D  &  2.86 & 10.53  & 15.97  & 0.69  & 0.14 \\
Sinusoidal(5), D &  \textbf{0.07} &  \textbf{0.28} & \textbf{27.30}  & \textbf{0.92}  & \underline{0.06 }\\
Sinusoidal(10), D & \textbf{0.07}   & \textbf{0.28}   & 27.20  & \underline{0.91}  & \textbf{0.05}  \\
Sigmoid, D  & 14.31  & 37.21  & 11.28  & 0.67  & 0.55 \\
Sinusoidal(10) c2f, D  & 14.74   & 49.07   &  9.78 & 0.60  & 0.60 \\
\bottomrule
\end{tabular}}
\vspace{1mm}
\caption{\textbf{Ablation study.} We investigate the effectiveness of our PoseNet with diverse architecture. LE refers to linear encoder and C, D refer to coupled and decoupled representations. RE, TE refer to rotational and translation error. The best and second-best results are in bold and underlined.
}
\label{tab:barf_encoder}
\end{table}

\subsubsection{Synthetic and Real NeRF (3D)}
We explore the more challenging problem of learning 3D Neural Radiance Field with inaccurate poses. For the synthetic data, we render Lego~\cite{mildenhall2021nerf} with a circular movement as shown in Figure~\ref{fig:lego_qualtivve}. The simulated camera orbits the Lego model in the xy-plane, moving up and down at a constant speed in the z-direction.

\noindent \textbf{Experimental settings.} Similar to the 2D experiment, we introduce temporal correlation between neighboring $SE(3)$ disturbances with interpolation. We use spherical linear interpolation for rotation. The introduced error corresponds to $55^\circ$ in rotation and 110\% in translation. For real data, we use the Fern, Fortress, Orchids, and Room datasets in LLFF~\cite{mildenhall2019llff}, since these sequences allow us to perform experiments with varying numbers of images, thus simulating fast-moving cameras. Unlike in the synthetic case, \emph{ we do not use any pose initialization} in the real data experiments. Following \cite{lin2021barf} we report the MSE distance and rotational angle after alignment using Procrustes analysis for registration evaluation and PSNR, SSIM~\cite{wang2004image} and LPIPS~\cite{zhang2018unreasonable} to evaluate view synthesis quality.

\noindent \textbf{Results.} We report our experimental results in Table~\ref{tab:barf_encoder} and Table~\ref{tab:barf_real_overall}, for synthetic and real data, respectively. In Table~\ref{tab:barf_real_overall}, the proposed continuous pose representation clearly offers better results than the discrete BARF. Ablation experiments further illustrate that the rotation and translation decoupled representation, i.e., two MLPs instead of one, performs better, offering the best results with the embedding frequency bands $F = 5$.
In real data experiments with completely unknown camera poses, PoseNet performs significantly better than BARF. These results are reported in Table~\ref{tab:barf_real_overall}, where dataset/n refers to $1/n^{th}$ fraction of uniformly downsampled cases. It can be seen that PoseNet successfully handles all three failure cases of BARF. This is particularly evident when only sparse image frames are available. At the same time, even in the cases when BARF succeeds, PoseNet performs significantly better than BARF. More results and experiments using B-Spline can be checked in supplementary material.

\begin{table}[t]
\resizebox{\linewidth}{!}{
\begin{tabular}{ccccccccc}
\toprule

 & \multicolumn{2}{|c|}{TE $\downarrow$ } & \multicolumn{2}{|c|}{PSNR $\uparrow$  } & \multicolumn{2}{|c|}{SSIM $\uparrow$}  & \multicolumn{2}{|c}{LPIPS $\downarrow$ }   \\
\cmidrule{2-9}
 Num & \multicolumn{1}{|c}{without}  & ours & \multicolumn{1}{|c}{without}  & ours &  \multicolumn{1}{|c}{without}  & ours & \multicolumn{1}{|c}{without}  & ours \\
\midrule

&  \multicolumn{8}{c}{Chair} \\

\midrule

20  & \multicolumn{1}{|c}{3.66 } & \textbf{1.74}  &  \multicolumn{1}{|c}{26.36}  & \textbf{26.58}   & \multicolumn{1}{|c}{0.89}  & \textbf{0.91} & \multicolumn{1}{|c}{0.19}  & \textbf{0.15} \\

10 & \multicolumn{1}{|c}{15.63 } & \textbf{3.38}  &  \multicolumn{1}{|c}{22.48}  & \textbf{25.02}   & \multicolumn{1}{|c}{0.81}  & \textbf{0.86} & \multicolumn{1}{|c}{0.34}  & \textbf{0.18} \\

 6 & \multicolumn{1}{|c}{59.31 } & \textbf{22.68}   &  \multicolumn{1}{|c}{21.45}  & \textbf{22.06}  & \multicolumn{1}{|c}{0.70}  & \textbf{0.80}  & \multicolumn{1}{|c}{0.57}  & \textbf{0.34} \\

\midrule
 &  \multicolumn{8}{c}{Hotdog}  \\
 
\midrule

 20  & \multicolumn{1}{|c}{ 3.66} &  \textbf{2.42} &  \multicolumn{1}{|c}{23.59}  & \textbf{25.64}   & \multicolumn{1}{|c}{0.90}  & \textbf{0.92} & \multicolumn{1}{|c}{0.14}  & \textbf{0.10} \\

 10 & \multicolumn{1}{|c}{15.63 } & \textbf{4.87}  &  \multicolumn{1}{|c}{21.85}  & \textbf{23.15}   & \multicolumn{1}{|c}{0.85}  & \textbf{0.87} & \multicolumn{1}{|c}{0.20}  & \textbf{0.16} \\

 6 & \multicolumn{1}{|c}{59.31 } & \textbf{6.70}  &  \multicolumn{1}{|c}{21.06}  & \textbf{22.03}   & \multicolumn{1}{|c}{0.79}  & \textbf{0.85} & \multicolumn{1}{|c}{0.34}  & \textbf{0.18} \\


\bottomrule
\end{tabular}}
\caption{\textbf{Interpolation error experiments.} We improve the EventNeRF~\cite{rudnev2022eventnerf} using the proposed PoseNet. A small number of sparsely known poses are used to estimate the poses in between. Our method improves EventNeRF significantly in all six experimental setups.}
\label{tab:eventnerf_interpolation}
\end{table}

\begin{figure}[t]
  \centering
  \resizebox{\linewidth}{!}{
  \begin{tabular}{ccc}
     & offset 0.2388 \textdegree{} & offset 2.85 \textdegree{} \\
    \multirow{2}{*}{\includegraphics[width=3.5cm]{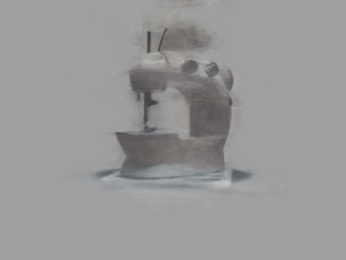}} & \includegraphics[width=3.2cm]{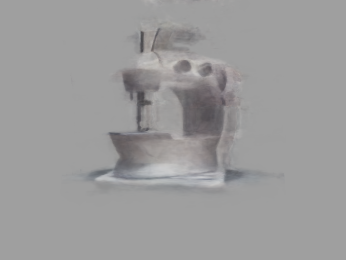}  & \includegraphics[width=3.2cm]{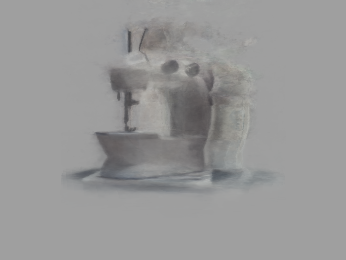}  \\
     & \multicolumn{2}{c}{EventNeRF} \\

     & \includegraphics[width=3.2cm]{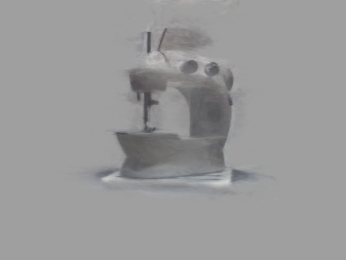}  & \includegraphics[width=3.2cm]{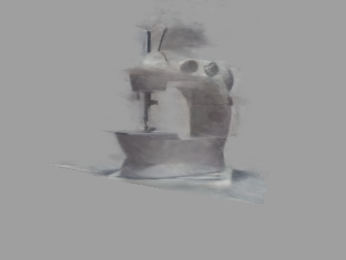} \\

     RGB  & \multicolumn{2}{c}{Ours} \\

     with known calibration & \multicolumn{2}{c}{} \\
  \end{tabular}}

\caption{\textbf{With and without calibration experiments.}  We investigate the effectiveness of our method under different deviations from the actual rotational axis. Our method can successfully reposition the object back to the center without additional calibration.}
\label{fig:calibration}
\end{figure}

\subsection{Continuous Pose for Asynchronous Events}\label{subsec:events}

By virtue of continuous pose representation, handling asynchronous event streams acquired by event cameras becomes natural. Hence, we use our PoseNet to learn the radiance field-based 3D scene representation from only colour event streams. This experimental setup is similar to recent work EventNeRF~\cite{rudnev2022eventnerf}. Note that EventNeRF accumulates asynchronous events to high-frequency synchronous event frames. The poses of each of those event frames are then assumed to be known. We argue that these assumptions limit the potential of the event cameras which come from their asynchronous nature. Therefore, we query for the pose of every event precisely at their trigger times. We conduct two experiments to address two practical issues of using events in EventNeRF setup using both synthetic and real datasets.

\vspace{-1em}
\subsubsection{Unknown continuous pose for single event}
Events are triggered asynchronously, and in practice where there is no precisely measured control available such as with a turntable ~\cite{rudnev2022eventnerf} or a motorized linear slider ~\cite{rebecq2016emvs}, event pose can only be interpolated from measured discrete poses (from Vicon or Colmap \cite{hidalgo2022event}). However, this introduces interpolation errors. 

\noindent \textbf{Experimental settings.} For synthetic data, we use \emph{chair} and \emph{hotdog} sequences from \cite{rudnev2022eventnerf}. The events are simulated using the model in~\cite{rudnev2021eventhands}. While EventNeRF performs interpolation, our method jointly learns intermediate poses as a continuous function of time. 


\noindent \textbf{Results.} In Table~\ref{tab:eventnerf_interpolation}, we reveal that integration of our PoseNet significantly enhances the overall performance with a notable reduction in translation errors and better scene reconstruction. More visual results can be found in supplementary material.

\subsubsection{Unknown calibration in practice}
EventNeRF~\cite{rudnev2022eventnerf} uses turntable to achieve stable and consistent object rotation speed. This setup also requires the actual rotational axis. Therefore, an additional checkerboard-based calibration technique, to estimate the axis offset, is also proposed in~\cite{rudnev2022eventnerf}. 

\noindent \textbf{Experimental setting.} For real cases, we use \emph{sewing machine} datasets, which hold difficulties in reconstructing thin structures, view-dependent effects, and colored texts. 

\noindent \textbf{Results.} We show that when PoseNet is used, additional calibration may not be required. The qualitative results of these experiments are shown in Figure~\ref{fig:chick_and_sew}. We demonstrate that when some offset is introduced, the 3D object deviates from the center for EventNeRF, while our method can reduce artifacts, learn the offset angle, and reposition the object back to the image center.

\subsection{Visual SLAM with Depth and IMUs}\label{subsec:vslam}
While the previously discussed tasks are offline, vSLAM is an online method with different considerations. In this application we approach the problem as incremental SLAM. For each incoming frame, our objective is to determine its transformation with respect to the last frame $T_{relative}$. Similar to NICE-SLAM \cite{zhu2022nice} we maintain a list of all optimal relative poses. It is trivial to solve the forgetting issue by retraining our PoseNet with such a list. 

\noindent \textbf{Experimental settings.} We report the tracking results of our method compared with the standard NICE-SLAM. We report results of intrinsic motion on Replica \cite{straub2019replica}, Scannet \cite{dai2017scannet} and TUM-RGBD\cite{sturm12iros}. Note that during tracking we assume intrinsic motion reference slowly changes over time and only optimize $f_o$ for every keyframe, with a frequency set to $10$ for our experiments.
In EUROC dataset \cite{burri2016euroc} we follow the same pre-processing step as \cite{gordon2019depth} and use nearest interpolation to get dense depth map. In order to evaluate the trajectory quality we report the ATE-RMSE [cm] of all sequences. More details regarding convergence rate and run-time can be found in the supplementary material.

\noindent \textbf{Results.} We report all tracking results using ATE-RMSE [cm]. The numbers for the baselines are taken from \cite{sandstrom2023point} except EUROC. We showcase the effectiveness of our method for tracking across all scenes in Table~\ref{tab:replica},~\ref{tab:scannet},~\ref{tab:tum-rgbd}. We observe significant improvements in both relatively easy and challenging scenarios. 

\begin{figure}
    \centering
    \includegraphics[width=0.9\linewidth, trim={1cm 0 0.5cm 1cm},clip]{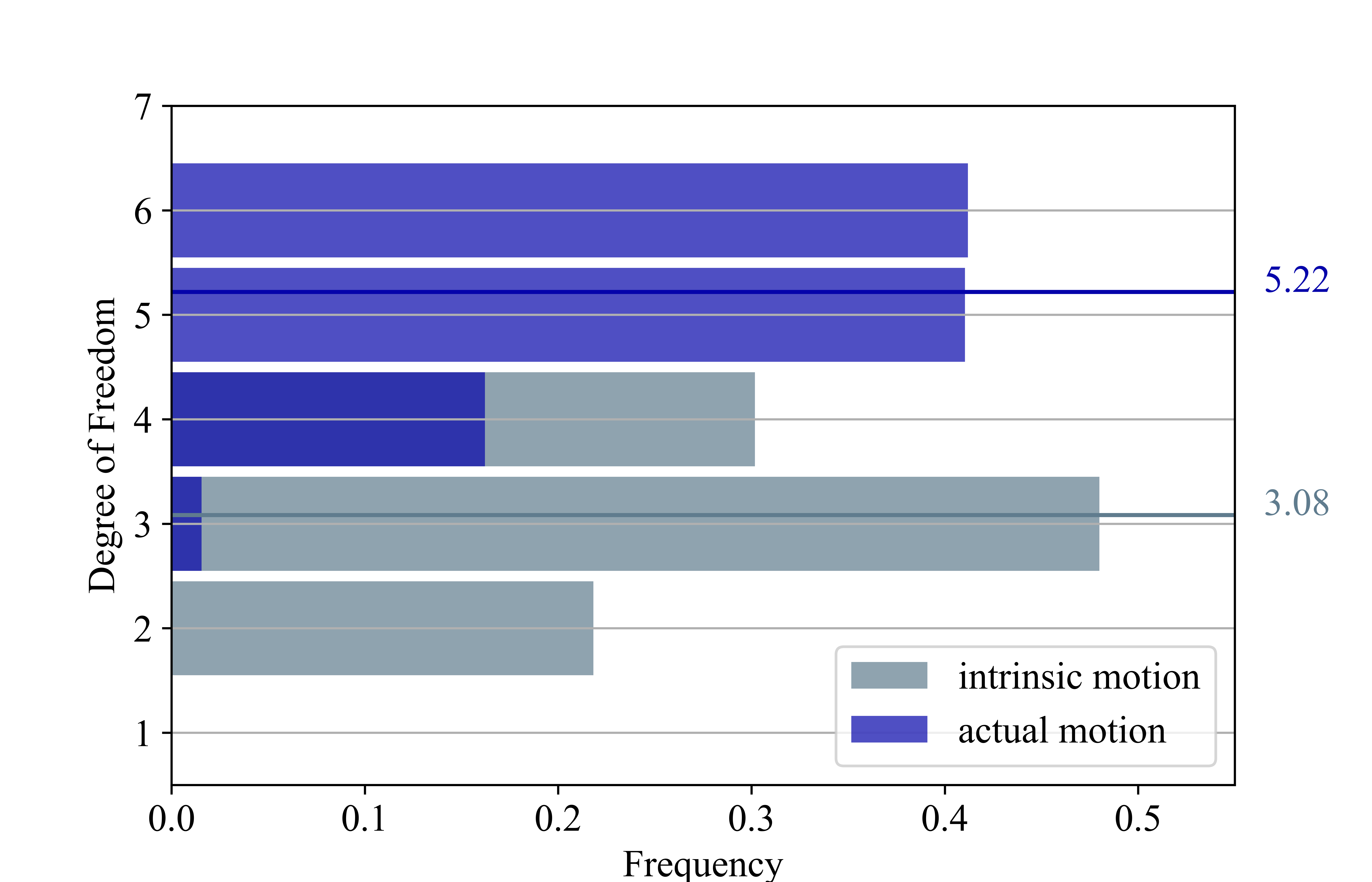}
    \caption{{DOF Comparison} on Replica room 1 dataset, we report that DOF of actual motion drop 41\% from 5.22 to 3.08, demonstrating the sparsity of intrinsic motion.}
    \label{fig:dof}
\end{figure}

Moreover as illustrated in Table \ref{tab:replica}, we underscore the importance of defining the coordinate system for relative pose optimization. The tracking is unstable and difficult when fixed on world origin or random coordinates. Figure \ref{fig:dof} further demonstrates that, through our estimation of intrinsic motion and its transformation with PoseNet, we attain pose within a low-dimensional manifold, resulting in a substantial enhancement of tracking performance. 

Finally, we validate the effectiveness of our IMU-Fusion method. While baseline methods fail in the face of large illumination changes and noisy depth, our approach maintains robust tracking and achieves accuracy comparable to state-of-the-art sparse feature-based tracking methods.

\begin{table}
\resizebox{\linewidth}{!}{
\begin{tabular}{lcccccccccc}
\toprule        
Method & Rm 0 & Rm 1 & Rm 2 & Off 0 & Off 1 & Off 2 & Off 3 & Off 4 & Avg  \\ 
\midrule

Vox-Fusion* \cite{yang2022vox} & 1.37 0 & 4.70 & 1.47 & 8.48 & 2.04 & 2.58 & 1.11 & 2.94 & 3.09  \\ 

ESLAM\cite{johari2023eslam}  & 0.71 0 & 0.70 & \textbf{0.52} & 0.57 & 0.55 & 0.58 & 0.72 & 0.63 & 0.63  \\ 

NICE-SLAM\cite{zhu2022nice}  & 0.97 0 & 1.31 & 1.07 & 0.88  & 1.00 & 1.06 & 1.10 & 1.13 & 1.06  \\ 

\midrule
Ours  & \textbf{0.53} & \textbf{0.45} & 0.84  & 0.54  & 0.33 & 0.48 & 0.66 & 0.51 & 0.54  \\ 
Ours(world)  & 0.62  & 0.52 &  0.91  & 0.60 & 0.62 & 0.36 & 0.54 & 0.72 & 0.58\\ 
Ours(rand)  & 35.84  & 9.29 &  34.67 & N/A  &  9.69 & 26.92 & N/A & N/A & N/A  \\ 
Ours (intrinsic)  & \textbf{0.53} & 0.47 &  0.81  &  \textbf{0.35} &  \textbf{0.24} & \textbf{0.43} & \textbf{0.64} & \textbf{0.50} & \textbf{0.49} \\ 

\bottomrule

\end{tabular}}
\caption{\textbf{Tracking performance on Replica~\cite{straub2019replica}}. By integrating our method into the tracking branch of NICE-SLAM, we observe significant improvements. We investigate the impact of varying reference coordinates on tracking. It is evident that our proposed low DOF motion further improves the tracking performance.}
\label{tab:replica}
\end{table}

\begin{table}
\resizebox{\linewidth}{!}{
\begin{tabular}{lcccccc}
\toprule        
Method & 0000 & 0059 & 0106 & 0181 & 0207 & Avg  \\ 
\midrule
DI-Fusion \cite{huang2021di} & 62.99 & 128.00 & 18.50 & 87.88 & 100.19 & 78.89 \\ 
Vox-Fusion* \cite{yang2022vox} & 68.84  & 24.18 & 8.41 & 23.30 & 9.41 & 26.90   \\ 
NICE-SLAM\cite{zhu2022nice} & 12.00  & 14.00 & 7.90 & 13.40 & 6.20  & 10.70    \\ 

\midrule
Ours  & \textbf{10.98}  & 11.98 & \textbf{7.10} &  13.50  &  5.76 & 9.86 \\ 

Ours(intrinsic) & 11.21 & \textbf{8.78} & 7.57  & \textbf{12.21}   & \textbf{4.87} & \textbf{8.93} \\ 

\bottomrule

\end{tabular}}
\caption{\textbf{Tracking performance on ScanNet~\cite{dai2017scannet}.} Our approach yields consistently better results than the baseline. Note that the gain of utilizing intrinsic motion is relatively small, possibly attributed to the challenges posed by the noisy ground truth poses. }
\label{tab:scannet}
\end{table}

\begin{table}
\centering
\resizebox{0.9\linewidth}{!}{
\begin{tabular}{lcccc}
\toprule        
Method & fr1/desk & fr2/xyz & fr3/office & Avg  \\ 
\midrule
DI-Fusion \cite{huang2021di} & 4.4 & 2.0 & 5.8 & 4.1 \\ 
Vox-Fusion* \cite{yang2022vox} & 3.52  & 1.49 & 26.01 & 10.34   \\ 
NICE-SLAM\cite{zhu2022nice} & 4.26 & 31.73 & 3.87 &  13.28    \\ 

\midrule
Ours  & 2.97 & 7.38 & 3.76  & 4.70 \\ 
Ours(intrinsic) & \textbf{2.72} & \textbf{1.98} & \textbf{2.74} & \textbf{2.48}  \\ 

\bottomrule

\end{tabular}}
\caption{\textbf{Tracking performance on TUM-RGBD~\cite{sturm12iros}} Our method consistently outperforms NICE-SLAM and other dense neural RGBD methods. The effectiveness of intrinsic motion is also demonstrated for reducing the tracking error significantly.}
\label{tab:tum-rgbd}
\end{table}

\begin{table}
\resizebox{\linewidth}{!}{
\centering
    \centering
    \vspace{1mm}

    \centering
        \begin{tabular}{lccccccc}
        \toprule
        Method & v101 & v102 & v103 & v201& v202 & v203  & Avg  \\ 
        \midrule
        VINS-MONO\cite{qin2017vins}  & 7.9  & 11.0 & 18.0 & 8.0  & 16.0 & 27.0 & 14.6   \\ 
        ORB-SLAM \cite{mur2015orb} & \textbf{1.5} & 2.0 & N/A & 2.1 & 1.8 & N/A & N/A \\ 
        DROID-SLAM \cite{teed2021droid} & 3.7 & \textbf{1.2} & \textbf{2.0} & \textbf{1.7} & \textbf{1.3} & \textbf{1.4} & \textbf{2.2} \\ 
        \midrule
        NICE-SLAM\cite{zhu2022nice}  & 2.58  & N/A  & 5.66 &  6.56  & N/A & N/A & N/A \\
        Ours(loose) & 2.20  & 6.74  &\textbf{5.04} &  \textbf{4.52}  & 3.87 & 19.06 & 6.77 \\  
        Ours(tight)  & 1.98  & \textbf{6.09}  & 5.55 &  4.99  & \textbf{3.03} & \textbf{15.34} & \textbf{6.16}  \\ 
        \bottomrule
        \end{tabular}}

    \caption{\textbf{Tracking performance on EUROC~\cite{burri2016euroc}.} Our IMU-fusion improves tracking with lower error and robustness, outperforming NICE-SLAM. We report results with sparse tracking method for reference. Despite the gap, our method narrows differences with state-of-the-art sparse tracking.}
\label{tab:nice-slam-IMU}

\end{table}

\section{Conclusion}
We proposed a simple yet effective technique for optimizing camera pose as a continuous function of time. The benefits of this approach were illustrated through several experiments of diverse applications, namely NeRF from the inaccurate pose, NeRF using Event Cameras, and visual SLAM with Depth and IMUs. We also studied different designs of the time-to-pose mapping continuous function, leading us to prefer a decoupled architecture. Furthermore, we justified the ease of using the proposed PoseNet in a plug-and-play manner. We first propose IMU-Fusion in NeRF-SLAM and analyze the advantage of adopting intrinsic motion frame for camera tracking tasks. Clear advantages in terms of performance were also observed in all settings, thanks to the continuous motion prior. 


\paragraph{Acknowledgements:} Research is partially funded by VIVO Collaboration Project and also partially funded by the Ministry of Education and Science of Bulgaria (support for INSAIT, part of the Bulgarian National Roadmap for Research Infrastructure).

\clearpage
\setcounter{page}{1}
\maketitlesupplementary

In this document, we provide additional details of our method and implementations. We further provide qualitative examples corresponding to the main paper results. Please also refer to the supplementary video for additional qualitative visualizations. We have also attached an example code in a separate file.

\section{PoseNet Coordinate Frames}
\label{sec:math_explainn}
In this section, we report the details of the PoseNet outputs concerning the coordinate frame under different applications. For the first two applications involving inaccurate pose and asynchronous events, we follow the work \cite{lin2021barf} and output refined pose from the noisy pose with respect to the $i$th camera frame. $ T_{c_{i}', w} = T_{init_i} \circ T_{refine_i} $, $ T_{init_i} = T_{c_{i}, w} \circ T_{noise_i}$, while $T_{a,b}$ represents the rigid-body transformation matrix that transforms homogeneous points defined in frame $b$ to the equivalent points in frame $a$. Note $c_{i}i$ refers to pose estimation of camera $i$ and $c_{i}$ denotes the ground truth. The target of PoseNet is to learn the cancellation of noise perturbations, essentially to serve as the inverse of $T_{noise_i}$. In the real experiment, we assume unknown initial pose so $T_{c_{i}', w} = I \circ T_{refine_i} $ making the objective of PoseNet to directly estimate $T_{c_{i}, w}$.

In the RGB-D SLAM application in Table \ref{tab:replica}, we analyze the impact of varying reference coordinates on tracking. We denote the PoseNet output with respect to frames $x$. So the estimation of $i$the camera pose: $T_{w,c_{i}} = T_{w,c_{i-1}} \circ T_{c_{i-1},x} \circ P(f(\theta_p, t_i)) $.  The output with respect to different reference frames is shown in Table \ref{tab:coordinates}. Note we get the random frame by perturbing the pose of $c_{i-1}$.

\begin{table}[b]
    \centering
    \begin{tabular}{cc}
    \toprule    
          Reference Frame &  Transformation \\ 
    \midrule
          Default &   $T_{c_{i-1}, c_{i}}$   \\
          World  &    $T_{w, c_{i}}$  \\
          Random &    $T_{r, c_{i}}$ \\ 
          Intrinsic  &    $T_{I, c_{i}}$ \\
          IMU  &  $T_{b_{i-1}, b_{i}}$\\
    \bottomrule
    \end{tabular} 
    \caption{\textbf{PoseNet on different reference coordinates.}}
    \label{tab:coordinates}
\end{table}

\begin{figure}
    \centering
    \includegraphics[width=\linewidth]{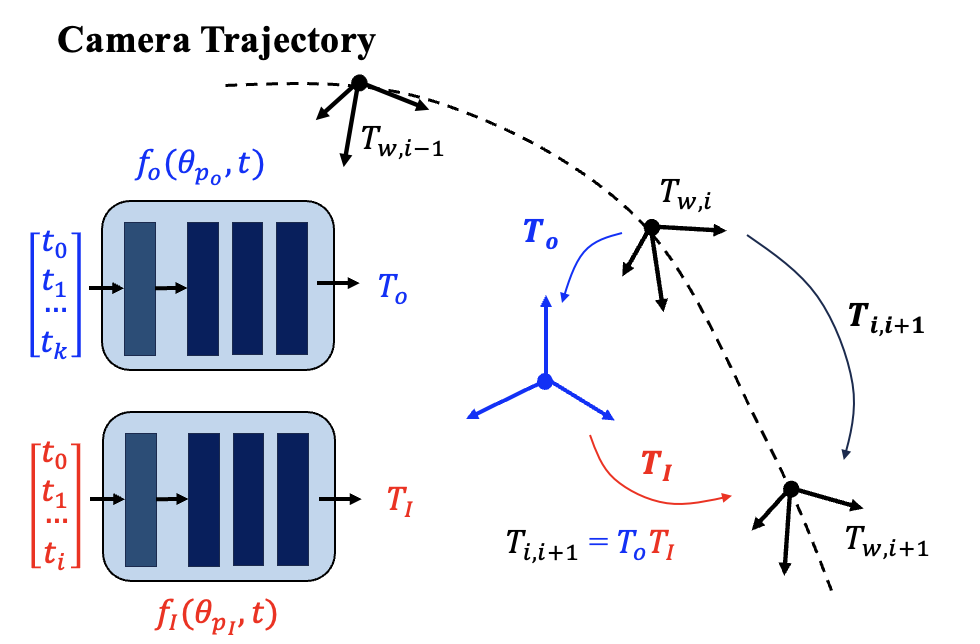}
    \caption{\textbf{Intrinsic Motion Frame.} We decompose the relative motion $T_{i,i+1}$ with a slowly changing rigid transform \textcolor{blue}{$T_0$} and a low dimensional frame-wise motion \textcolor{red}{$T_I$} using two separate PoseNets.}
    \label{fig:intr-explain}
\end{figure}

For the experiments of IMU, PoseNet outputs the pose of the agent which is fixed as the IMU sensor. Then we transform the pose to camera frame with $T_{w,c_{i}} = T_{w,b_{i}} \circ T_{b_{i},c_{i}}$ while $_{b_{i},c_{i}}$ is constant and read from the sensor extrinsic.

\section{NeRF from Inaccurate Pose}
\label{sec:supp_noisy_pose}

\paragraph{Implementation details.}
Compared to \cite{lin2021barf} we make the following modifications and extensions: (1) BARF perturbs the ground truth pose in synthetic datasets by independently sampling 6 dimensions Gaussian noise in  $SE(3)$. We introduce time-dependent noise which is closer to the real-world scenario, for monocular cameras, where the error of pose estimation increases with time due to drift and error accumulation. Furthermore, we also separate the rotation and translation perturbation and instead of sampling noise across all frames we only sample a subset of frames and interpolate the poses for the rest. By doing so we can explicitly set the maximal deviation on translation or rotation. (2) Unlike BARF when we optimize one camera pose it also affects the surrounding poses, therefore a larger batch size is important for stable training. We use 4096 random rays for each iteration to optimize camera poses collectively. 

For joint training with the radiance field, we use the Adam~\cite{kingma2014adam} optimizer for both translation and rotation networks with different learning rates. We use a smaller learning rate for rotation since quaternion rotation expression is highly nonlinear and difficult to train compared to translation~\cite{li2018undeepvo}. We use 1e-3 for TransNet and 2e-4 for RotsNet and exponentially decay the schedule to 1e-5 and 1e-6 respectively for stable training.


\begin{table}[b]
\resizebox{\linewidth}{!}{
\begin{tabular}{lccccc}
\toprule
Parameter & Rotation error $\downarrow$ & Translation error $\downarrow$ & PSNR $\uparrow$ & SSIM $\uparrow$ & LPIPS  $\downarrow$ \\
\midrule
Regularize = 1e-3 & 26.815 & 14.5 & 8.87 & 0.62 & 0.60\\
Regularize = 1e-2 & 74.586 & 350.301 & 9.37 & 0.71 & 0.55 \\
Regularize = 1e-1 & 115..65 & 581.81 & 4.51 & 0.39 & 0.73 \\
Regularize = 1 & 94.81 & 284.61 & 9.85 & 0.70 & 0.56 \\
\midrule
knots = 75 & 50.779 & 199.2 & 8.46 & 0.61 & 0.60 \\
knots = 50 & 3.009 & 9.523 & 14.46 & 0.69 & 0.21 \\
knots = 25 & 3.01 & 9.53 & 14.46 & 0.69 & 0.21 \\

\bottomrule
\end{tabular}}
\vspace{1mm}
\caption{\textbf{Quantitative results of BARF with B-Spline.} We use scipy B-spline interpolation implementation splrep. On top part we use knots = 25 and for bottom part we use s = 1e-3.}
\label{tab:b_spline_3d}
\end{table}

\begin{figure*}[t]
    \centering
    \begin{tabular}{cccc}
    \centering
     & corrected pose & RGB & Depth \\
    \multirow{2}{*}{\includegraphics[width=4cm, trim={0  0  0  1.4cm},clip]{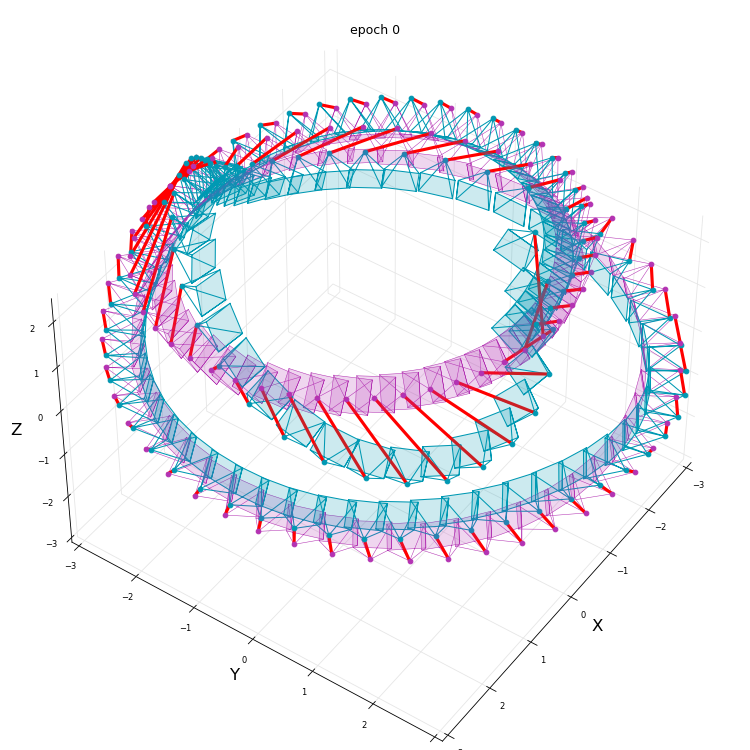}} & \includegraphics[width=3cm,trim={0  0  0  1.4cm},clip]{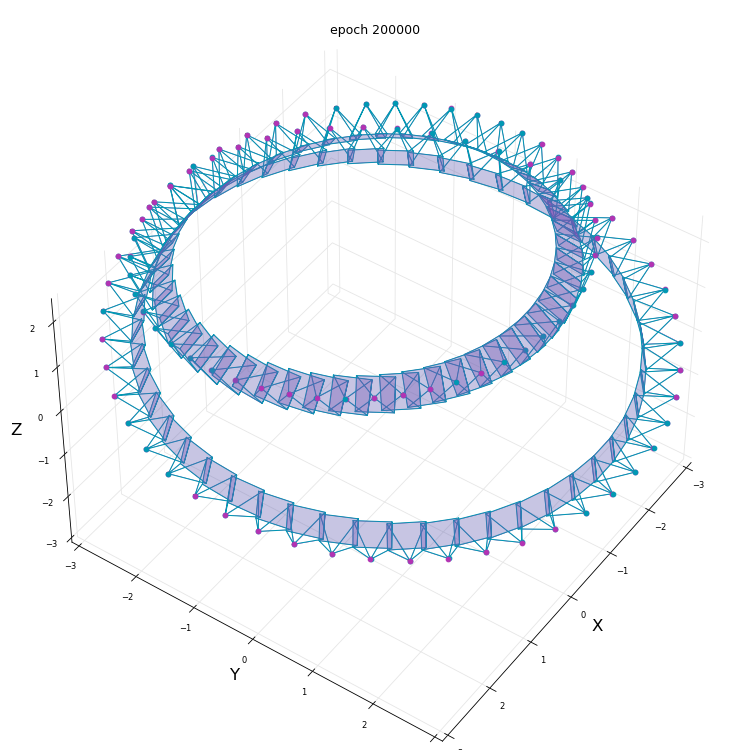} & \includegraphics[width=3cm]{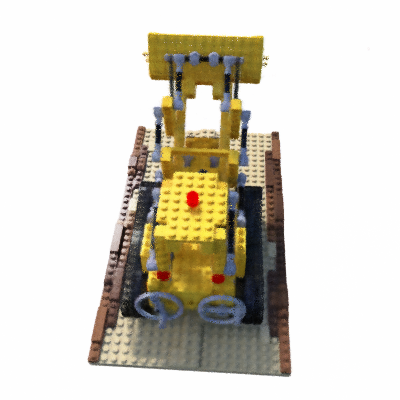} & \includegraphics[width=3cm]{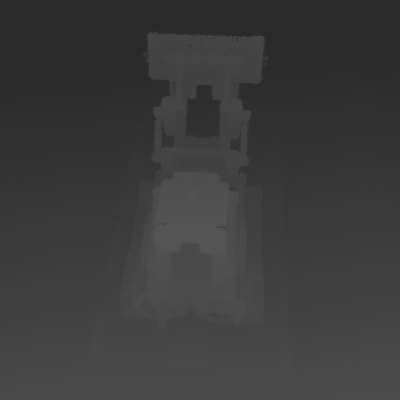} \\

    & \multicolumn{3}{c}{BARF} \\
    
    & \includegraphics[width=3cm,trim={0  0  0  1.4cm},clip]{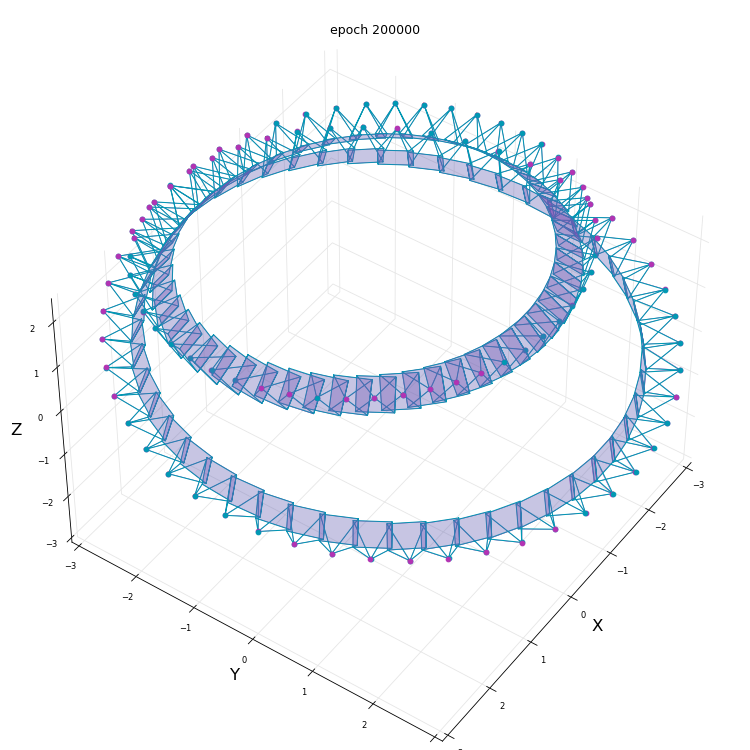} & \includegraphics[width=3cm]{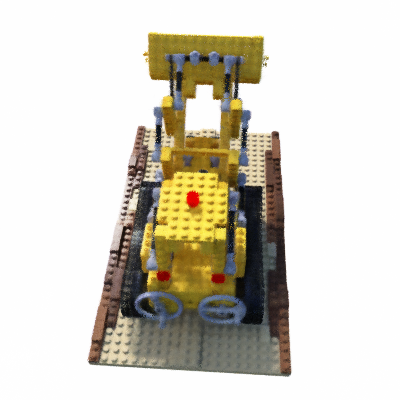} & \includegraphics[width=3cm]{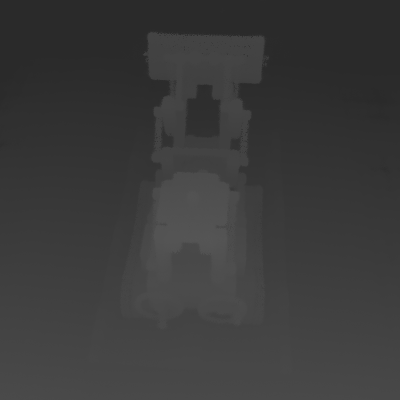} \\

    Scale 01 pose perturbation & \multicolumn{3}{c}{Ours} \\

  \hline
  \hline

    \multirow{2}{*}{\includegraphics[width=4cm, trim={0  0  0  1.4cm},clip]{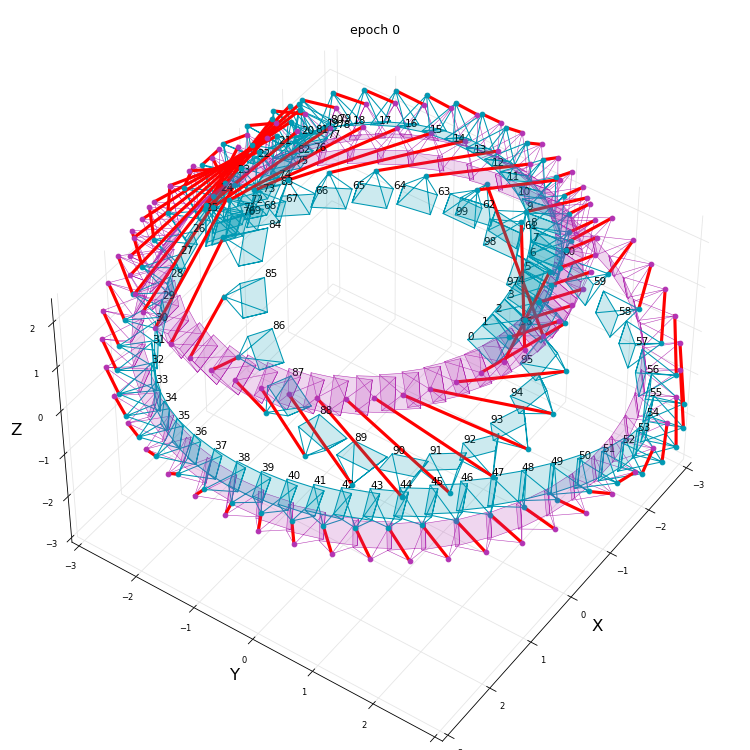}} & \includegraphics[width=3cm,trim={0  0  0  1.4cm},clip]{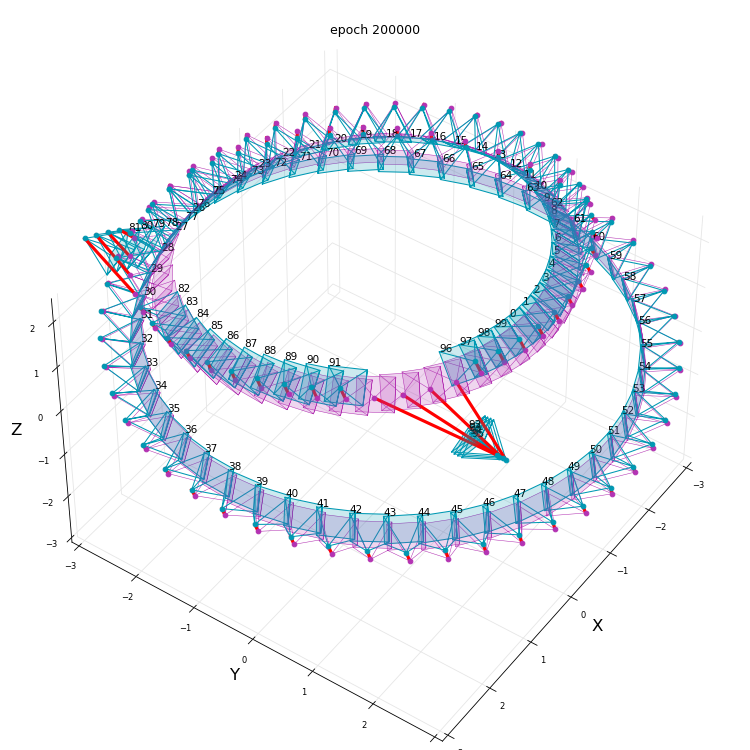} & \includegraphics[width=3cm]{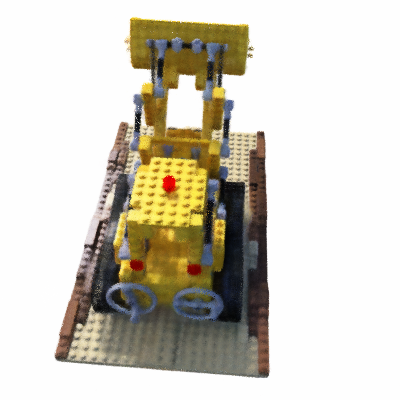} & \includegraphics[width=3cm]{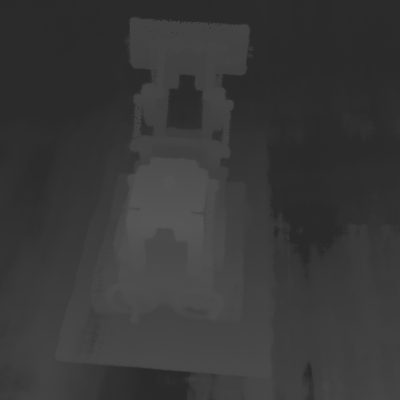} \\

    & \multicolumn{3}{c}{BaRF} \\
    
    & \includegraphics[width=3cm,trim={0  0  0  1.4cm},clip]{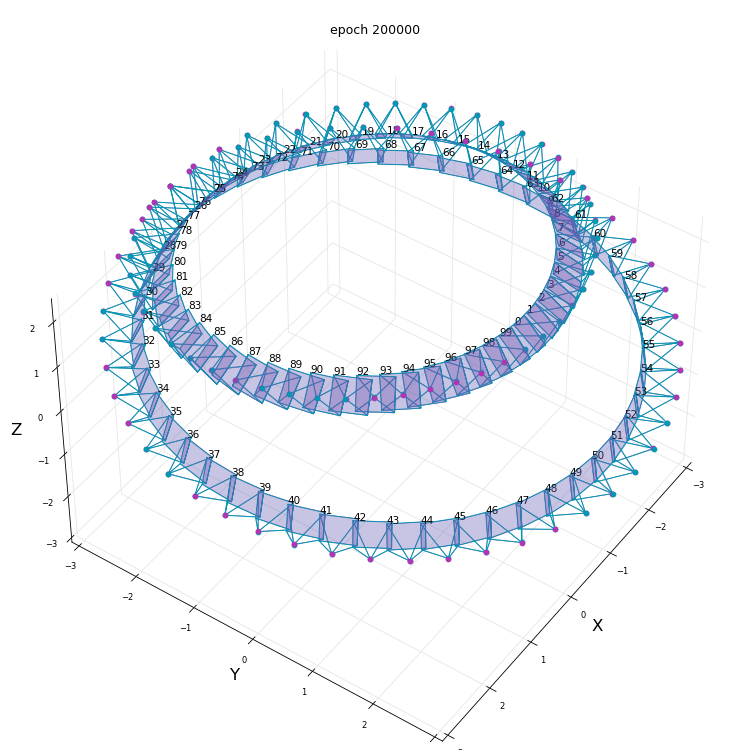} & \includegraphics[width=3cm]{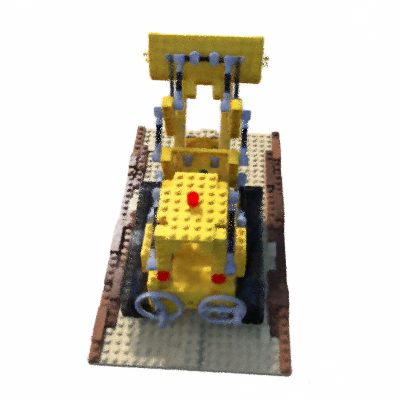} & \includegraphics[width=3cm]{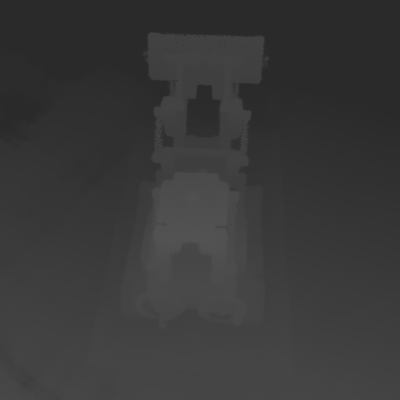} \\

    Scale 02 pose perturbation & \multicolumn{3}{c}{Ours} \\

  \hline
  \hline

    \multirow{2}{*}{\includegraphics[width=4cm, trim={0  0  0  1.4cm},clip]{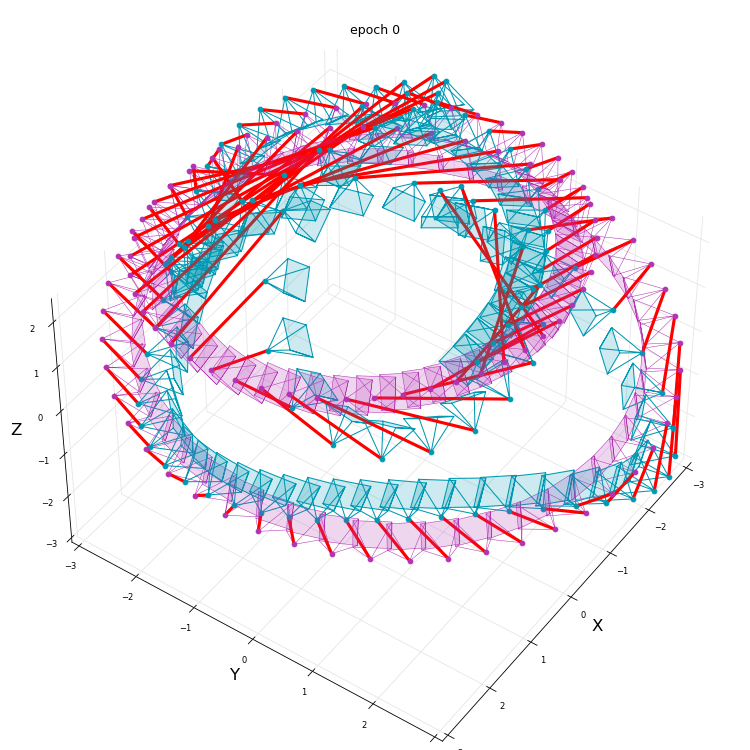}} & \includegraphics[width=3cm,trim={0  0  0  1.4cm},clip]{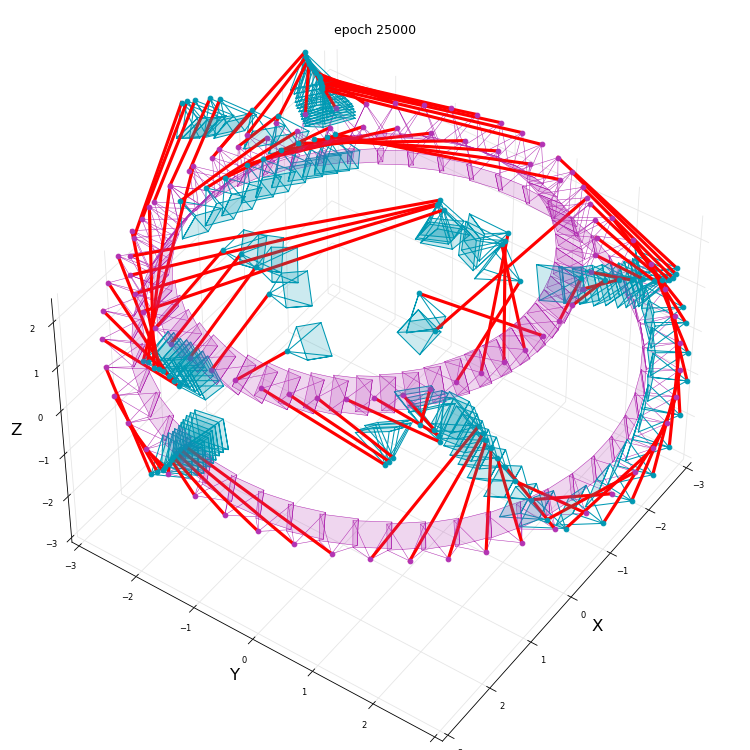} & \includegraphics[width=3cm]{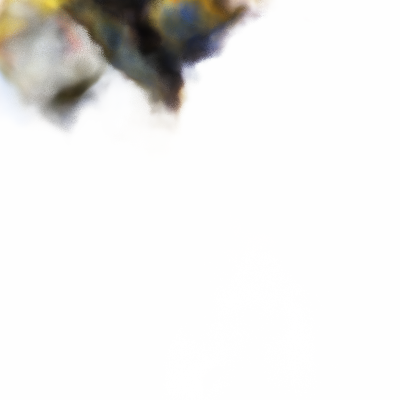} & \includegraphics[width=3cm]{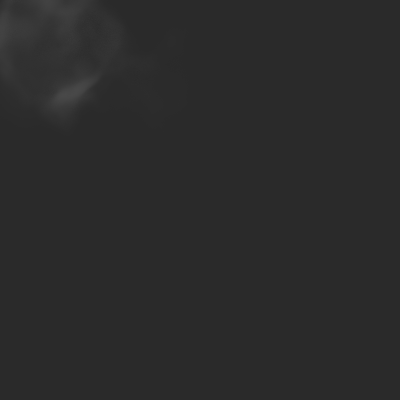} \\

    & \multicolumn{3}{c}{BaRF} \\
    
    & \includegraphics[width=3cm,trim={0  0  0  1.4cm},clip]{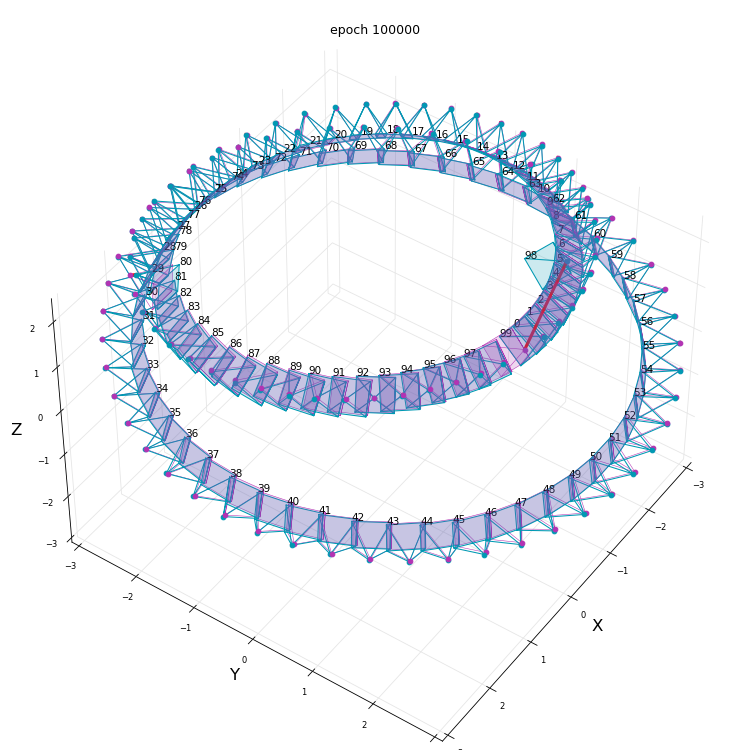} & \includegraphics[width=3cm]{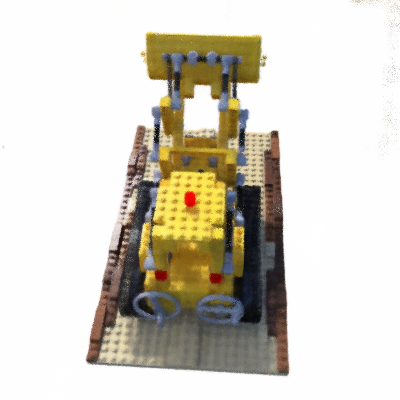} & \includegraphics[width=3cm]{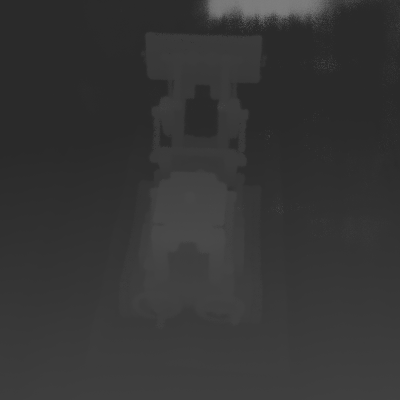} \\

    Scale 03 pose perturbation & \multicolumn{3}{c}{Ours} \\
  
    \end{tabular}
    \vspace{1mm}

\caption{\textbf{Qualitative results of interpolated pose noise.}  Our method can handle large pose noise and render images in the centre with the correct detphs.
\label{fig:synthetic_noise_pertubation} }

\end{figure*}

\vspace{-1em}
\paragraph{More results on the synthetic dataset.} 
From Table~\ref{tab:barf_noise_interpolation} we can see our method is robust to large translation noise of up to 40\% of the whole scene and is also robust to large rotation deviations of up to 90 degrees. BARF fails to register the camera frame under 20\% translation and 60-degree rotation perturbation and although the 3D object is correctly reconstructed with largely correct poses, certain novel view synthesis yields bad PSNR as the object deviates from the image centre. This can be clearly seen in qualitative results comparison in Figure \ref{fig:synthetic_noise_pertubation}.

\begin{table}[htbp]
\centering

\begin{subtable}[t]{0.5\textwidth}
\centering
\resizebox{0.75\linewidth}{!}{
\begin{tabular}{lccccc}
\toprule

TM & \multicolumn{5}{c}{80 (10\%) } \\ 
\midrule 
Method & RE & TE & PSNR & SSIM & LPIPS  \\
\midrule
BaRF\cite{lin2021barf} & 0.06 & 0.254 & 27.72 & 0.92 & 0.04 \\
Ours & 0.03  & 0.196 & 27.91 & 0.92 & 0.04 \\
\midrule
TM  & \multicolumn{5}{c}{160 (20\%) } \\ 
\midrule 
Method & RE & TE & PSNR & SSIM & LPIPS  \\
\midrule
BaRF\cite{lin2021barf} & 24.76 & 57.342 & 9.79 & 0.61 & 0.52 \\
Ours & 0.05  & 0.292 & 26.74 & 0.91 & 0.06 \\
\midrule
TM  & \multicolumn{5}{c}{240 (30\%) } \\ 
\midrule 
Method & RE & TE & PSNR & SSIM & LPIPS  \\
\midrule
BaRF\cite{lin2021barf} & 19.77 & 95.631 & 6.97 & 0.50 & 0.73 \\
Ours & 0.03 & 0.178 & 28.44 & 0.93 & 0.04  \\
\midrule
TM  & \multicolumn{5}{c}{320 (40\%) } \\ 
\midrule 
Method & RE & TE & PSNR & SSIM & LPIPS  \\
\midrule
BaRF\cite{lin2021barf} & 18.66 & 127.3 & 7.39 & 0.53 & 0.71 \\
Ours & 0.03  & 0.200 & 28.25 & 0.93 & 0.04 \\

\bottomrule
\end{tabular}}
\caption{\textbf{Interpolated translational noise experiments.}}
\end{subtable}
\vspace{1mm}

\begin{subtable}[t]{0.5\textwidth}
\centering
\resizebox{0.75\linewidth}{!}{
\begin{tabular}{lccccc}
\toprule

RM & \multicolumn{5}{c}{30 \textdegree{}} \\ 
\midrule 
Method & RE & TE & PSNR & SSIM & LPIPS  \\
\midrule
BaRF\cite{lin2021barf} & 0.067 & 0.265 & 27.75 & 0.92 & 0.05 \\
Ours & 0.049  & 0.105 & 28.22 & 0.93 & 0.04 \\
\midrule
RM & \multicolumn{5}{c}{60 \textdegree{}} \\ 
\midrule 
Method & RE & TE & PSNR & SSIM & LPIPS  \\
\midrule
BaRF\cite{lin2021barf} & 0.101 & 0.378 & 26.82 & 0.91 & 0.06 \\
Ours & 0.050  & 0.141 & 28.13 & 0.93 & 0.04 \\
\midrule
RM & \multicolumn{5}{c}{90 \textdegree{}} \\ 
\midrule 
Method & RE & TE & PSNR & SSIM & LPIPS  \\
\midrule
BaRF\cite{lin2021barf} & 12.103 & 37.380 & 10.40 & 0.61 & 0.42 \\
Ours & 0.061 & 0.181 & 28.03 & 0.93 & 0.04  \\
\midrule
RM & \multicolumn{5}{c}{120 \textdegree{}} \\ 
\midrule 
Method & RE & TE & PSNR & SSIM & LPIPS  \\
\midrule
BaRF\cite{lin2021barf} & 40.526 & 122.454 & 6.62 & 0.54 & 0.66 \\
Ours  & 19.279 & 66.572 & 8.79 & 0.56 & 0.52 \\

\bottomrule
\end{tabular}}
\caption{\textbf{Interpolated rotational noise experiments.}}
\label{tab:barf_rotation_modulate}
\end{subtable}
\vspace{1mm}

\begin{subtable}[t]{0.5\textwidth}
\centering
\resizebox{0.75\linewidth}{!}{
\begin{tabular}{lccccc}
\toprule

TM+RM $R|t$ & \multicolumn{5}{c}{30 \textdegree{} + 80(10\%) } \\ 
\midrule 
Method & RE & TE & PSNR & SSIM & LPIPS  \\
\midrule
BaRF\cite{lin2021barf} & 0.062  & 0.306 & 27.78 & 0.92 & 0.04 \\
Ours & 0.064 & 0.266  & 28.97  & 0.93 & 0.04  \\
\midrule
TM+RM $R|t$ & \multicolumn{5}{c}{60 \textdegree{} + 160(20\%) } \\ 
\midrule 
Method & RE & TE & PSNR & SSIM & LPIPS  \\
\midrule
BaRF\cite{lin2021barf} & 5.835  & 29.560 & 11.63 & 0.63 & 0.35  \\
Ours & 0.077 & 0.293 & 26.64 & 0.91 & 0.06 \\
\midrule
TM+RM $R|t$ & \multicolumn{5}{c}{90 \textdegree{} + 240(30\%)} \\ 
\midrule 
Method & RE & TE & PSNR & SSIM & LPIPS  \\
\midrule
BaRF\cite{lin2021barf} & 46.352 & 160.639 & 8.17 & 0.63 & 0.60  \\
Ours & 0.378 & 2.813 & 22.10 & 0.83 & 0.09  \\
\midrule
TM+RM $R|t$t & \multicolumn{5}{c}{120 \textdegree{} + 320(40\%) } \\ 
\midrule 
Method & RE & TE & PSNR & SSIM & LPIPS  \\
\midrule
BaRF\cite{lin2021barf} & 55.640 & 195.134 & 7.7 & 0.63 & 0.63  \\
Ours & 16.122 & 52.527 & 9.16 & 0.56 & 0.51  \\

\bottomrule
\end{tabular}}
\vspace{1mm}
\caption{\textbf{Interpolated translational and rotation noise experiments.}}
\label{tab:barf_scale_modulate}

\end{subtable}
\vspace{1mm}
\caption{\textbf{Interpolated pose noise experiments.} TM refers to Translational maximal deviation and RM refers to Rotational maximal deviation. The diameter of the circular trajectory is 800, the maximal deviation of the translation perturbation is set to be 10\%, 20\%, 30\%, and 40\%.}
\label{tab:barf_noise_interpolation}
\end{table}

\begin{figure*}[h]
    \begin{tabular}{ccc|cc}
     & \multicolumn{2}{c}{RGB} & \multicolumn{2}{c}{Depth} \\
     & ours & BaRF & ours & BaRF  \\

    Trex (50)& \includegraphics[width=3cm]{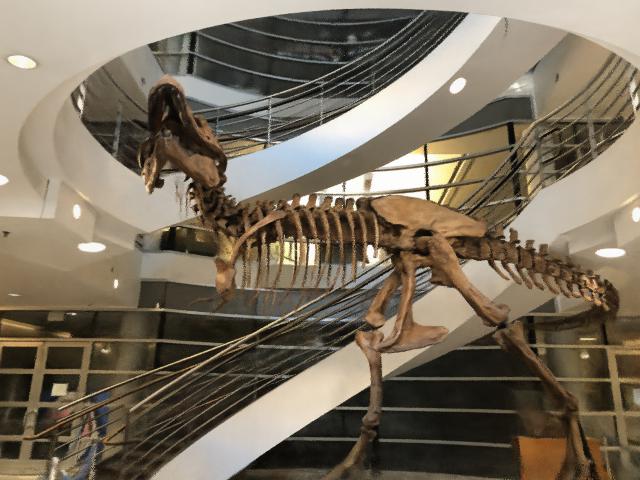} & \includegraphics[width=3cm]{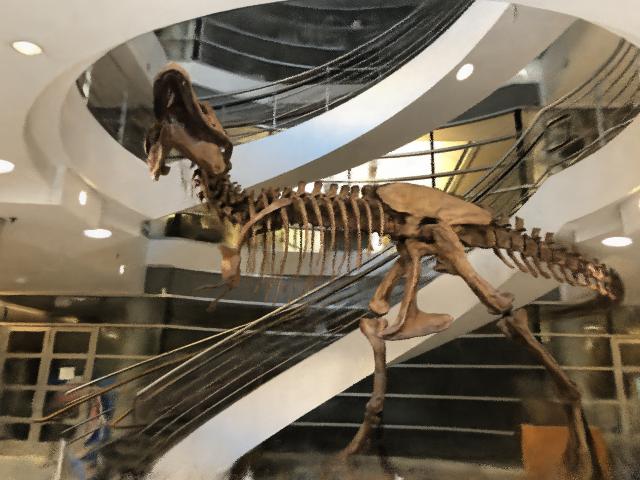}
    & \includegraphics[width=3cm]{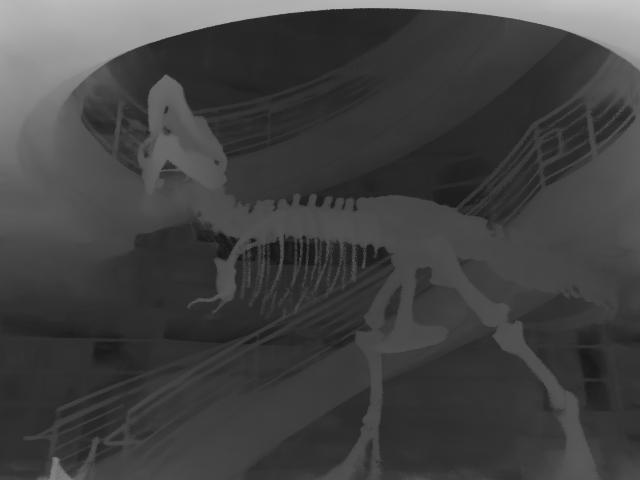} & \includegraphics[width=3cm]{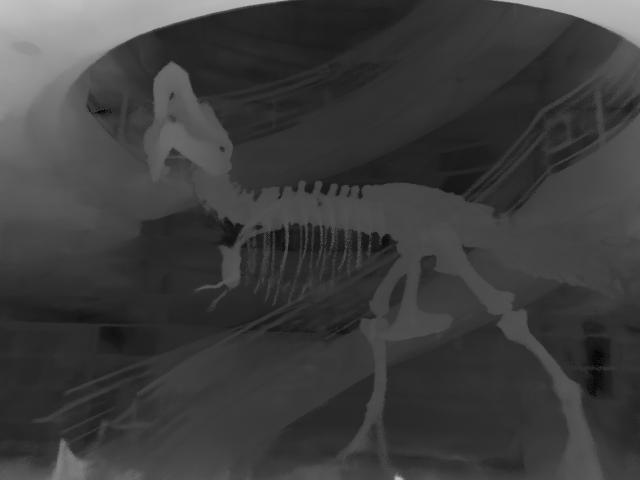}
    \\
    Trex/2 (25) & \includegraphics[width=3cm]{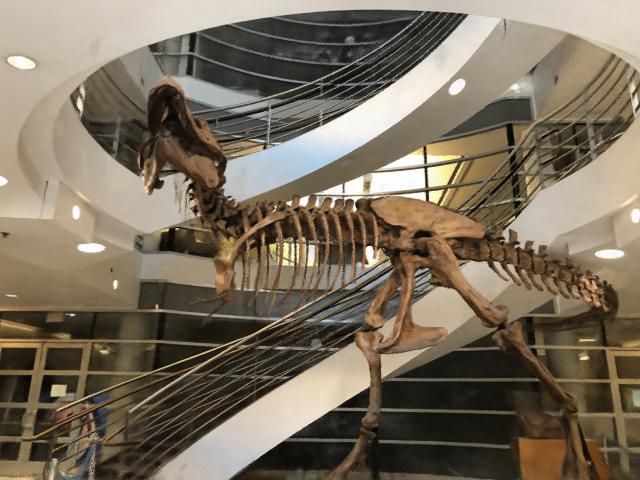} & \includegraphics[width=3cm]{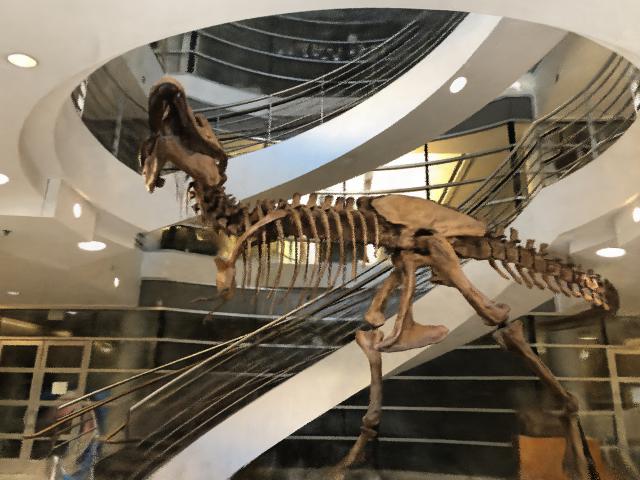}
    & \includegraphics[width=3cm]{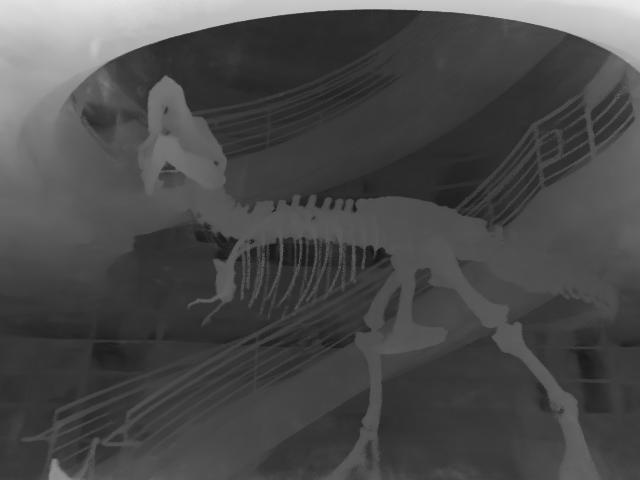} & \includegraphics[width=3cm]{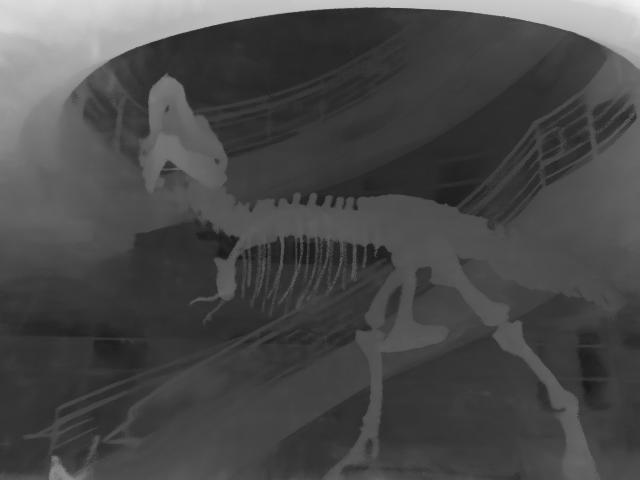}
    \\
    Trex/4 (13) & \includegraphics[width=3cm]{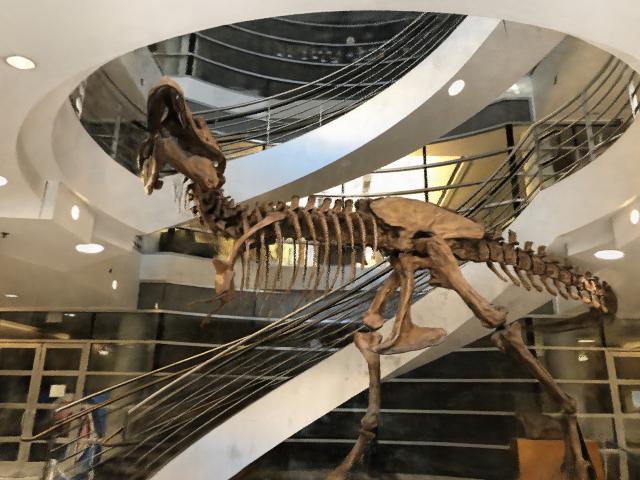} & \includegraphics[width=3cm]{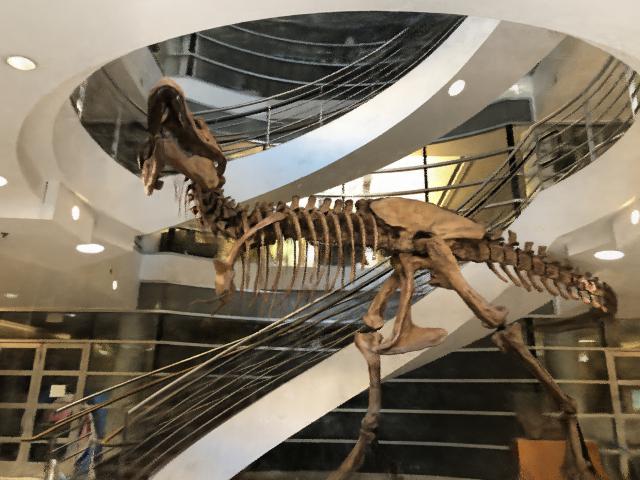}
    & \includegraphics[width=3cm]{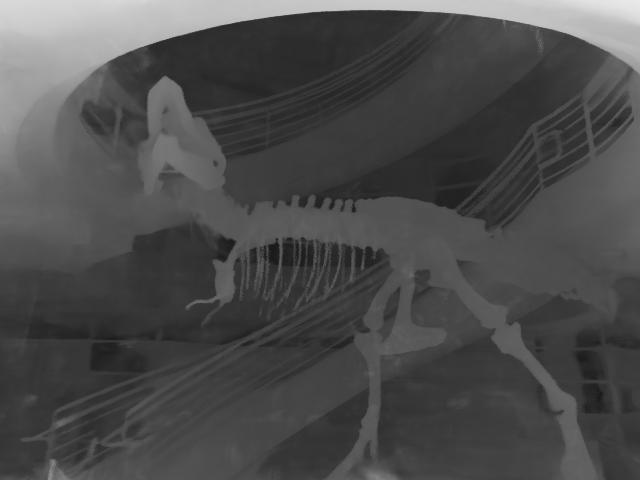} & \includegraphics[width=3cm]{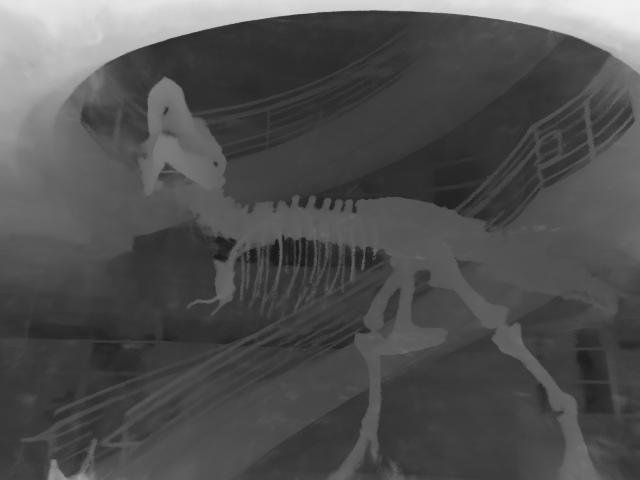}
    \\
    
    \end{tabular}
    \vspace{1mm}
    
\caption{\textbf{Qualitative results of novel view synthesis real datasets~\cite{mildenhall2019llff} with unknown pose.} Corresponding to Table 2 in the main text we report the real dataset results in which the camera moves in a smooth trajectory. We denote BaRF failure cases with * and the number in the parentheses is the number of frames used in training. }
\label{fig:Real_dataset_qualtiative} 
\end{figure*}

\vspace{-1em}
\paragraph{More results of real dataset.} More results on other real scenes as well as qualitative results can be found in Table ~\ref{tab:more_llff} and Figure ~\ref{fig:Real_dataset_qualtiative}. Benefiting from neighboring temporal information our proposed pose representation performs consistently well on different speeds of camera motion. Similar to the above experiments we can find the novel view deviates from the image center in Fort/2(19) results. Furthermore, our method is robust to high-speed scenarios with slight artifacts while BARF diverges and provides very inaccurate results. Note that in the reported results we disable the test-time photometric optimization for better comparison of camera pose registration performance.

\vspace{-1em}
\paragraph{B-spline baseline experiments on 3D.} Similar to 2D planar experiments we report also the results using classical continuous B-spline to enforce continuity between neighbouring poses. We experimented with various parameter configurations to illustrate the challenge of tuning classical methods in the context of neural radiance fields.
\vspace{-1em}
\paragraph{Ablation on network size.} We report the performance evaluations with different network sizes. The reduction in network size affects camera localization performance. We use the 8-layer and 256 width model for other applications.

\begin{table}[t]
\resizebox{\linewidth}{!}{
\begin{tabular}{lccccc}
\midrule 
Method & Rotation error $\downarrow$ & Translation error $\downarrow$  & PSNR $\uparrow$ & SSIM $\uparrow$ & LPIPS $\downarrow$  \\

\midrule

8-layer, width 256 &  \textbf{0.07} &  \textbf{0.28} & \textbf{27.30}  & \textbf{0.92}  & \textbf{0.06}\\
8-layer, width 128 &  0.09 & 0.31  &  27.33 & 0.90  & 0.09 \\
4-layer, width 256 &  0.10 & 0.32  & 27.13  & 0.90  & 0.10 \\
4-layer, width 128 &  0.11 & 0.33  & 27.15  & 0.91  & 0.11 \\

\bottomrule
\end{tabular}}
\vspace{1mm}
\caption{\textbf{Ablation study on network sizes}. The performance of camera localization drops only slightly with decreased network size. The experiments is conducted using our synthetic dataset, consistent with Table \ref{tab:barf_encoder}.}
\label{tab:network_size}
\end{table}

\section{Continuous Pose for Asynchronous Events}
\label{sec:supp_events}
\paragraph{Implementation details.} \cite{rudnev2022eventnerf} shows EventNeRF reconstruction quality cannot handle inaccurate camera poses over 1 \textdegree{}. The real sequences angle offset reported by EventNeRF however can reach up to 2.85 \textdegree{}. Following its noise perturbation method, we introduce different magnitudes of pose inaccuracies in the real datasets. Furthermore, we also consider the pose inaccuracies due to unknown asynchronous event poses. \cite{rudnev2022eventnerf} provides in total 1000 ground truth poses from Blender which describes a circumferential movement. We uniformly sample different numbers of poses to linearly interpolate the whole circular path position and keep the orientation unchanged. Similarly to above, we use the Adam optimizer with an exponential learning rate schedule which decays from 2e-4 to 2e-6 for TransNet and 5e-5 to 5e-7 for RotsNet.

\begin{figure*}
  \centering
  \resizebox{\linewidth}{!}{
  \begin{tabular}{ccccc}
     & Interpolate 20 & Interpolate10 & Interpolate6  \\
    \multirow{2}{*}{\includegraphics[width=4cm]{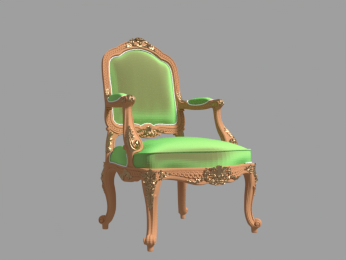}} & \includegraphics[width=3cm]{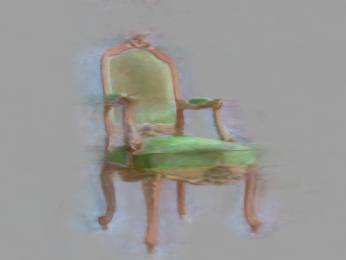} & \includegraphics[width=3cm]{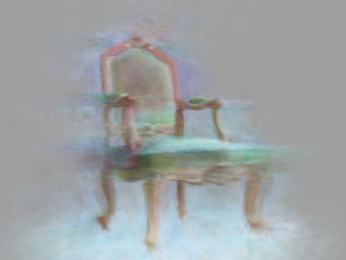} & \includegraphics[width=3cm]{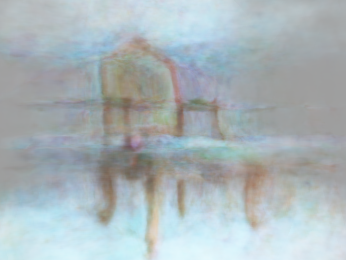} \\
    
     & \includegraphics[width=3cm, trim={0  0  2cm 0 },clip]{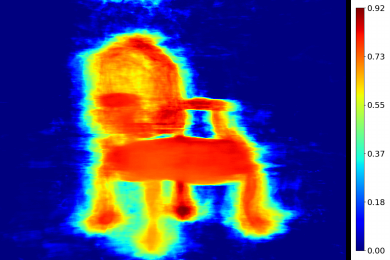} & \includegraphics[width=3cm, trim={0  0  2cm 0 },clip]{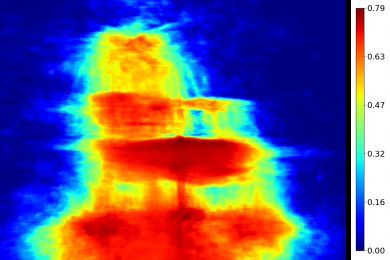} & \includegraphics[width=3cm, trim={0  0  2cm 0 },clip]{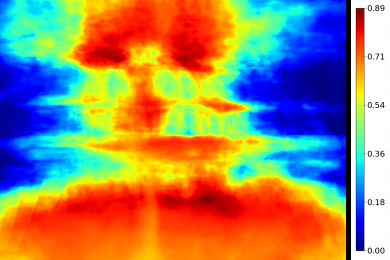} \\

    & \multicolumn{3}{c}{EventNeRF} \\

     & \includegraphics[width=3cm]{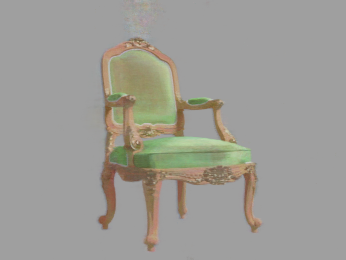} & \includegraphics[width=3cm]{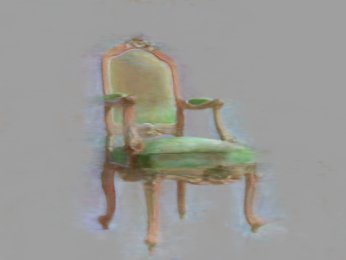} & \includegraphics[width=3cm]{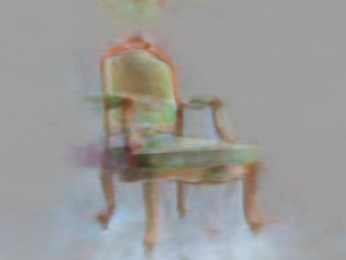} \\

     & \includegraphics[width=3cm, trim={0  0  2cm 0 },clip]{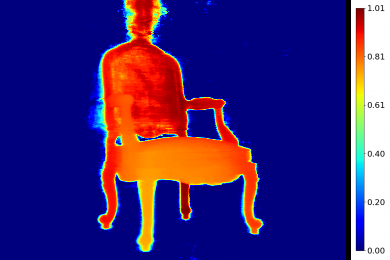} & \includegraphics[width=3cm, trim={0  0  2cm 0 },clip]{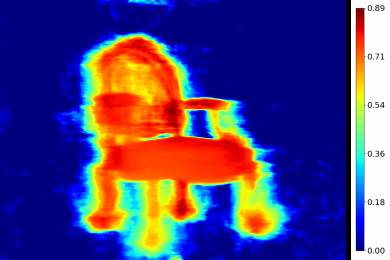} & \includegraphics[width=3cm, trim={0  0  2cm 0 },clip]{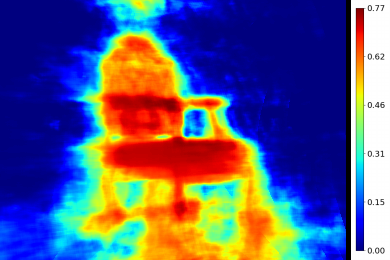} \\
    
    Chair Dataset ground truth  & \multicolumn{3}{c}{Ours} \\
    \hline 
    \hline 
  \end{tabular}}


    \resizebox{\linewidth}{!}{
  \begin{tabular}{ccccc}
    \multirow{2}{*}{\includegraphics[width=4cm]{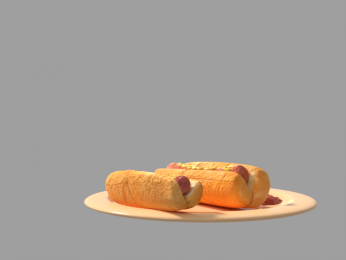}} & \includegraphics[width=3cm]{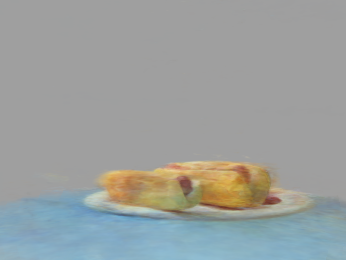} & \includegraphics[width=3cm]{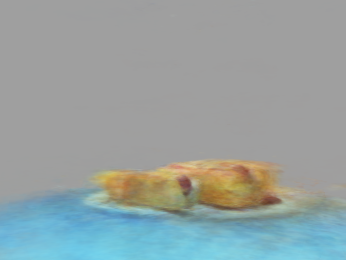} & \includegraphics[width=3cm]{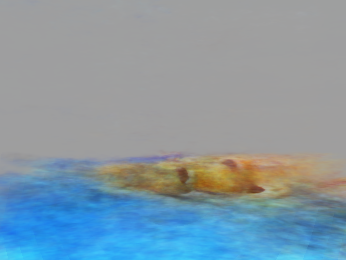} \\

     & \includegraphics[width=3cm, trim={0  0  2cm 0 },clip]{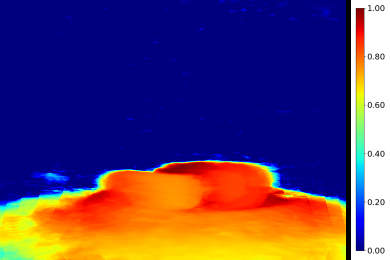} & \includegraphics[width=3cm, trim={0  0  2cm 0 },clip]{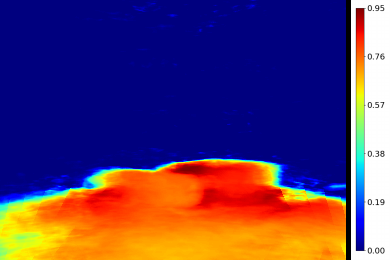} & \includegraphics[width=3cm, trim={0  0  2cm 0 },clip]{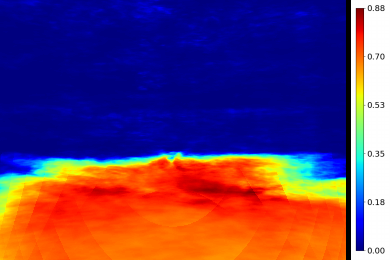} \\
    
    & \multicolumn{3}{c}{EventNeRF} \\

     & \includegraphics[width=3cm]{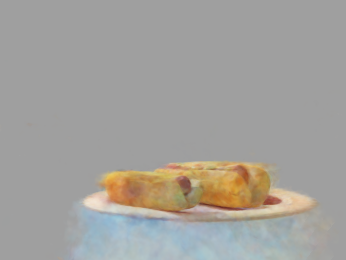} & \includegraphics[width=3cm]{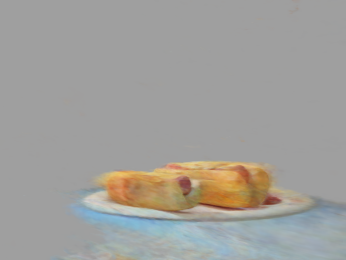} & \includegraphics[width=3cm]{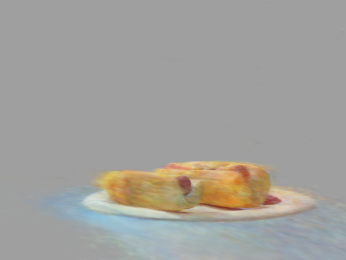} \\

     & \includegraphics[width=3cm, trim={0  0  2cm 0 },clip]{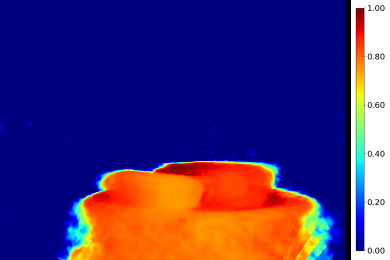} & \includegraphics[width=3cm, trim={0  0  2cm 0 },clip]{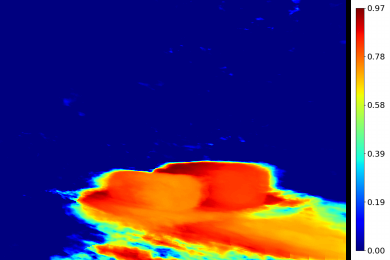} & \includegraphics[width=3cm, trim={0  0  2cm 0 },clip]{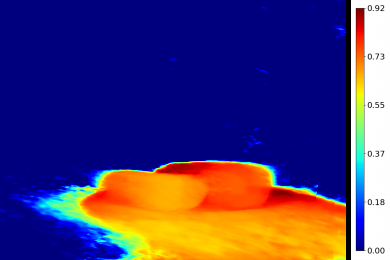} \\
     
    Hotdog Dataset ground truth & \multicolumn{3}{c}{Ours} \\
  \end{tabular}}


\caption{\textbf{Qualitative results of novel view depth and rgb synthesis Interpolation error experiments.}  Our method improves EventNeRF significantly in all six experimental setups.}
\label{fig:chair_and_hotdog_rgb} 

\end{figure*}

\vspace{-1em}
\paragraph{Qualitative results in interpolation error experiments.} 
In Figure~\ref{fig:chair_and_hotdog_rgb} we report the qualitative results of novel view synthesis on synthetic sequences of chair and hotdog which correspond to Table 4 of the main text. We can see EventNeRF suffers from strong fuzzy artifacts and the depth seems to dilate around the object while our method correctly learns the depth and reconstructs clearer 3D objects.

\vspace{-1em}
\paragraph{Qualitative results on the synthetic datasets in angle offset calibration experiments.} 

\begin{figure*}
  \centering
  \resizebox{\linewidth}{!}{
  \begin{tabular}{ccccc}
     & offset 0.2388 \textdegree{} & offset 1.5 \textdegree{} & offset 2.85 \textdegree{} \\
    \multirow{2}{*}{\includegraphics[width=4cm,trim={0  1cm 2cm 0 },clip]{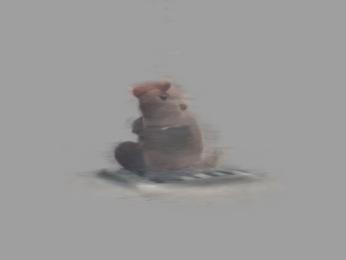}} & \includegraphics[width=3cm]{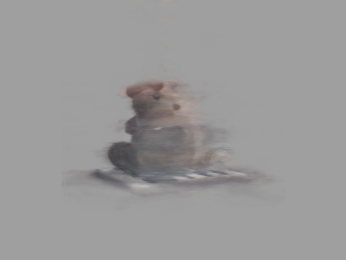} & \includegraphics[width=3cm]{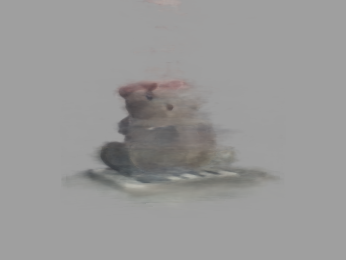} & \includegraphics[width=3cm]{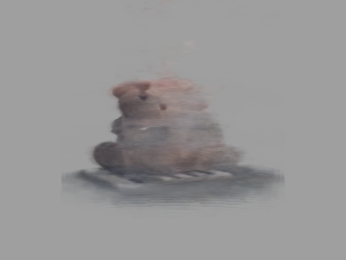}  \\
    & \multicolumn{3}{c}{EventNeRF} \\

     & \includegraphics[width=3cm]{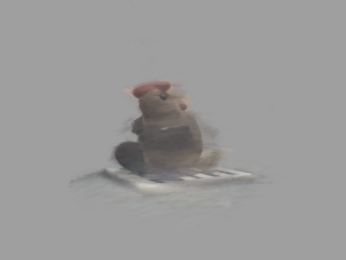} & \includegraphics[width=3cm]{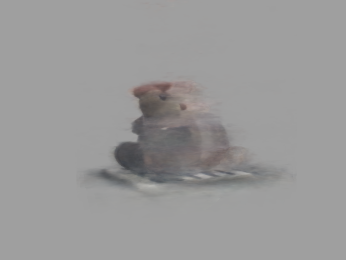} & \includegraphics[width=3cm]{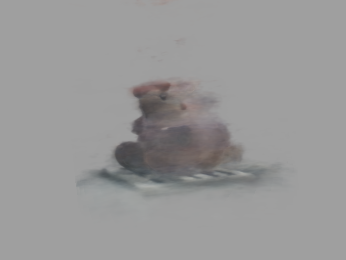}  \\
    
    Chick Dataset (with calibration)  & \multicolumn{3}{c}{Ours} \\

  \end{tabular}}

\caption{\textbf{Qualitative results of novel view depth and rgb synthesis in angle offset calibration experiments.}  Our method improves EventNeRF significantly in all six experimental setups.}
\label{fig:multi_chick} 

\end{figure*}

In Figure~\ref{fig:multi_chick} we report 3 more real dataset experiments on sequences of multimeter and plant. Similar to Figure~\ref{fig:calibration} in the main text, EventNeRF suffers from trailing artifacts and at large angle offsets it nearly reconstructs 2 separate objects around the image center. In contrast, our method learns the offset angle and re-positions the object back to the image center.

\section{Visual SLAM with Depth and IMUs}
\label{sec:supp_events}

\paragraph{Full IMU fusion.} In the main text, we elaborate on harnessing gyroscope readings through both loose and tight coupling methods. However, direct utilization of accelerometer readings poses challenges as it provides acceleration instead of velocity in the body frame, resulting in a significant error when integrating with an unknown initial speed. Additionally, effective processing of acceleration data necessitates critical steps such as gravity removal and denoising ~\cite{campos2021orb, qin2017vins,mourikis2007multi,forster2016svo}. Therefore we first show the experiment with simulated IMU on the modified ScanNet dataset with simulated IMU as shown in Table~\ref{tab:nice-slam-syn-IMU}. Given accelerator reading on time $t$, $\hat{\alpha}_t = (\hat{\alpha}_x, \hat{\alpha}_y, \hat{\alpha}_z)$. We first transform the reading to the last body frame with captured image, $\hat{\alpha}_{t_{i-1}, t} = R_{t_{i-1},t} \circ \hat{\alpha}_t $. We calculate $R_{t_i,t} $ from loosely coupled method mentioned above.
We then use auto-differentiation to calculate the second derivative of TransNet with respect to input time and supervise it with $\mathcal{L}1$ loss: 

\begin{equation}
\mathcal{L}_{acc} = | \ddot{f}({\theta_p}, t)  - \hat{\alpha}_{t_{i-1}, t}| ;
\label{eq:accelerator}
\end{equation}
\vspace{-1em}
\paragraph{Implementation details.} 

\begin{figure*}
    \centering
    
    \begin{subfigure}{0.5\textwidth}
        \centering
        \includegraphics[width=0.9\linewidth, trim={0  0  0  1.4cm},clip]{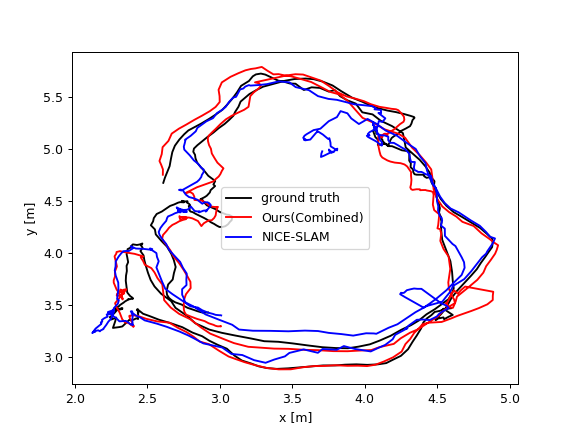}
        \caption{scan/059}
        \label{fig:059_quality}
    \end{subfigure}%
    \hfill
    \begin{subfigure}{0.5\textwidth}
        \centering
        \includegraphics[width=0.9\linewidth, trim={0  0  0  1.4cm},clip]{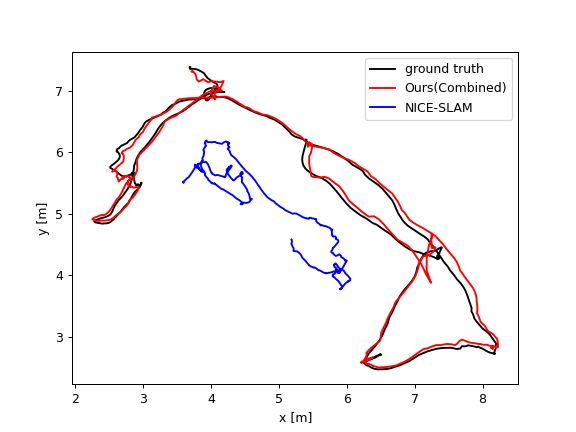}
        \caption{scan/v106}
        \label{fig:106_quality}
    \end{subfigure}%
    \hfill
    \begin{subfigure}{0.5\textwidth}
        \centering
        \includegraphics[width=0.9\linewidth, trim={0  0  0  1.4cm},clip]{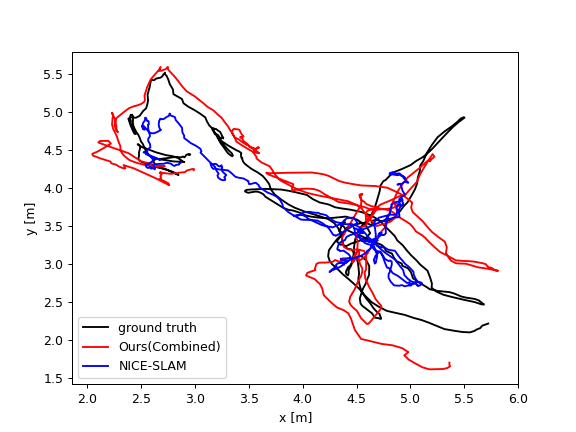}
        \caption{scan/181}
        \label{fig:181_quality}
    \end{subfigure}%
    \hfill
    \begin{subfigure}{0.5\textwidth}
        \centering
        \includegraphics[width=0.9\linewidth, trim={0  0  0  1.4cm},clip]{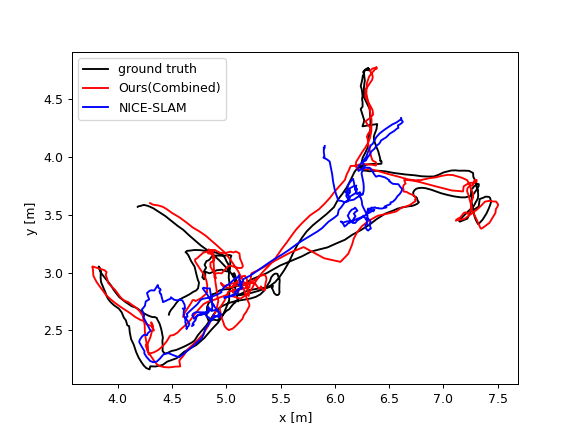}
        \caption{scan/207}
        \label{fig:207_quality}
    \end{subfigure}%

    \caption{\textbf{Qualitative results of tracking on challenging ScanNet} With the assistance of simulated IMU information, our method maintains robust tracking and preserves scale accuracy.}
    \label{fig:SLAM_IMU_quality}
\end{figure*}

In the NICE-SLAM experiments, we follow the original work~\cite{zhu2022nice} and the bundle adjustment is disabled. Learning rate for TransNet is set to 1e-3 and for RotsNet 2e-4. For IMU experiment we use $\lambda_{gyro} = 1$ and $\lambda_{acc} = 1$. For IMU simulation we interpolate the ground truth from 20 Hz to 200 Hz and calculate the numerical derivatives. We use the cubic interpolator for translation and $SLERP$ for rotation. We downsample the dataset from 20Hz to 5 Hz to highlight the importance of using IMU which is 100 Hz.
\vspace{-1em}
\paragraph{More experiments on RGB-D SLAM with IMU.}
 As Table ~\ref{tab:nice-slam-syn-IMU} shows, by fusing the acceleration and angular velocity we improve NICE-SLAM significantly and can maintain tracking to the end on challenging ScanNet. Taking advantage of both temporal information yields the best tracking performance. Qualitative results can be seen in Figure~\ref{fig:SLAM_IMU_quality}. We then use our method on EUROC \cite{burri2016euroc}. We first use EKF-SLAM to denoise accelerator readings with sensor-fusion from gyroscope and Vicon Pose. As Table~\ref{tab:Acceleration} demonstrates, fusing accelerator is beneficial especially under challenging scenes such as v103 and v203, and combining both sensor data yields the best results on average.


\begin{table}[t]

\resizebox{\linewidth}{!}{
\centering
    \centering
        \begin{tabular}{lcccccc}
        \toprule        
         & \multicolumn{5}{c}{With IMU}   \\ 
        \midrule 
         & scan/059 & scan/106  & scan/181 & scan/207 & Average  \\ 
        \midrule 
        Nice-SLAM \cite{zhu2022nice} & 37.28 & 174.27  & 71.94  & 80.00  & 89.75   \\
        Ours(Gyro)  & 14.51 & 12.78 & 43.98   & 18.23 & 22.36  \\
        Ours(Acceleration)  & 14.98 & 11.49 &  44.13  & 19.38  & 22.49 \\
        Ours(Combined)  & \textbf{13.80} & \textbf{10.68} &  \textbf{38.20} & \textbf{14.80}  & \textbf{19.37}  \\
        \bottomrule
    \end{tabular}}

    \vspace{1mm}
    
    \caption{\textbf{Tracking performance on challenging Scannet \cite{dai2017scannet}.} Our PoseNet improves the tracking performance of NICE-SLAM significantly by fusing the IMU tightly. Using full IMU reading yields the best results over all experiment sequences.}
\label{tab:nice-slam-syn-IMU}

\end{table}

\begin{table}
\resizebox{\linewidth}{!}{
\centering
    \centering
    \vspace{1mm}

    \centering
        \begin{tabular}{lccccccc}
        \toprule
        Method & v101 & v102 & v103 & v201& v202 & v203  & Avg  \\ 
        \midrule
        No IMU & 2.17  & N/A  & 5.82 &  7.76  & 5.04 & N/A & N/A \\  
        Gyro & \textbf{1.98}  & 6.09  & 5.55 &  \textbf{4.99}  & \textbf{3.03} & 15.34 & 6.16  \\ 
        Accelerator  & 2.16 & 4.76  & 5.10 &  6.72  &  4.14 & 15.10 & 6.33  \\ 
        Combined  & 2.40 & \textbf{5.33}  &  \textbf{3.63}   & 5.84 & 3.46 & \textbf{13.63} & \textbf{5.71}   \\ 
        \bottomrule
        \end{tabular}}

    \caption{\textbf{Tracking performance on EUROC~\cite{burri2016euroc}.} Note that here we report only PoseNet based results. Utilizing both gyroscope and accelerometer data proves beneficial, particularly in challenging scenes, as compared to not using IMU.}
\label{tab:Acceleration}

\end{table}


\begin{table}[t]

\begin{subtable}[h]{0.5\textwidth}
\centering
\resizebox{0.7\linewidth}{!}{
\begin{tabular}{lccc}
\toprule

Method  & GFLOPs & Params$[×10^3]$  & Time-cost[s/it]  \\
\midrule
BARF & 65.60 & 514  & 0.133   \\
Ours & 65.62 & 791 &  0.138   \\
\bottomrule
\end{tabular}}

\end{subtable}
\begin{subtable}[h]{0.5\textwidth}
\centering    
\resizebox{1\linewidth}{!}{
\begin{tabular}{lccc}
\toprule

Method  & Tracking time-Cost[ms/iter] & Convergence rate[iter]  \\
\midrule
NICE-SLAM & 27.1 & 11.96 \\
Ours & 31.5 &  13.21   \\
\bottomrule
\end{tabular}}
\end{subtable}
\vspace{1mm}
\caption{
\textbf{Left--Computation \& Runtime.} Computation of a 1024 batch ray using RTX 3090, with the negligible inclusion of extra computation and time expense.
\textbf{Right--Runtime \& Convergence rate.} We follow the default setting of Replica.yaml. We assume convergence when the tracking loss remains unchanged.}
\label{tab:computation}

\end{table}

\clearpage

\begin{figure*}[t]

    \centering
    \begin{tabular}{ccc|cc}
     & \multicolumn{2}{c}{RGB} & \multicolumn{2}{c}{Depth} \\
     & ours & BaRF & ours & BaRF  \\
    Fern (18) & \includegraphics[width=3cm]{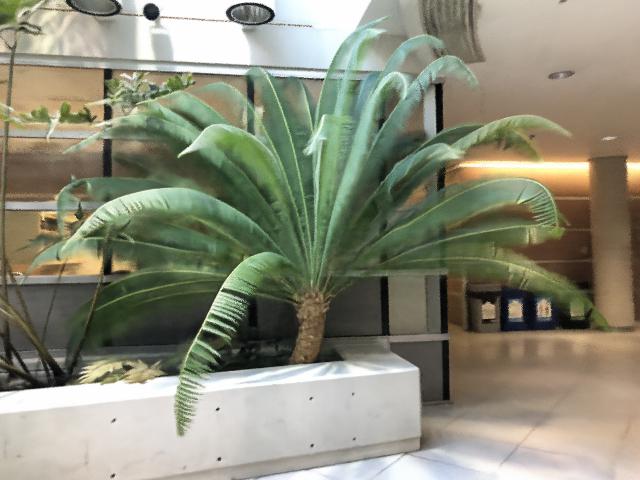} & \includegraphics[width=3cm]{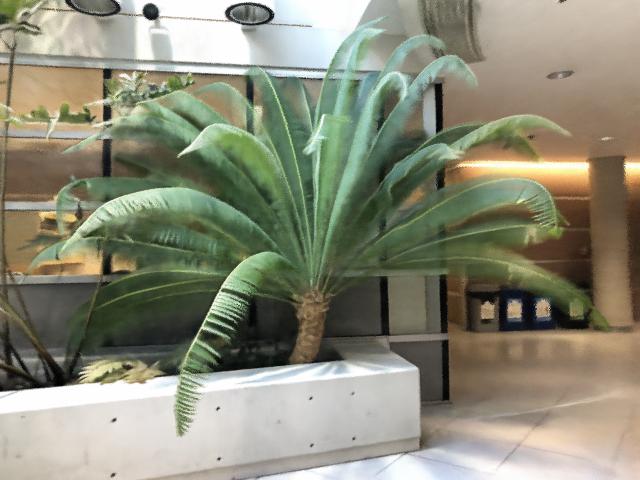}
    & \includegraphics[width=3cm]{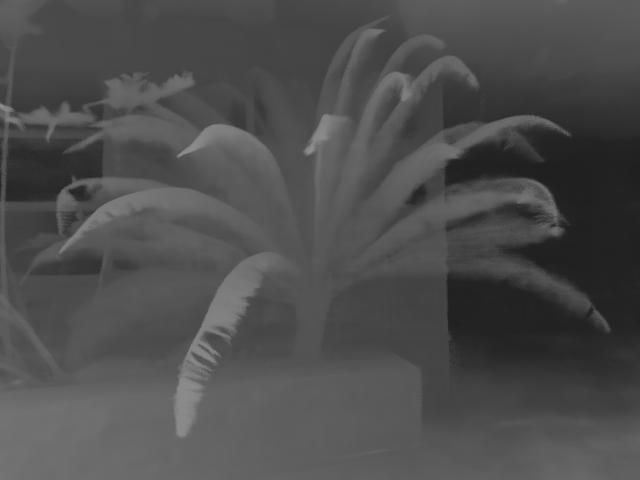} & \includegraphics[width=3cm]{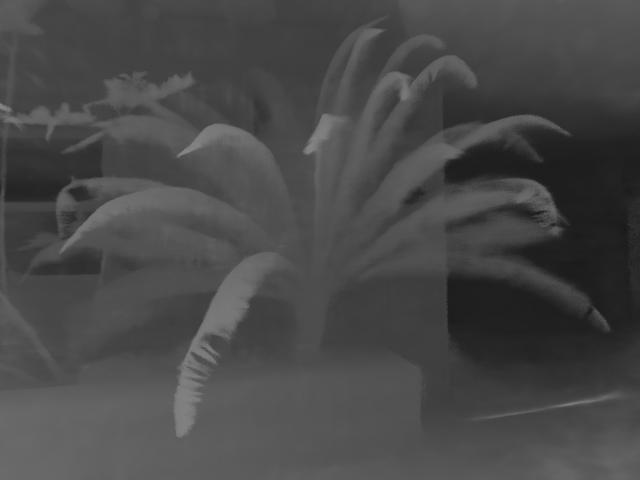}
    \\
    Fern/2 (9) & \includegraphics[width=3cm]{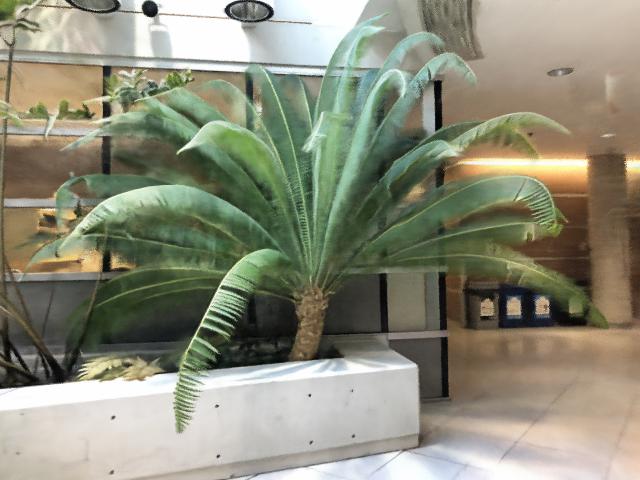} & \includegraphics[width=3cm]{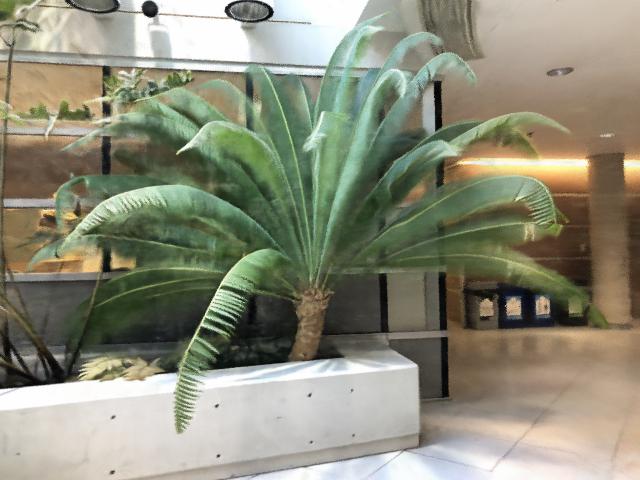}
    & \includegraphics[width=3cm]{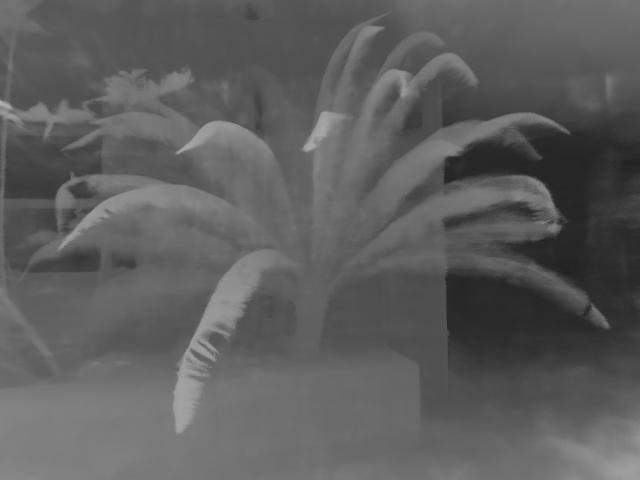} & \includegraphics[width=3cm]{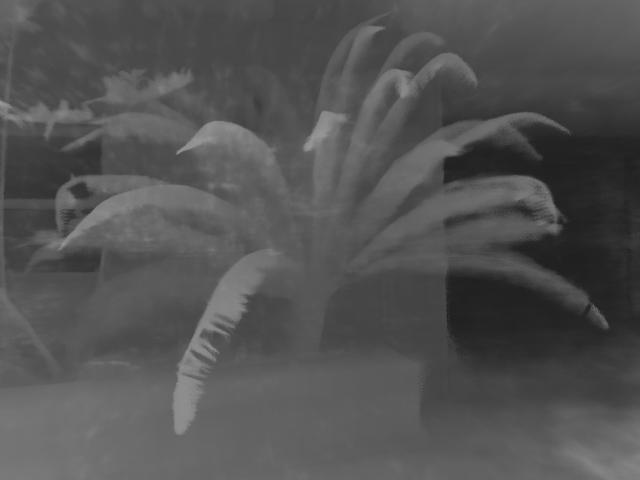}
    \\
    Fern/4 (5) & \includegraphics[width=3cm]{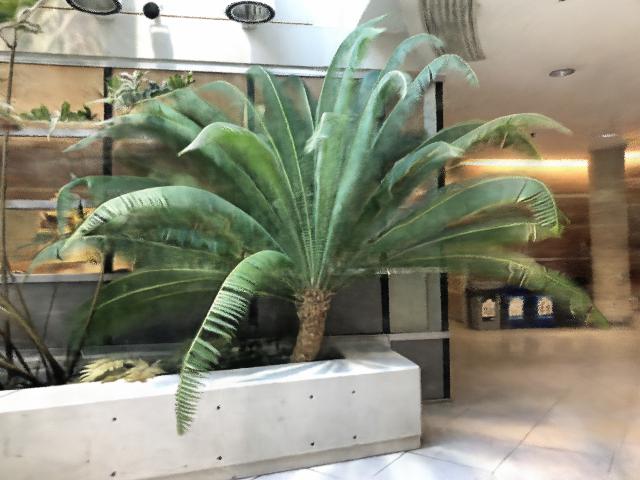} & \includegraphics[width=3cm]{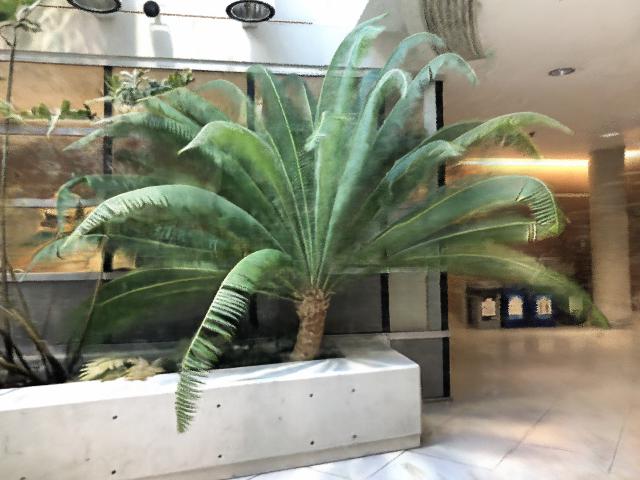}
    & \includegraphics[width=3cm]{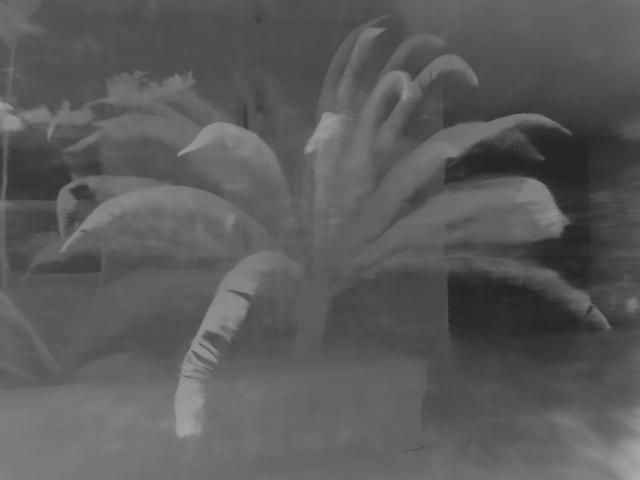} & \includegraphics[width=3cm]{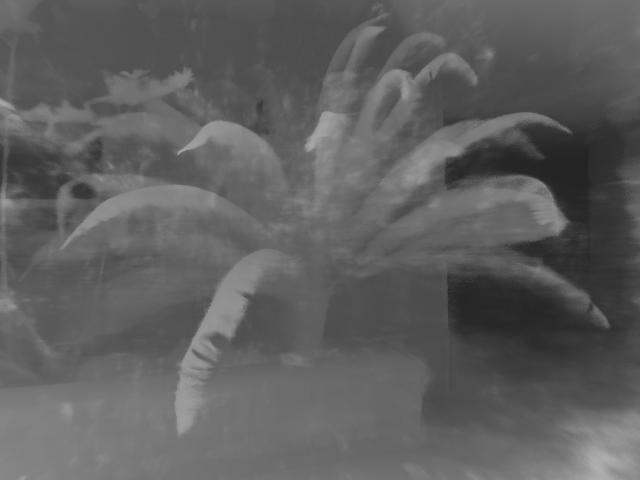}
    \\
  \hline
  \hline
    Fort (38)& \includegraphics[width=3cm]{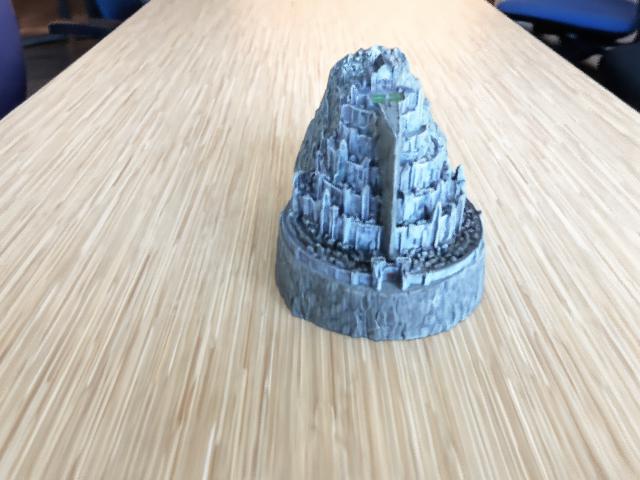} & \includegraphics[width=3cm]{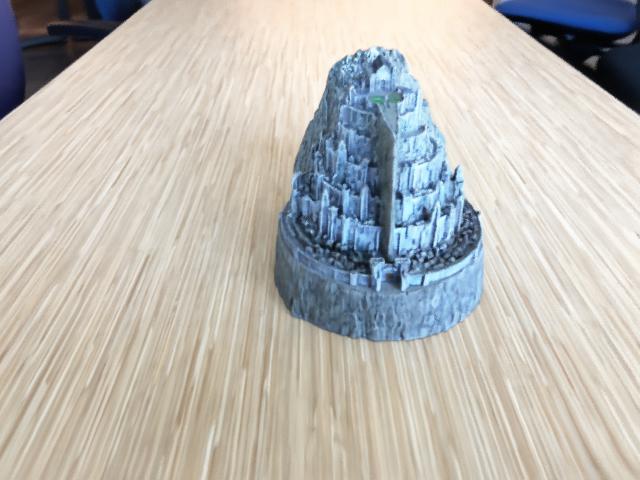}
    & \includegraphics[width=3cm]{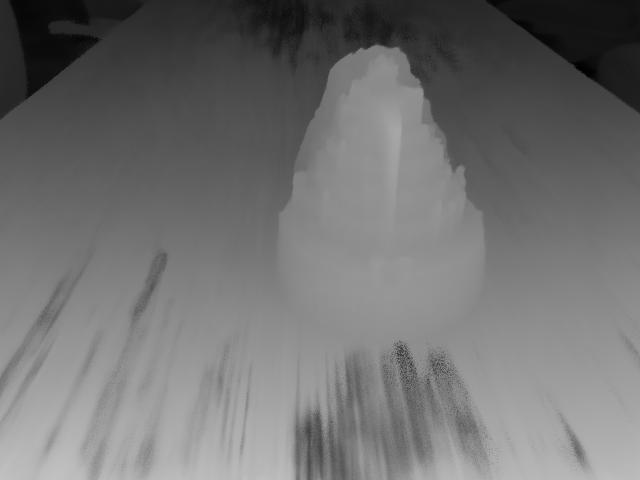} & \includegraphics[width=3cm]{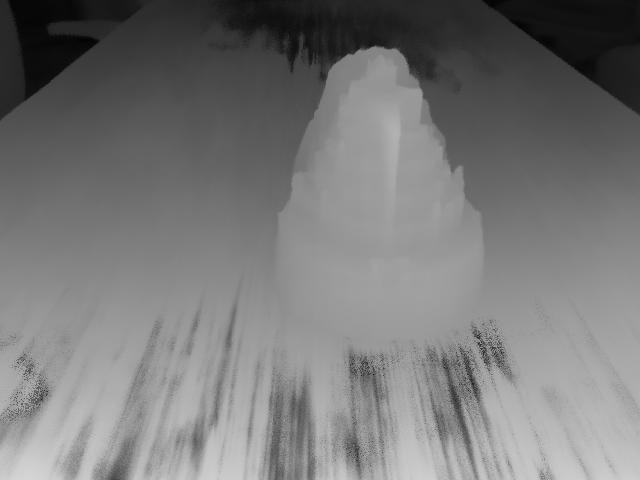}
    \\
    * Fort/2 (19)& \includegraphics[width=3cm]{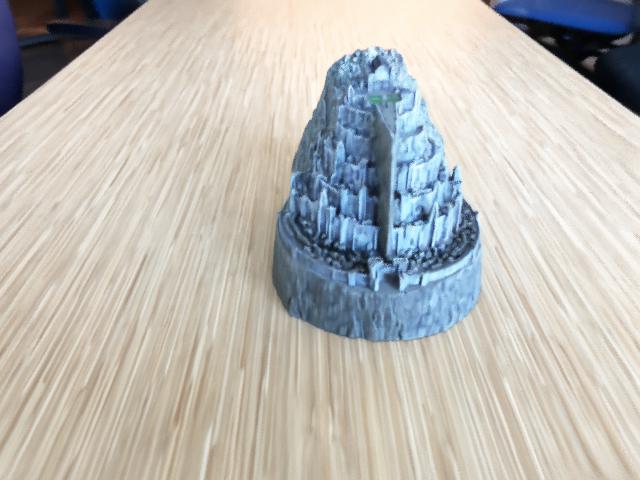} & \includegraphics[width=3cm]{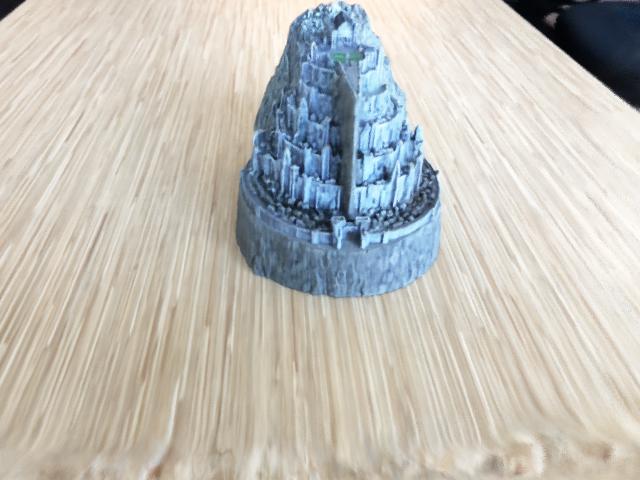}
    & \includegraphics[width=3cm]{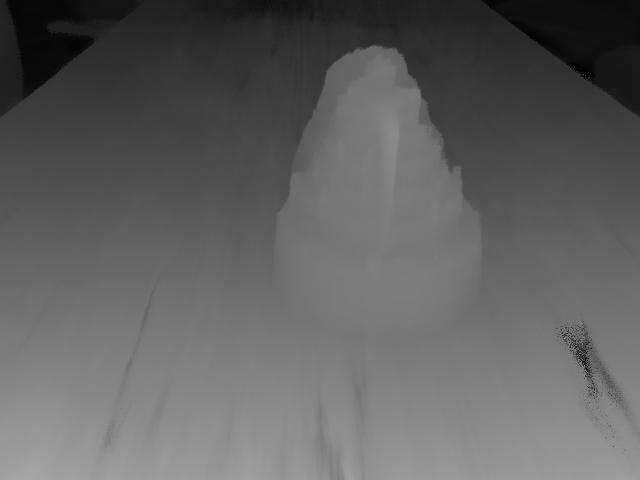} & \includegraphics[width=3cm]{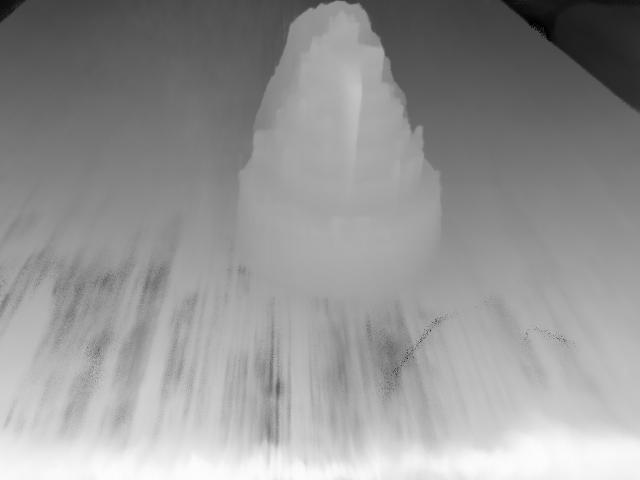}
    \\
    Fort/4 (10) & \includegraphics[width=3cm]{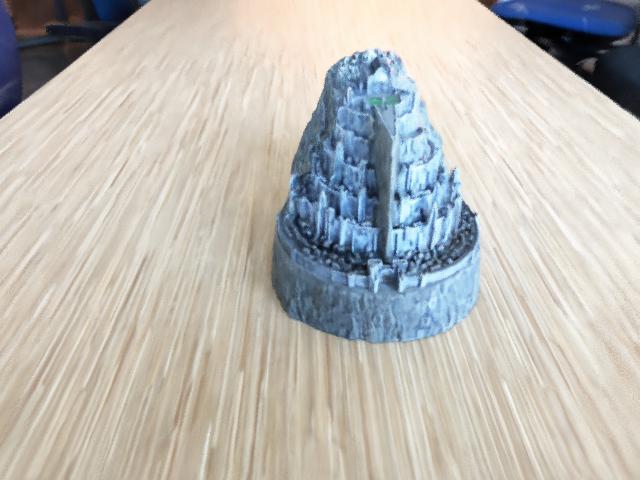} & \includegraphics[width=3cm]{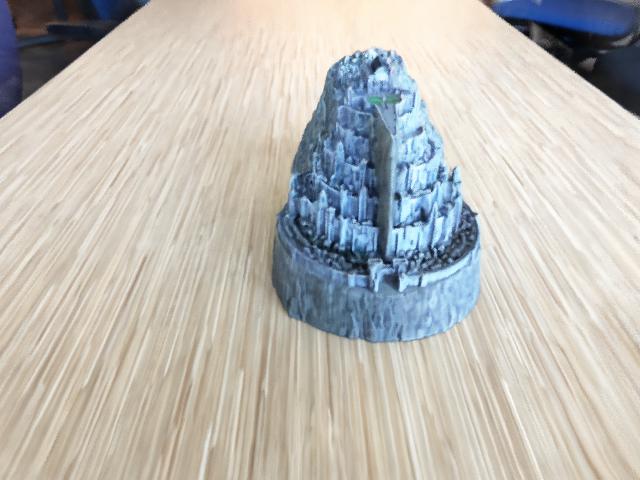}
    & \includegraphics[width=3cm]{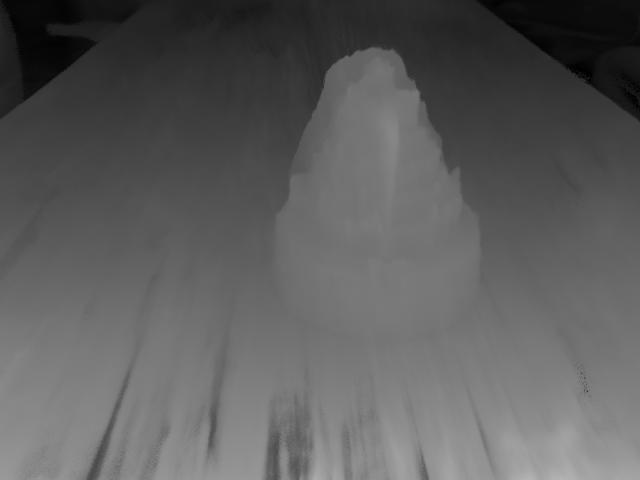} & \includegraphics[width=3cm]{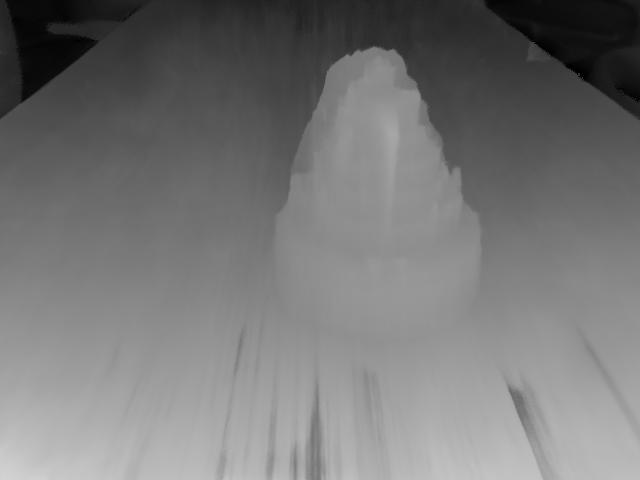}
    \\
  \hline
  \hline
    Orchids (23)& \includegraphics[width=3cm]{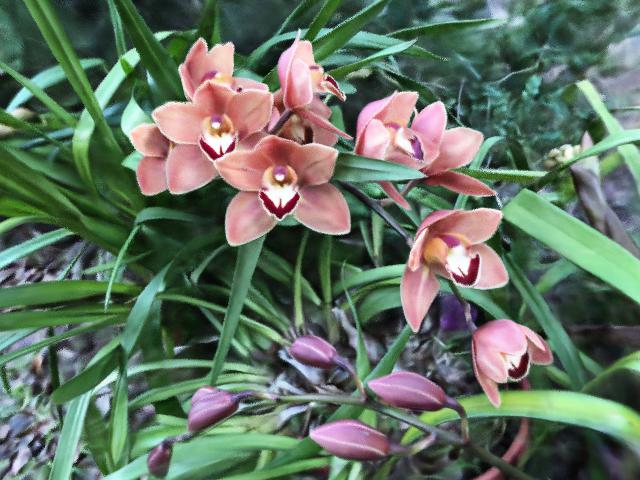} & \includegraphics[width=3cm]{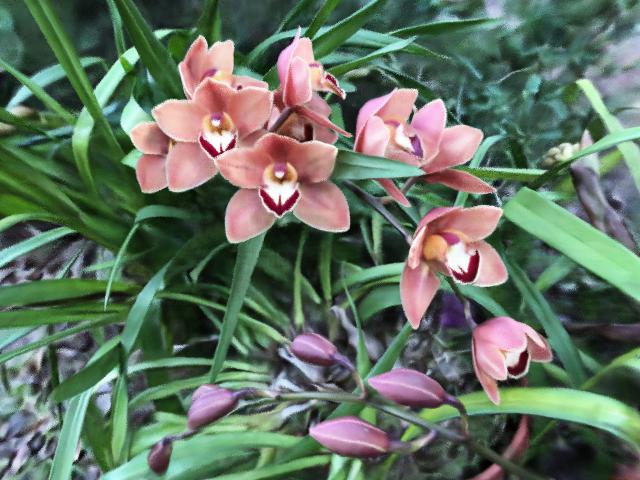}
    & \includegraphics[width=3cm]{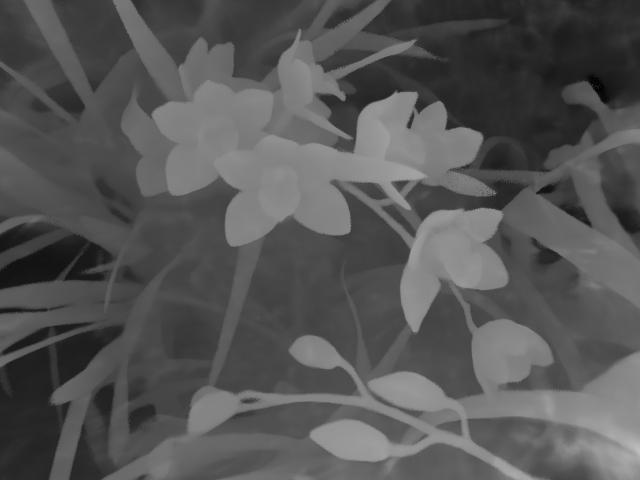} & \includegraphics[width=3cm]{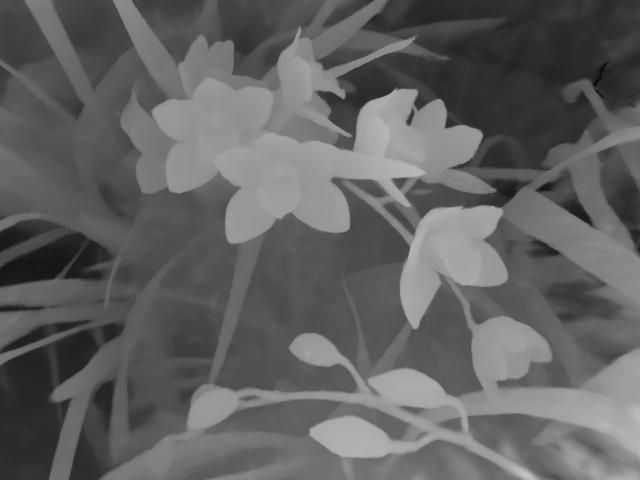}
    \\
    Orchids/2 (12) & \includegraphics[width=3cm]{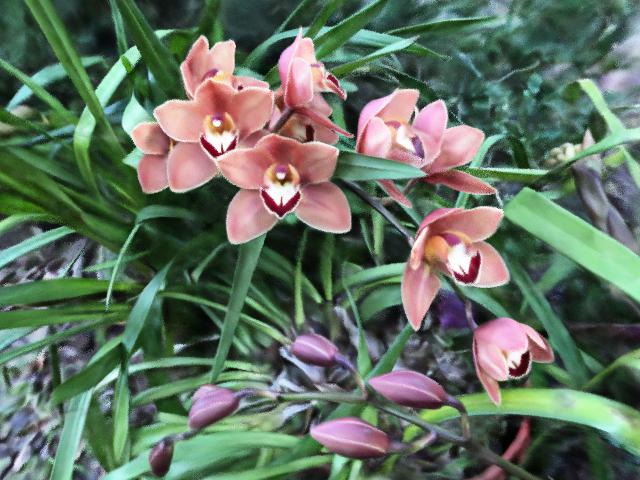} & \includegraphics[width=3cm]{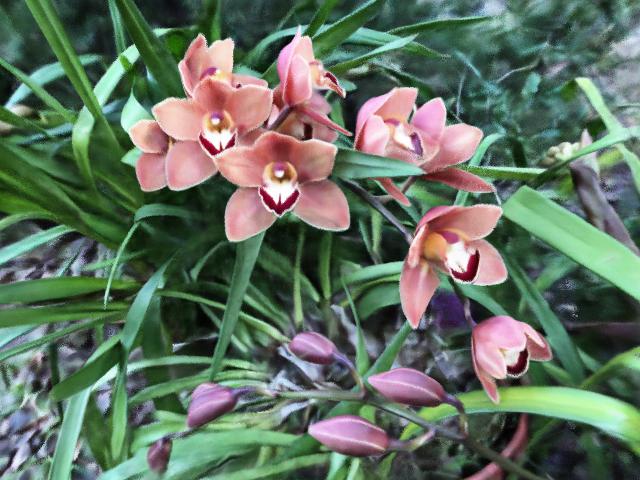}
    & \includegraphics[width=3cm]{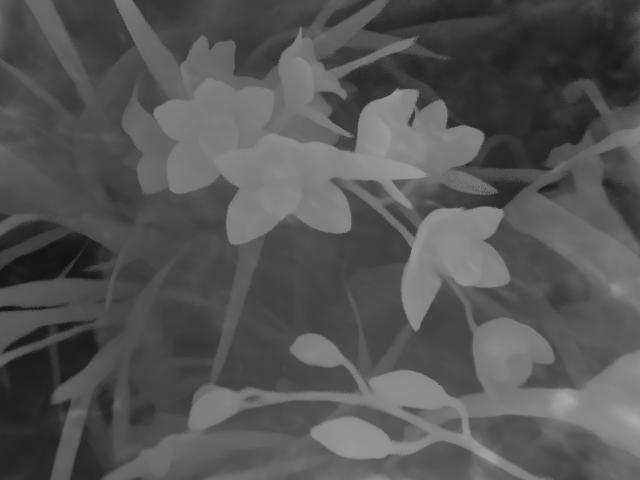} & \includegraphics[width=3cm]{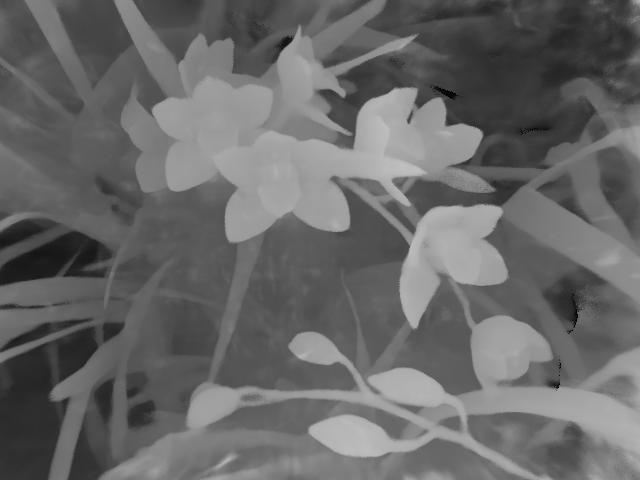}
    \\
    * Orchids/4 (6) & \includegraphics[width=3cm]{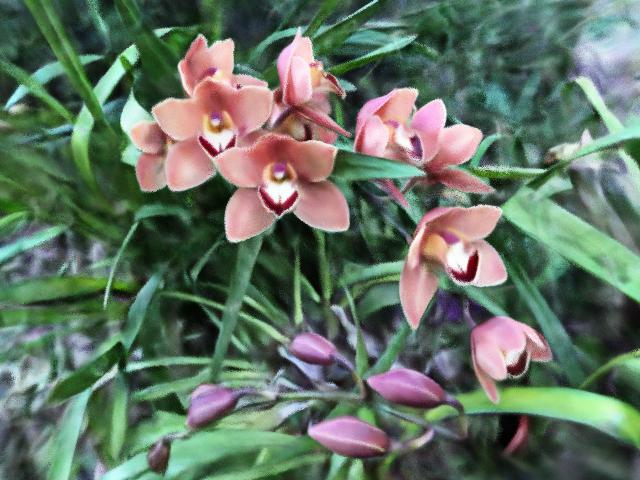} & \includegraphics[width=3cm]{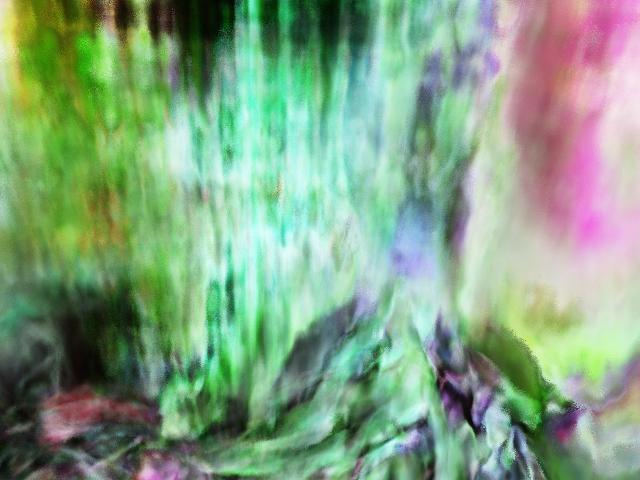}
    & \includegraphics[width=3cm]{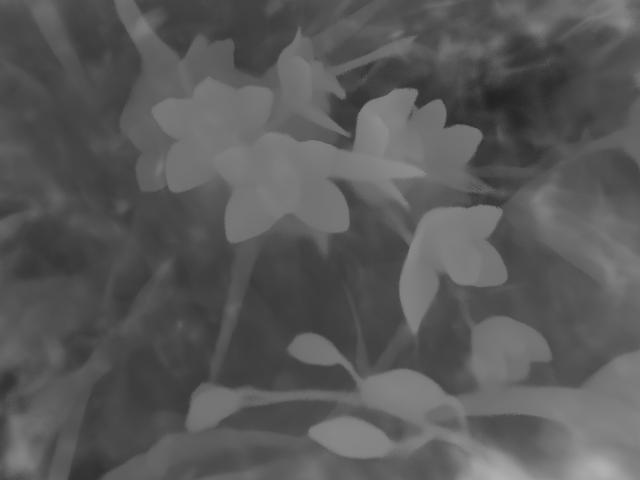} & \includegraphics[width=3cm]{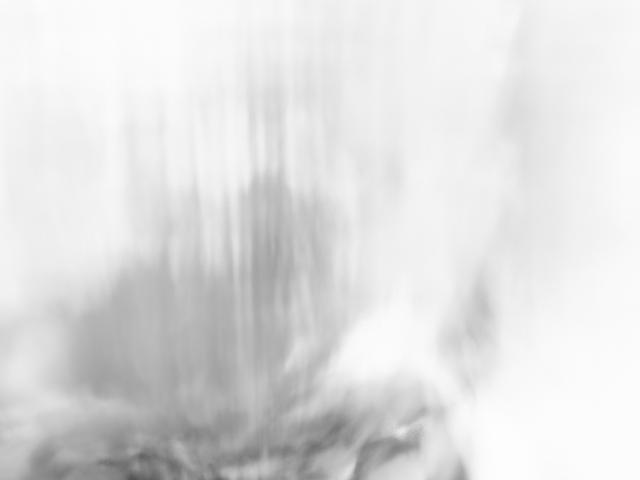}
    \\
  \hline
  \hline
    \end{tabular}

\end{figure*}

\begin{figure*}[h]
    \begin{tabular}{ccc|cc}
     & \multicolumn{2}{c}{RGB} & \multicolumn{2}{c}{Depth} \\
     & ours & BaRF & ours & BaRF  \\

    Room (37)& \includegraphics[width=3cm]{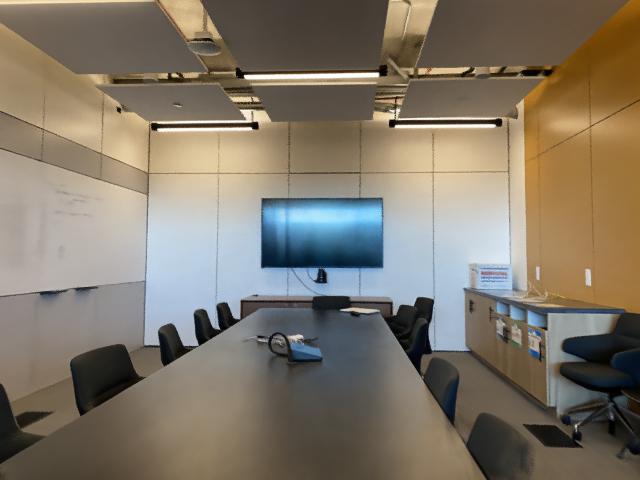} & \includegraphics[width=3cm]{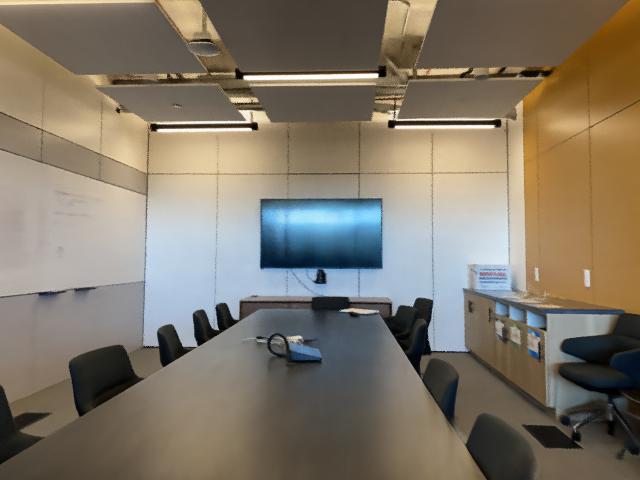}
    & \includegraphics[width=3cm]{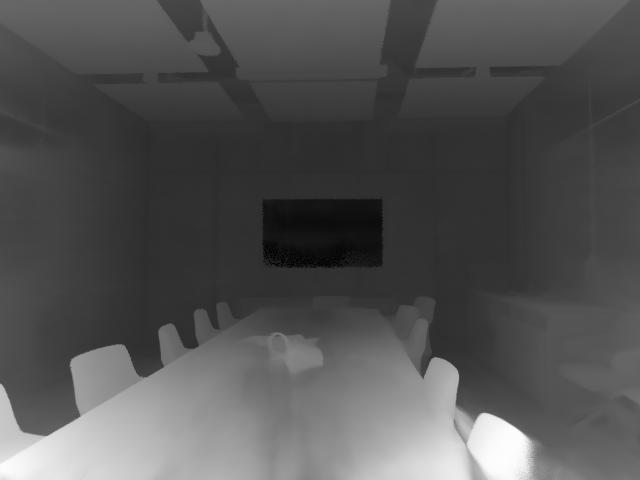} & \includegraphics[width=3cm]{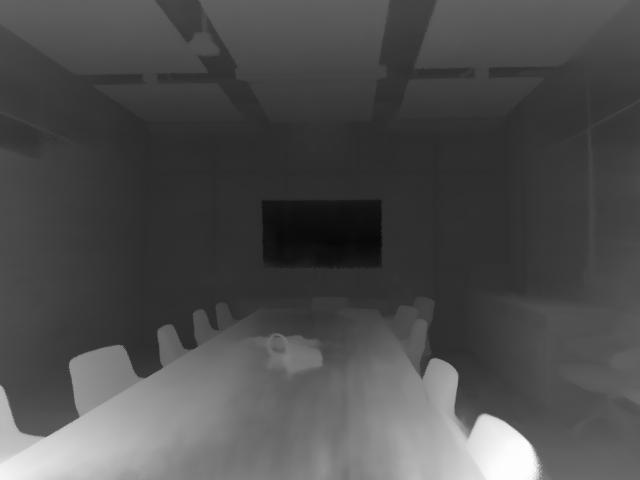}
    \\
    Room/2 (18) & \includegraphics[width=3cm]{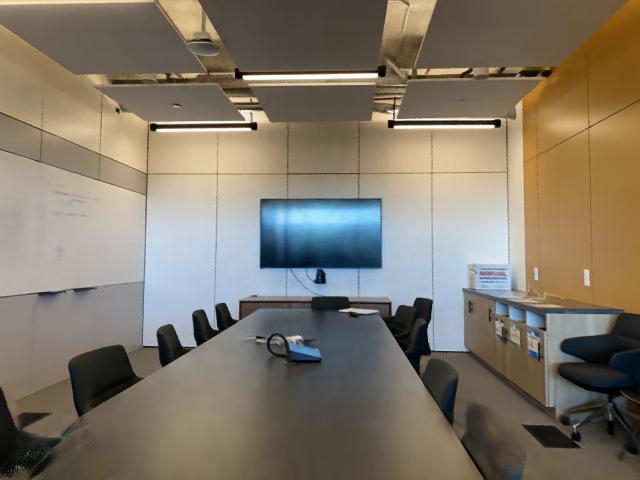} & \includegraphics[width=3cm]{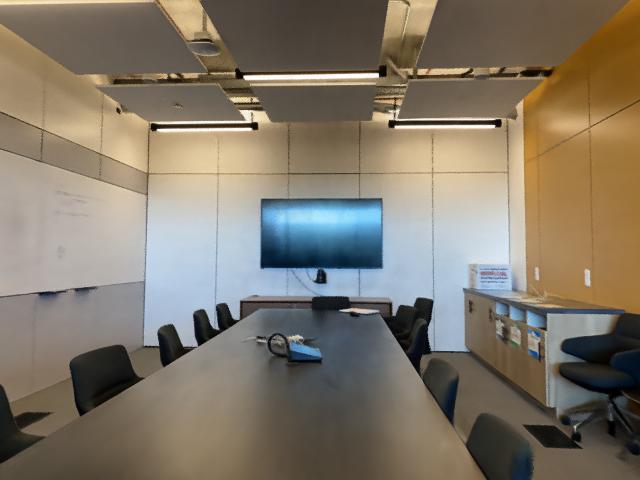}
    & \includegraphics[width=3cm]{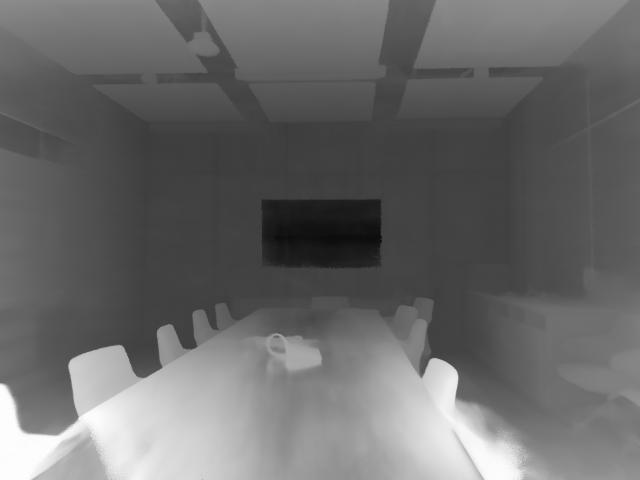} & \includegraphics[width=3cm]{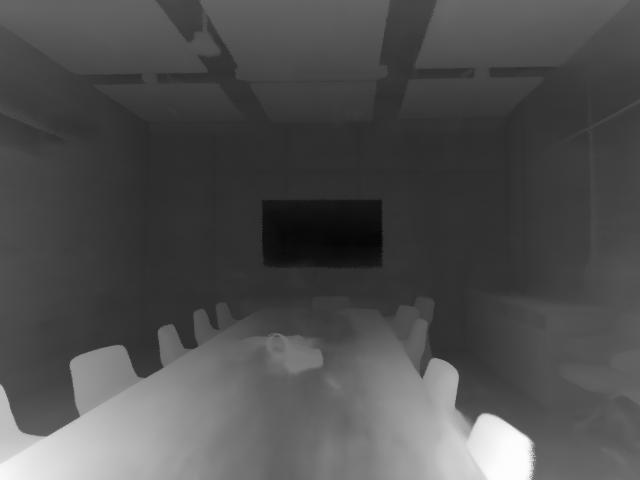}
    \\
    * Room/4 (9) & \includegraphics[width=3cm]{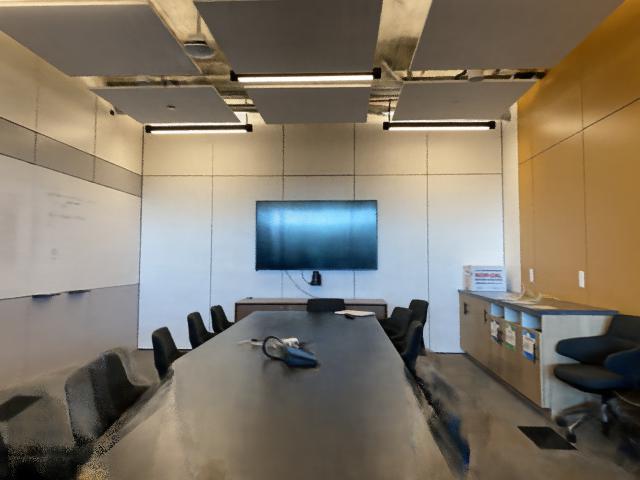} & \includegraphics[width=3cm]{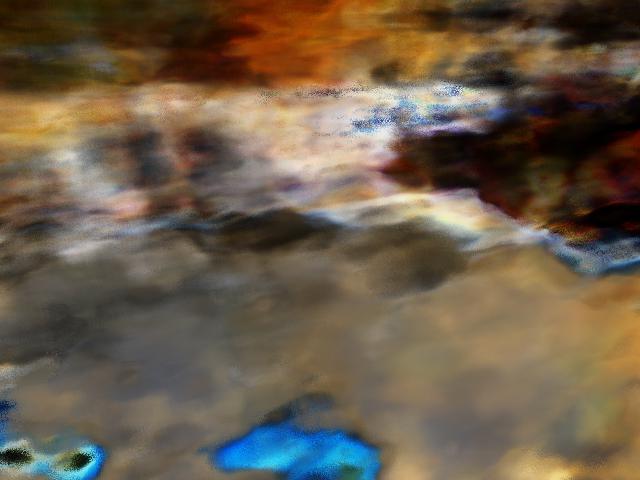}
    & \includegraphics[width=3cm]{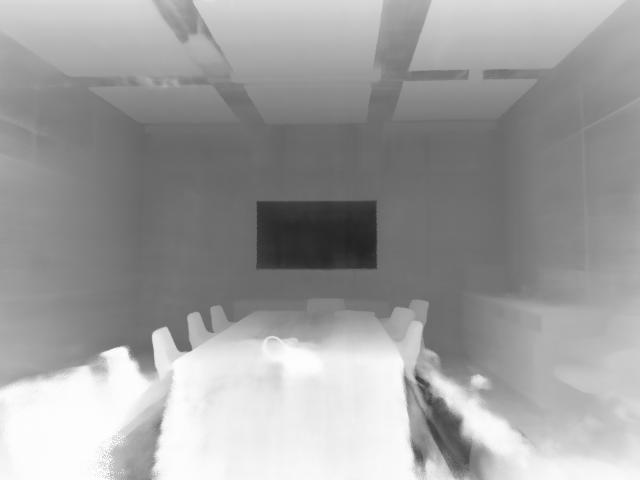} & \includegraphics[width=3cm]{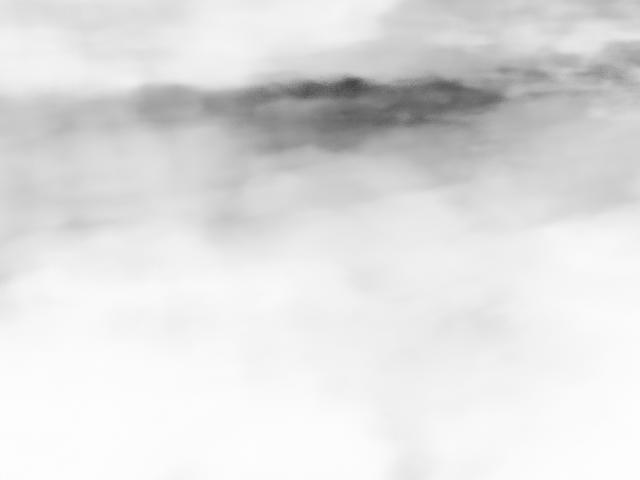}
    \\
  \hline
  \hline

    Flower (34)& \includegraphics[width=3cm]{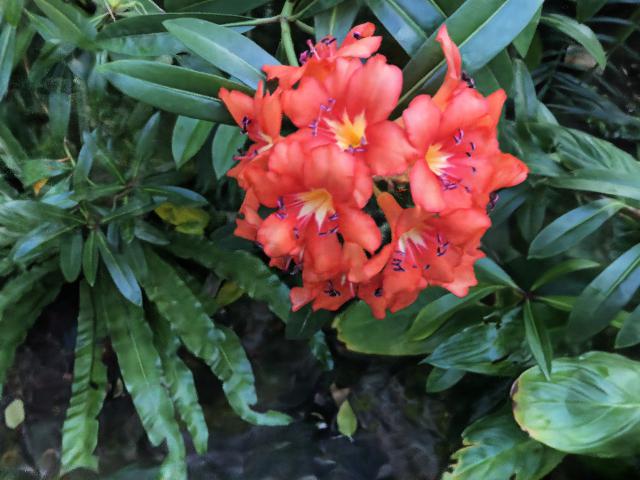} & \includegraphics[width=3cm]{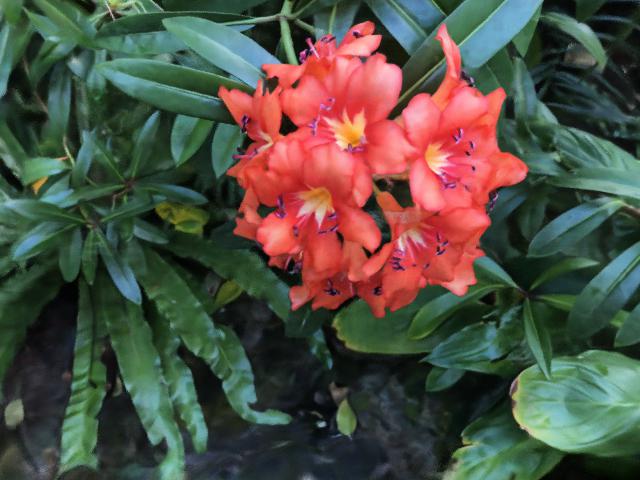}
    & \includegraphics[width=3cm]{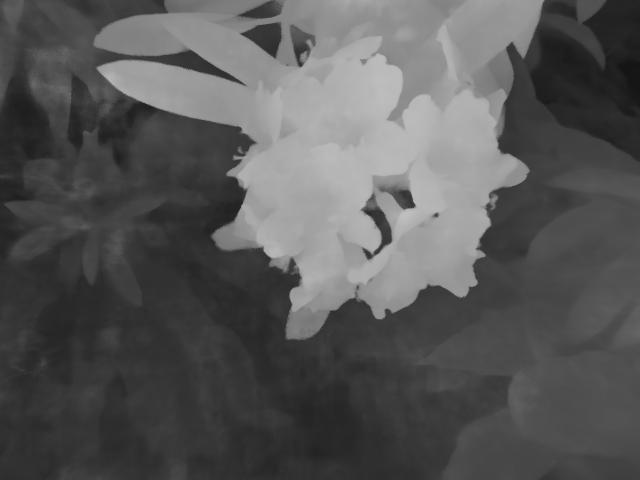} & \includegraphics[width=3cm]{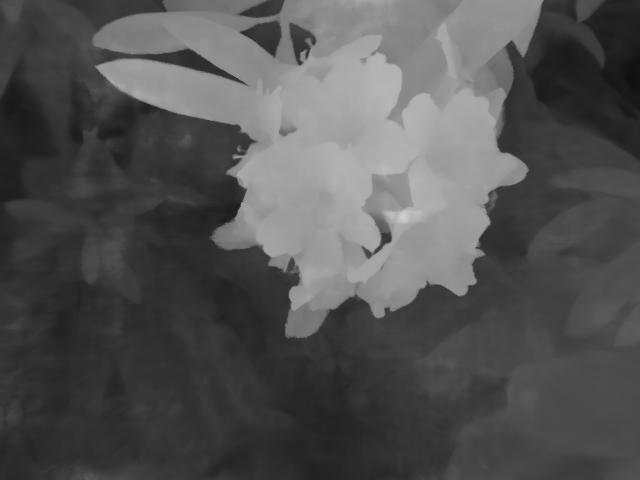}
    \\
    Flower/2 (17) & \includegraphics[width=3cm]{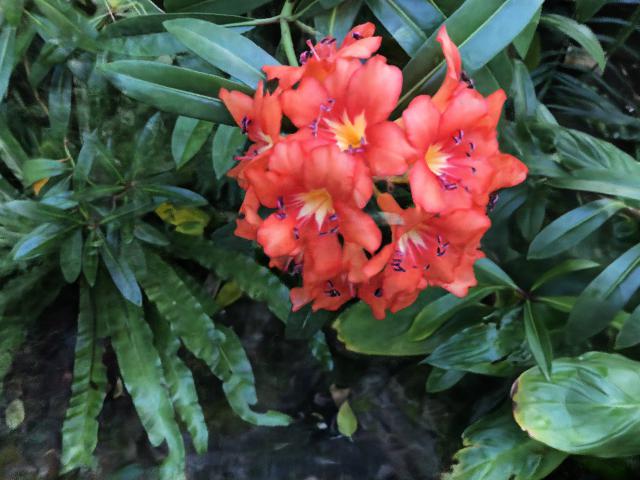} & \includegraphics[width=3cm]{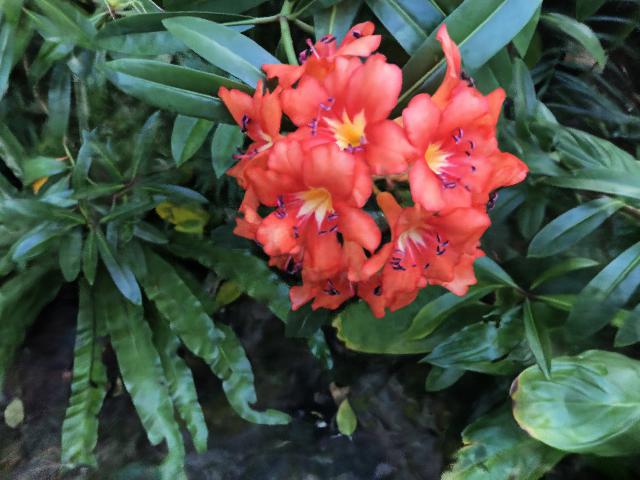}
    & \includegraphics[width=3cm]{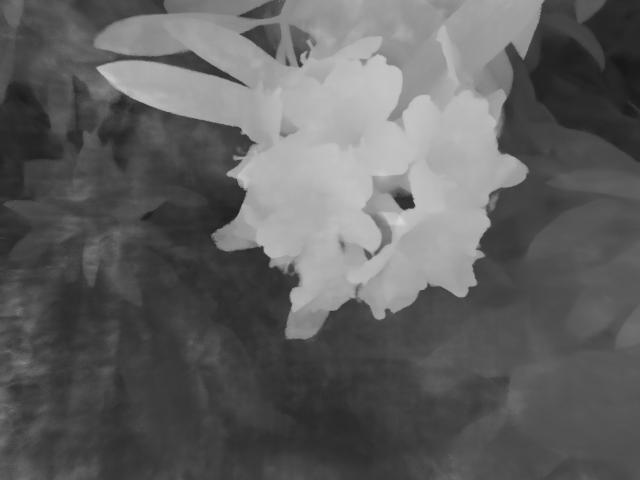} & \includegraphics[width=3cm]{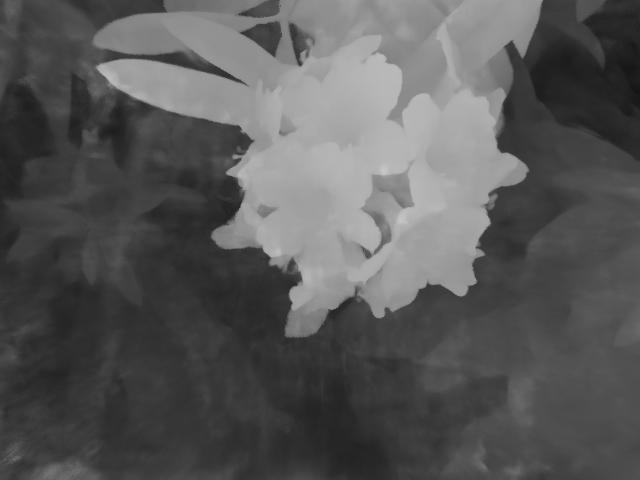}
    \\
    Flower/4 (9) & \includegraphics[width=3cm]{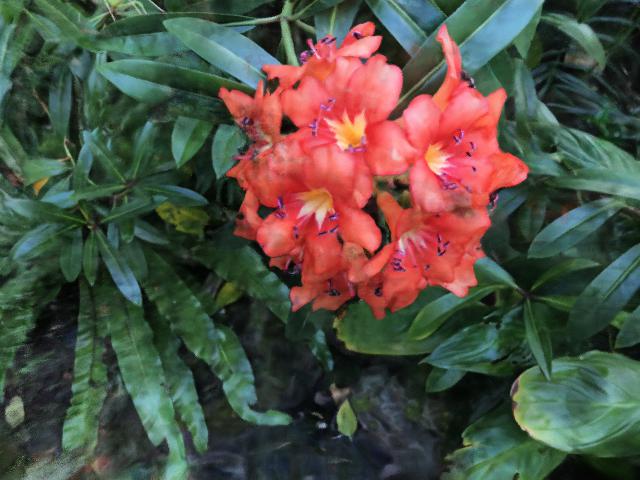} & \includegraphics[width=3cm]{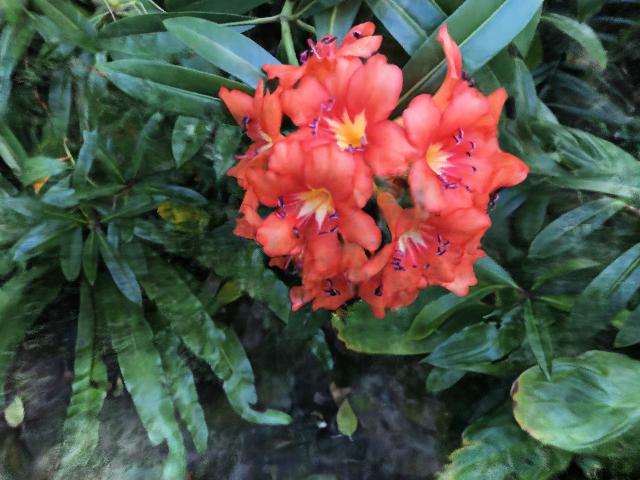}
    & \includegraphics[width=3cm]{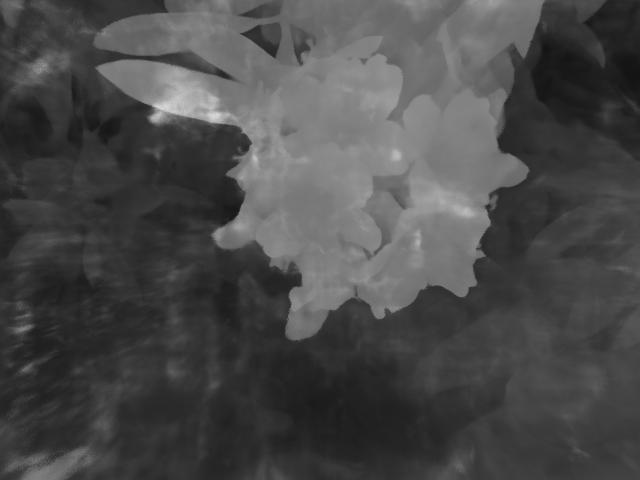} & \includegraphics[width=3cm]{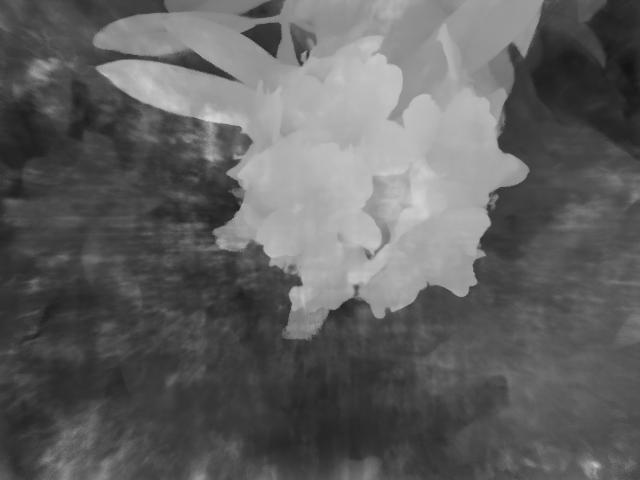}
    \\

  \hline
  \hline

    Horns (56)& \includegraphics[width=3cm]{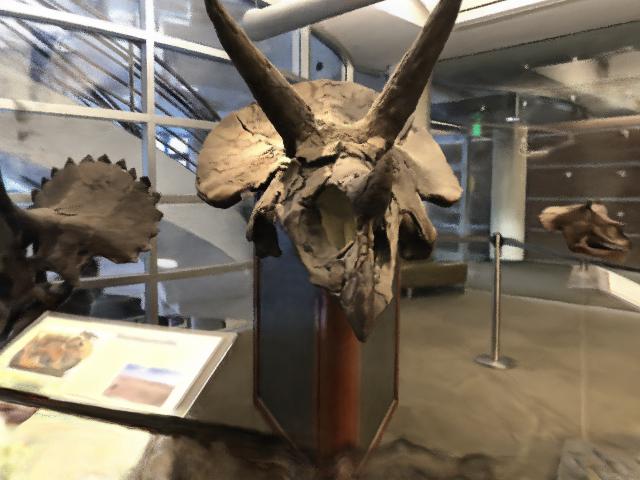} & \includegraphics[width=3cm]{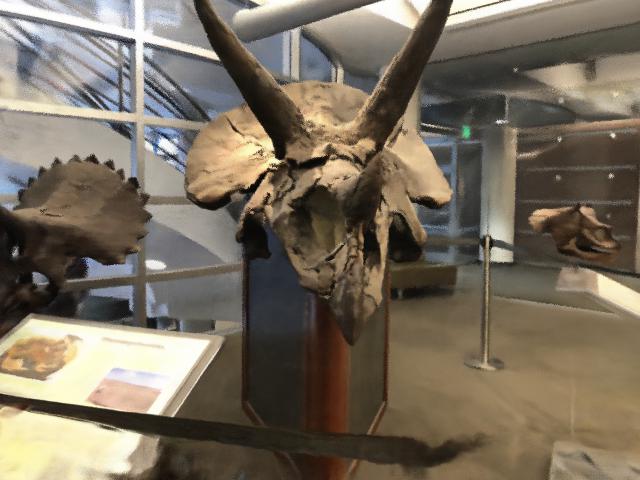}
    & \includegraphics[width=3cm]{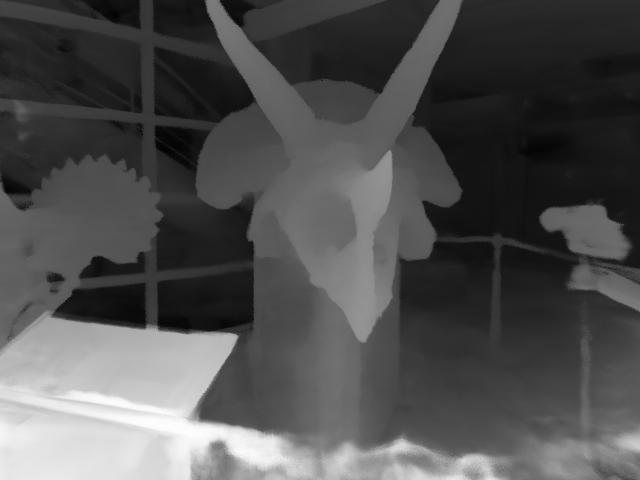} & \includegraphics[width=3cm]{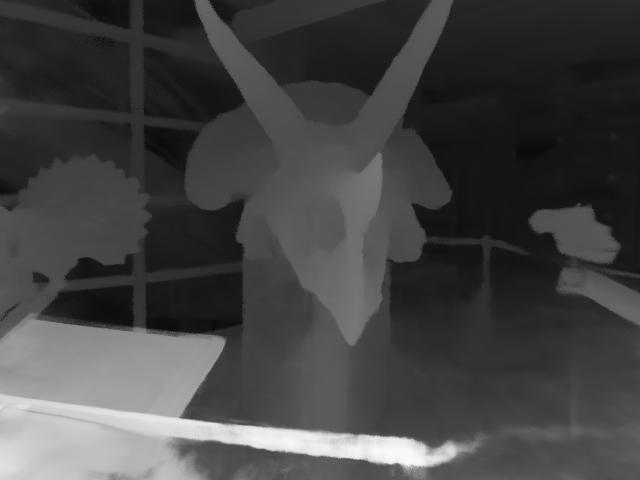}
    \\
    Horns/2 (28) & \includegraphics[width=3cm]{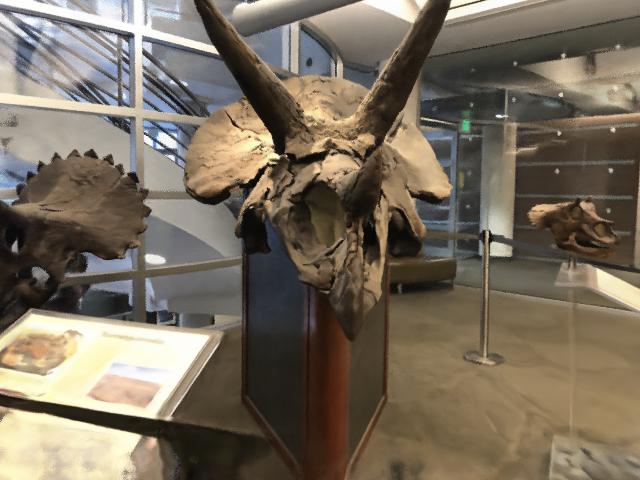} & \includegraphics[width=3cm]{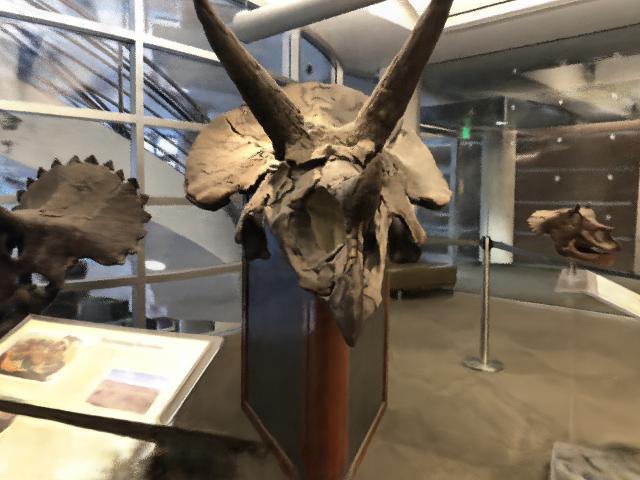}
    & \includegraphics[width=3cm]{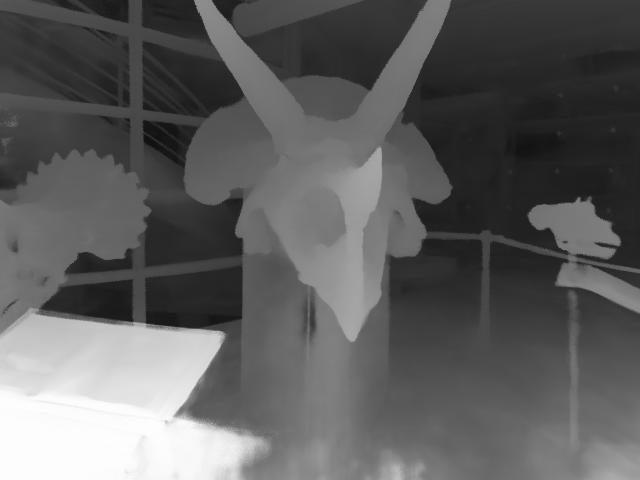} & \includegraphics[width=3cm]{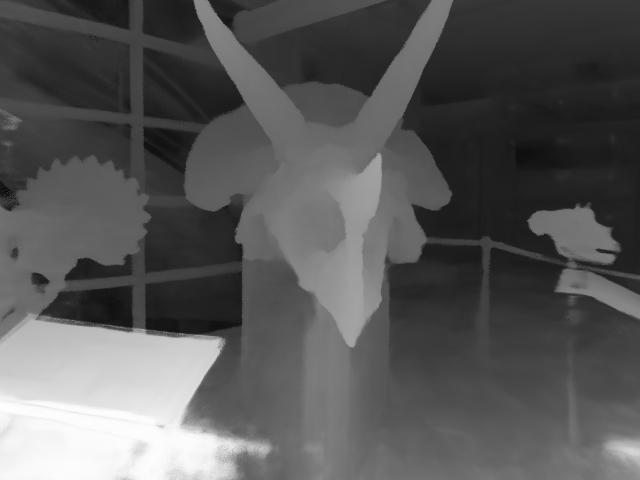}
    \\
    Horns/4 (14) & \includegraphics[width=3cm]{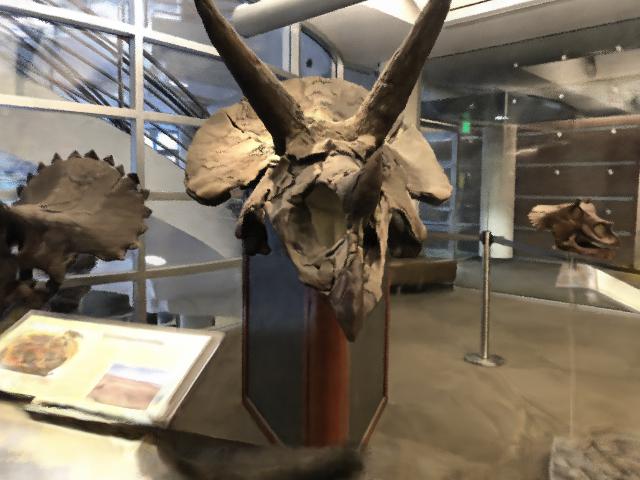} & \includegraphics[width=3cm]{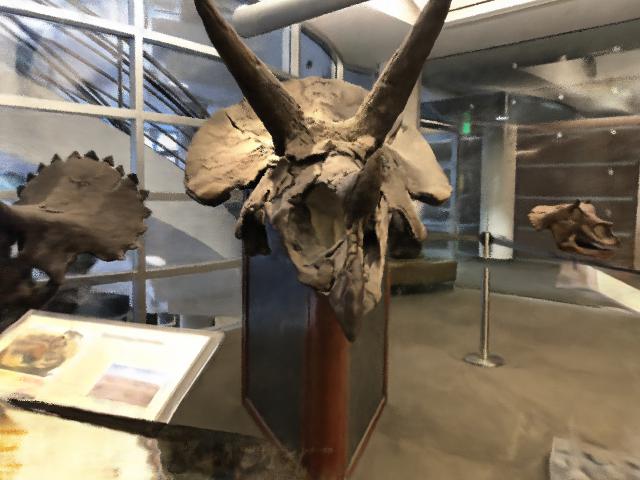}
    & \includegraphics[width=3cm]{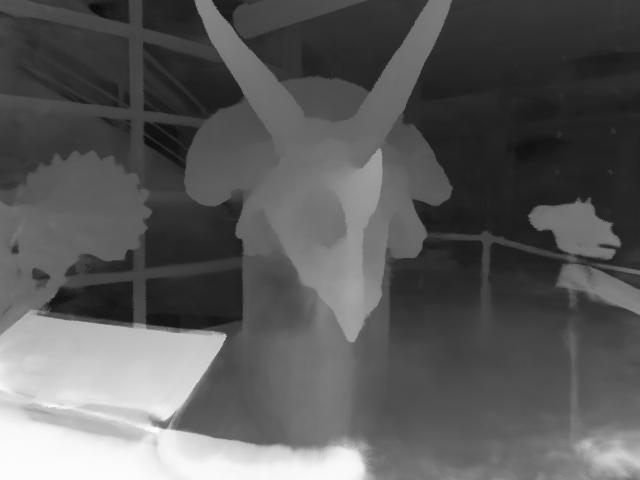} & \includegraphics[width=3cm]{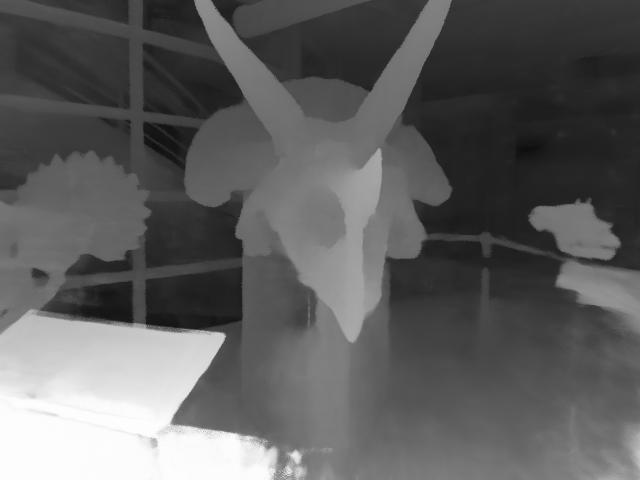}
    \\
    
    \end{tabular}
    \vspace{1mm}
    

\end{figure*}

\begin{table*}[b]
\centering
\resizebox{0.8\linewidth}{!}{
\begin{tabular}{lcccccccccc}
\toprule

\multirow{ 2}{*}{Scene}  & \multicolumn{2}{|c|}{Rotation$\textdegree{}$ $\downarrow$} & \multicolumn{2}{|c|}{Translation $\downarrow$} & \multicolumn{2}{|c|}{PSNR $\uparrow$} & \multicolumn{2}{|c|}{SSIM $\uparrow$}  & \multicolumn{2}{|c}{LPIPS $\downarrow$}   \\
\cmidrule{2-11}
 & \multicolumn{1}{|c}{BARF}  & ours  & \multicolumn{1}{|c}{BARF}  & ours & \multicolumn{1}{|c}{BARF}  & ours &  \multicolumn{1}{|c}{BARF}  & ours & \multicolumn{1}{|c}{BARF}  & ours \\
\midrule
Flower& \multicolumn{1}{|c}{0.64}  & 0.49 & \multicolumn{1}{|c}{0.27} &  0.25 &  \multicolumn{1}{|c}{17.18}  & 17.93  & \multicolumn{1}{|c}{0.34}  & 0.41 & \multicolumn{1}{|c}{0.27}  & 0.21 \\
Flower$/2$& \multicolumn{1}{|c}{0.62}  & 0.46  & \multicolumn{1}{|c}{0.28} &  0.25 &  \multicolumn{1}{|c}{17.18}  & 17.94  & \multicolumn{1}{|c}{0.34}  & 0.36 & \multicolumn{1}{|c}{0.27}  & 0.23 \\
Flower$/4$& \multicolumn{1}{|c}{0.59}  & 0.54 & \multicolumn{1}{|c}{0.31} & 0.29  &  \multicolumn{1}{|c}{17.07}  &  17.38 & \multicolumn{1}{|c}{0.33}  & 0.38  & \multicolumn{1}{|c}{0.29}  &0.27  \\

\midrule
Horns  &  \multicolumn{1}{|c}{0.18}  & 0.19  & \multicolumn{1}{|c}{0.18} &  0.18 &  \multicolumn{1}{|c}{19.58}  & 18.89  & \multicolumn{1}{|c}{0.59}  & 0.55 & \multicolumn{1}{|c}{0.32}  & 0.27  \\
Horns$/2$   &  \multicolumn{1}{|c}{0.27}  & 0.33  & \multicolumn{1}{|c}{0.20} & 0.17  &  \multicolumn{1}{|c}{16.24}  & 16.09  & \multicolumn{1}{|c}{0.49}  & 0.45 & \multicolumn{1}{|c}{0.31}  & 0.28 \\
Horns$/4$  &  \multicolumn{1}{|c}{0.21}  & 0.24 & \multicolumn{1}{|c}{0.16} &  0.17 &  \multicolumn{1}{|c}{16.91}  &  16.85 & \multicolumn{1}{|c}{0.54}  &  0.53 & \multicolumn{1}{|c}{0.32}  & 0.32 \\

\midrule

Trex  &  \multicolumn{1}{|c}{0.49}  & 0.41 & \multicolumn{1}{|c}{0.38} & 0.35  &  \multicolumn{1}{|c}{16.53}  & 17.04  & \multicolumn{1}{|c}{0.42}  & 0.45 & \multicolumn{1}{|c}{0.21}  & 0.19 \\

Trex$/2$ &  \multicolumn{1}{|c}{0.56}  & 0.26  & \multicolumn{1}{|c}{0.43} & 0.29  &  \multicolumn{1}{|c}{16.37}  &  18.96 & \multicolumn{1}{|c}{0.40}  & 0.61 & \multicolumn{1}{|c}{0.23}  & 0.16 \\

Trex$/4$  &  \multicolumn{1}{|c}{0.19}  &  0.20  & \multicolumn{1}{|c}{0.24} & 0.26  &  \multicolumn{1}{|c}{21.62}  &   20.74 & \multicolumn{1}{|c}{0.73}  & 0.70 & \multicolumn{1}{|c}{0.17}  & 0.15 \\

\midrule
Average  &  \multicolumn{1}{|c}{0.42}  & \textbf{0.34} & \multicolumn{1}{|c}{0.27} & \textbf{0.24}  &  \multicolumn{1}{|c}{17.63}  &  \textbf{17.95} & \multicolumn{1}{|c}{0.46}  & \textbf{0.49} & \multicolumn{1}{|c}{0.27}  &\textbf{0.23}  \\



\bottomrule
\end{tabular}}
\vspace{1mm}
\captionof{table}{\textbf{More quantitative results on real datasets}. In addition to Table \ref{tab:barf_real_overall} we report more results on LLFF \cite{mildenhall2019llff} dataset. Note that in this dataset images are captured in a top-down, left-right manner rather than following a continuous trajectory. Consequently, our method may not be fully leveraged. Nevertheless, when considering average values, our approach outperforms the baseline.}
\label{tab:more_llff}
\end{table*}

\begin{figure*}
  \centering
  \resizebox{\linewidth}{!}{
  \begin{tabular}{ccccc}
     & offset 0.2388 \textdegree{} & offset 1.5 \textdegree{} & offset 2.85 \textdegree{} \\
    \multirow{2}{*}{\includegraphics[width=4cm]{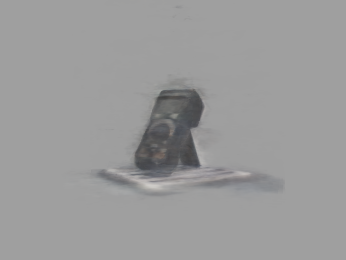}} & \includegraphics[width=3cm]{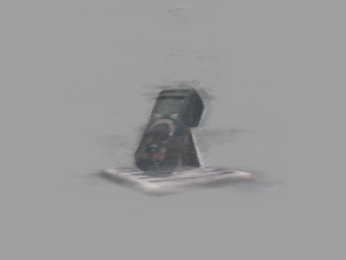} & \includegraphics[width=3cm]{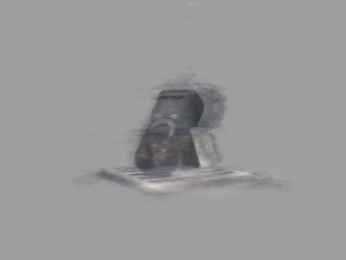} & \includegraphics[width=3cm]{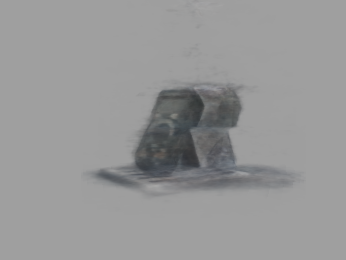} \\
    
     & \includegraphics[width=3cm, trim={0  0  2cm 0 },clip]{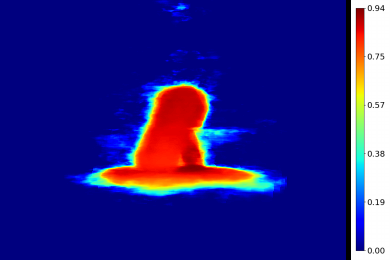} & \includegraphics[width=3cm, trim={0  0  2cm 0 },clip]{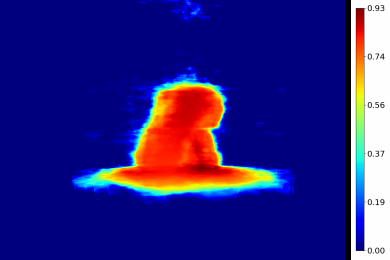} & \includegraphics[width=3cm, trim={0  0  2cm 0 },clip]{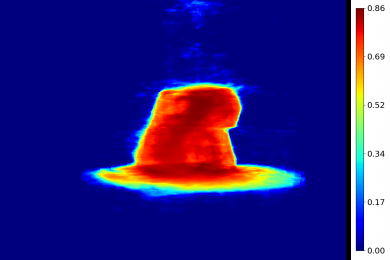} \\

    & \multicolumn{3}{c}{EventNeRF} \\

    \multirow{2}{*}{\includegraphics[width=4cm,trim={0  1cm 2cm 0 },clip]{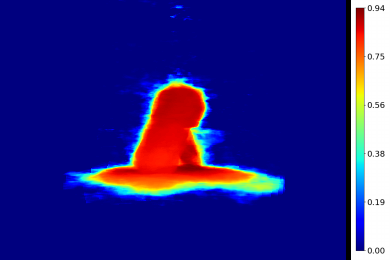}} & \includegraphics[width=3cm]{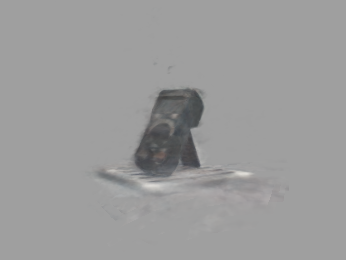} & \includegraphics[width=3cm]{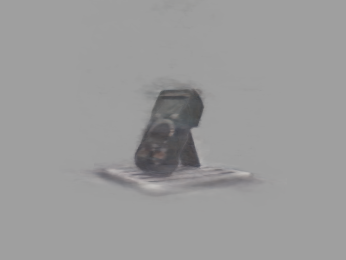} & \includegraphics[width=3cm]{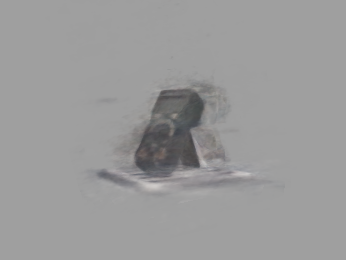} \\
    
     & \includegraphics[width=3cm, trim={0  0  2cm 0 },clip]{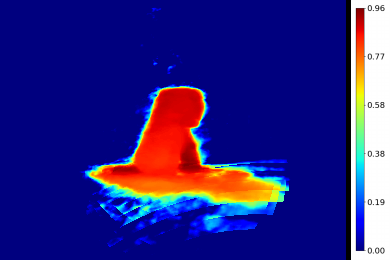} & \includegraphics[width=3cm, trim={0  0  2cm 0 },clip]{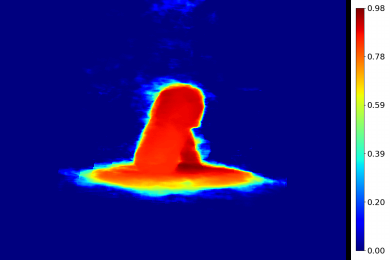} & \includegraphics[width=3cm, trim={0  0  2cm 0 },clip]{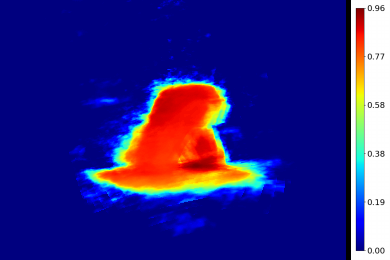} \\
    
    Multimeter Dataset (with calibration)  & \multicolumn{3}{c}{Ours} \\
  \hline
  \hline
  \end{tabular}}

  \resizebox{\linewidth}{!}{
  \begin{tabular}{ccccc}
    \multirow{2}{*}{\includegraphics[width=4cm]{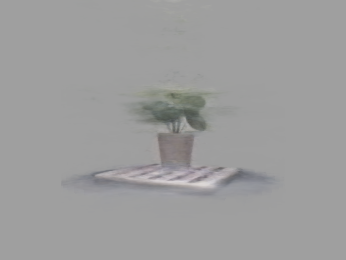}} & \includegraphics[width=3cm]{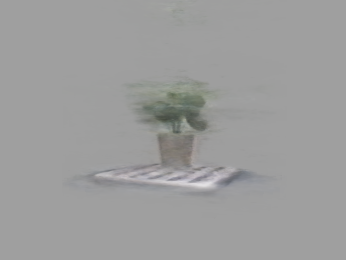} & \includegraphics[width=3cm]{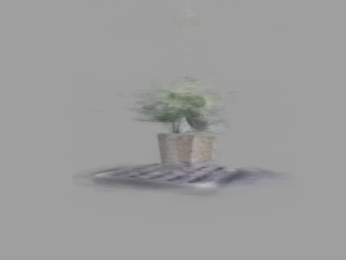} & \includegraphics[width=3cm]{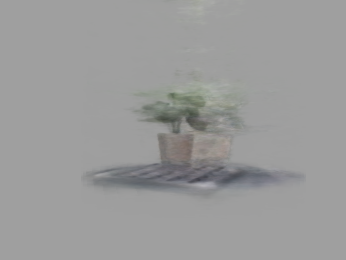} \\
    
     & \includegraphics[width=3cm, trim={0  0  2cm 0 },clip]{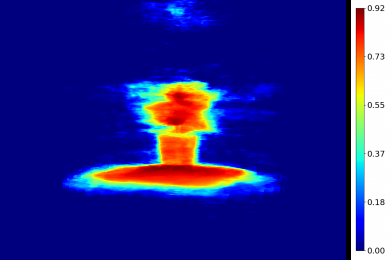} & \includegraphics[width=3cm, trim={0  0  2cm 0 },clip]{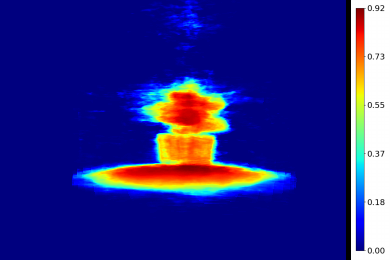} & \includegraphics[width=3cm, trim={0  0  2cm 0 },clip]{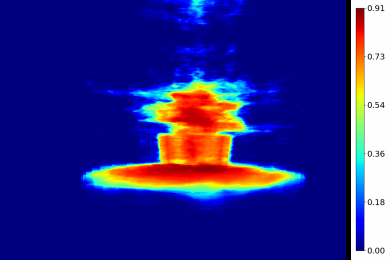} \\

    & \multicolumn{3}{c}{EventNeRF} \\

    \multirow{2}{*}{\includegraphics[width=4cm,trim={0  1cm 2cm 0 },clip]{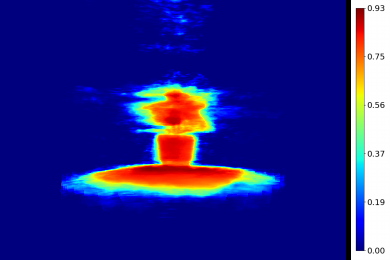}} & \includegraphics[width=3cm]{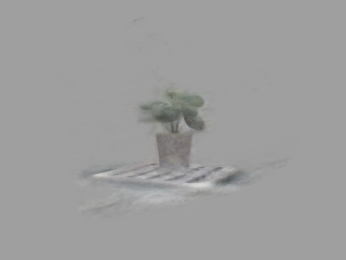} & \includegraphics[width=3cm]{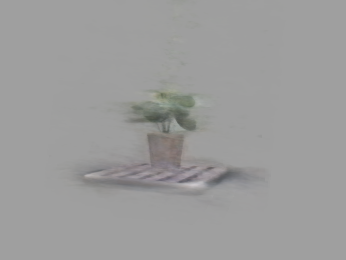} & \includegraphics[width=3cm]{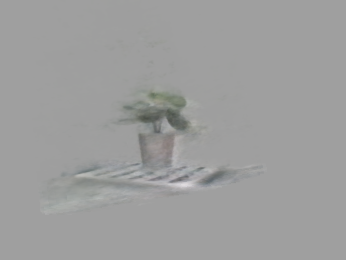} \\
    
     & \includegraphics[width=3cm, trim={0  0  2cm 0 },clip]{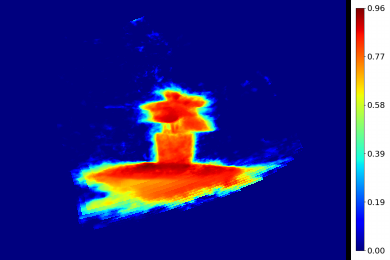} & \includegraphics[width=3cm, trim={0  0  2cm 0 },clip]{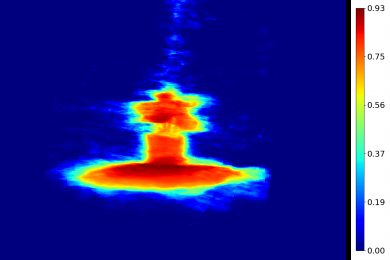} & \includegraphics[width=3cm, trim={0  0  2cm 0 },clip]{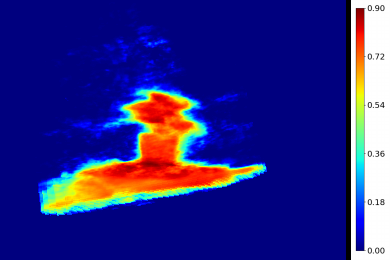} \\
    
    Plant Dataset (with calibration)  & \multicolumn{3}{c}{Ours} \\


    
  \end{tabular}}

\label{fig:multi_plant} 

\end{figure*}

\begin{figure}
    \centering
    \begin{subfigure}{0.5\textwidth}
        \centering
        \includegraphics[width=0.9\linewidth]{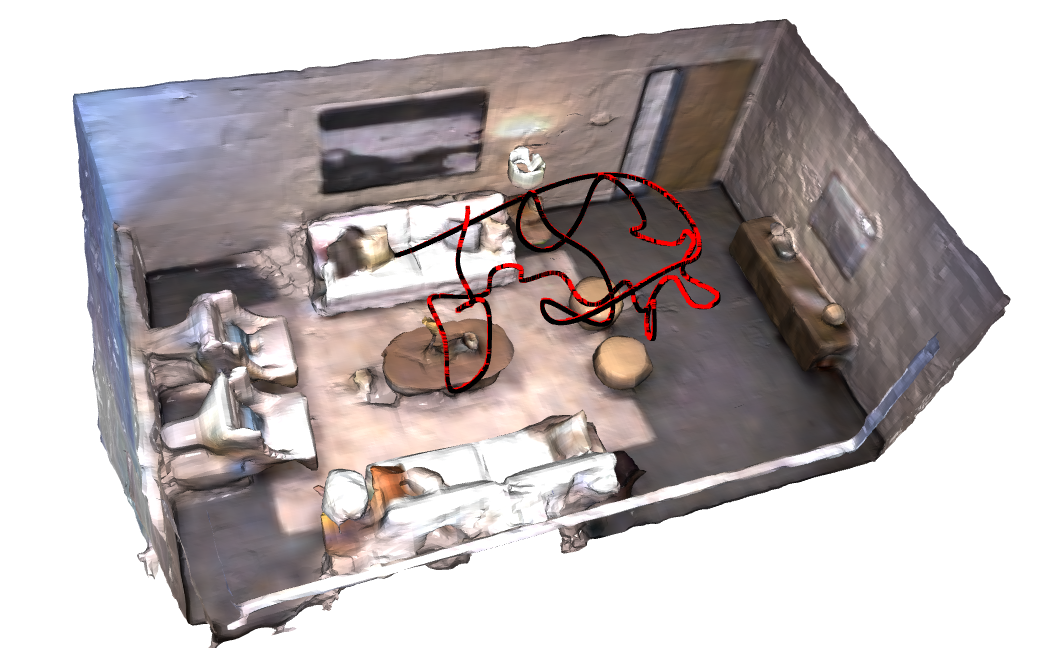}
        \caption{NICE-SLAM}
         \vspace{1em}
        \label{fig:render_nice-r0}
    \end{subfigure}
    \hfill
    \begin{subfigure}{0.5\textwidth}
        \centering
        \includegraphics[width=0.9\linewidth]{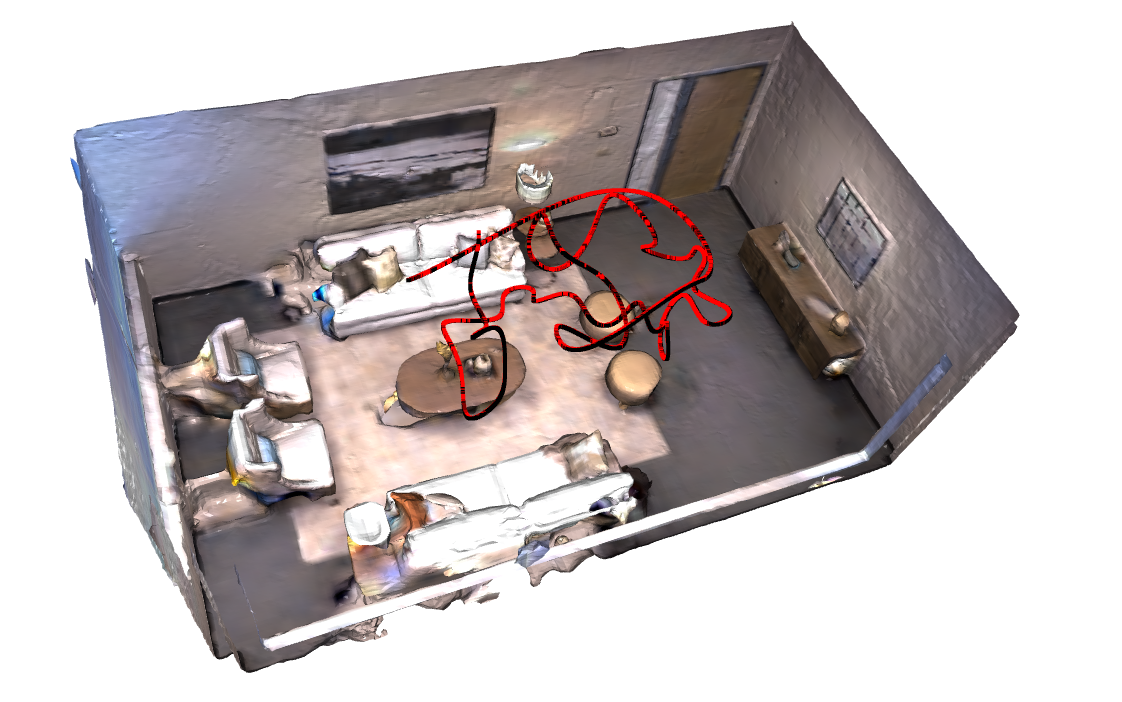}
        \caption{Ours}
        \vspace{1em}
        \label{fig:render_ours-r0}
    \end{subfigure}
    \begin{subfigure}{0.5\textwidth}
        \centering
        \includegraphics[width=0.9\linewidth]{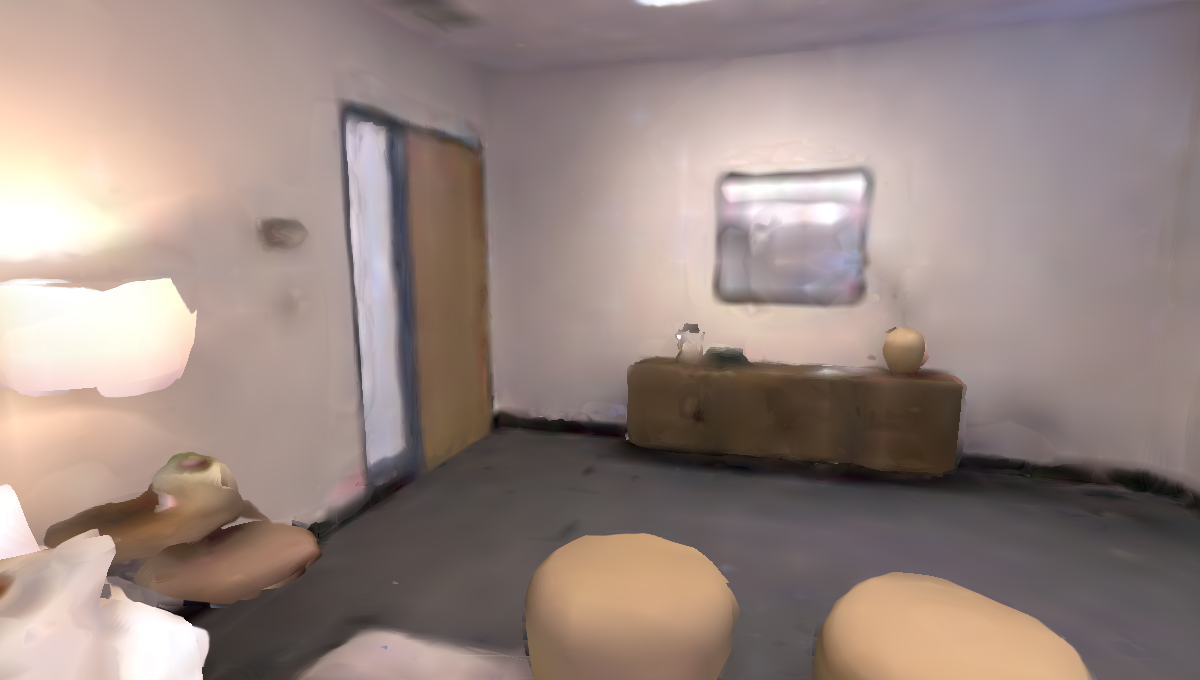}
        \caption{NICE-SLAM PSNR: 33.9}
        \vspace{1em}
        \label{fig:render_nice-r0-rgb}
    \end{subfigure}
    \begin{subfigure}{0.5\textwidth}
        \centering
        \includegraphics[width=0.9\linewidth]{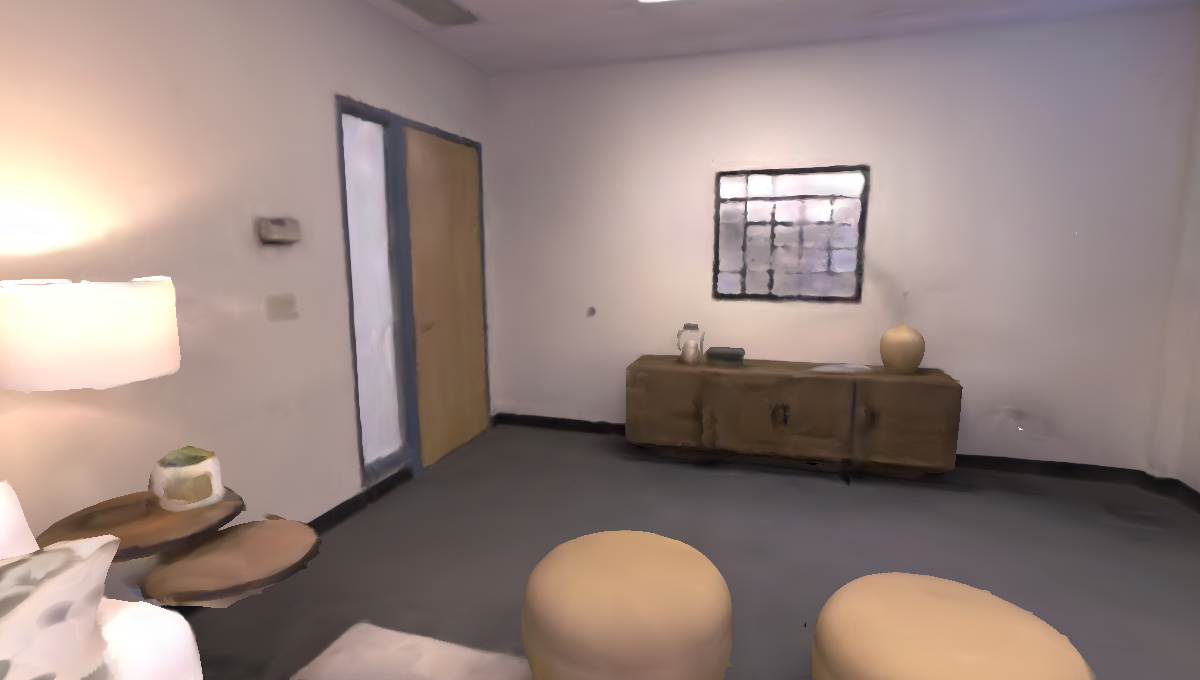}
        \caption{Ours PSNR: 36.9}
        \label{fig:render_ours-r0-rgb}
    \end{subfigure}%
    
    \caption{\textbf{Reconstruction and Rendering of Replica Room0.} Thanks to the improvement of tracking performance, our method
    is able to substantially increase the fidelity of the renderings. This is also supported by the quantitative results PSNR. We reconstruct clean details compared to NICE-SLAM.}
    \label{fig:Room0_recon}
\end{figure}

\begin{figure}
    \centering
    \begin{subfigure}{0.5\textwidth}
        \centering
        \includegraphics[width=0.85\linewidth]{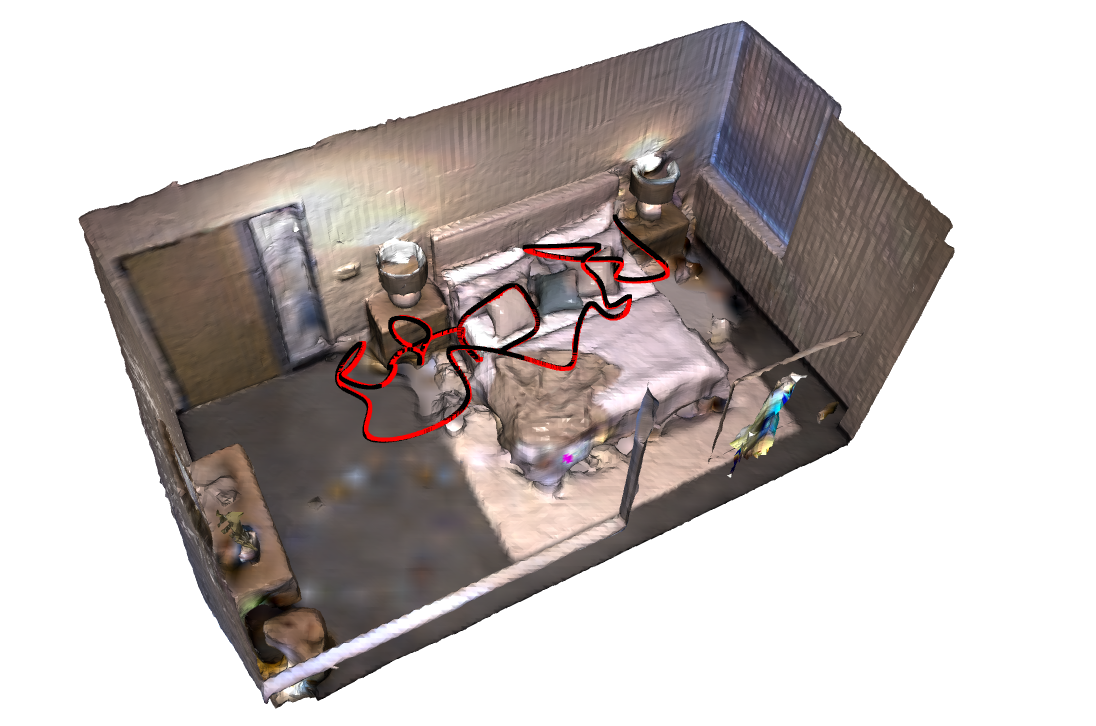}
        \caption{NICE-SLAM}
         \vspace{1em}
        \label{fig:render_nice-r1}
    \end{subfigure}
    \hfill
    \begin{subfigure}{0.5\textwidth}
        \centering
        \includegraphics[width=0.78\linewidth]{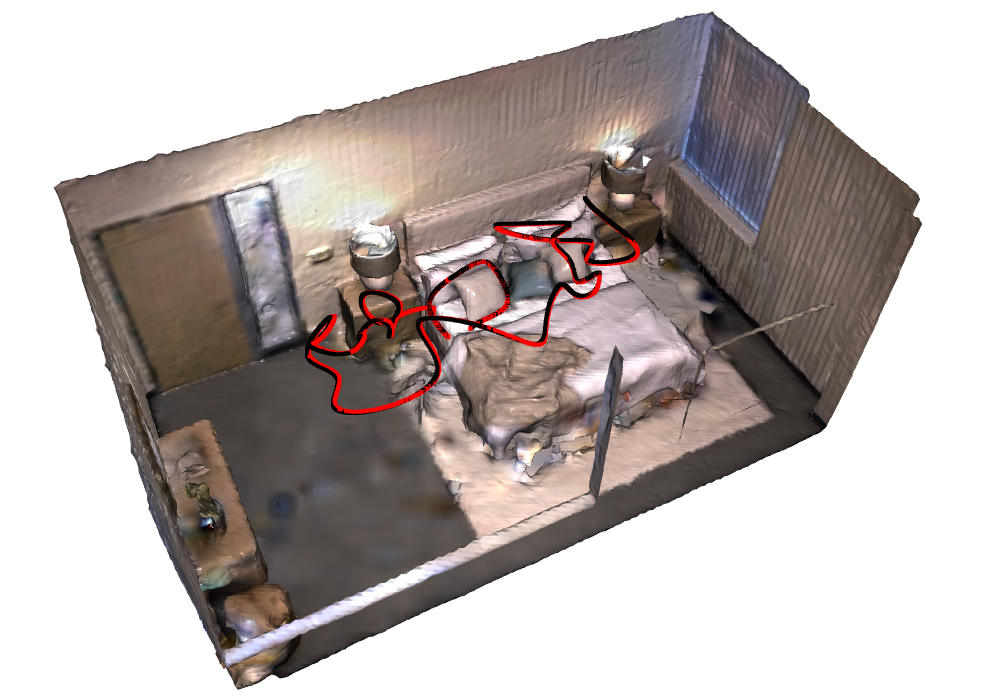}
        \caption{Ours}
        \vspace{1em}
        \label{fig:render_ours-r1}
    \end{subfigure}
    \begin{subfigure}{0.5\textwidth}
        \centering
        \includegraphics[width=0.9\linewidth]{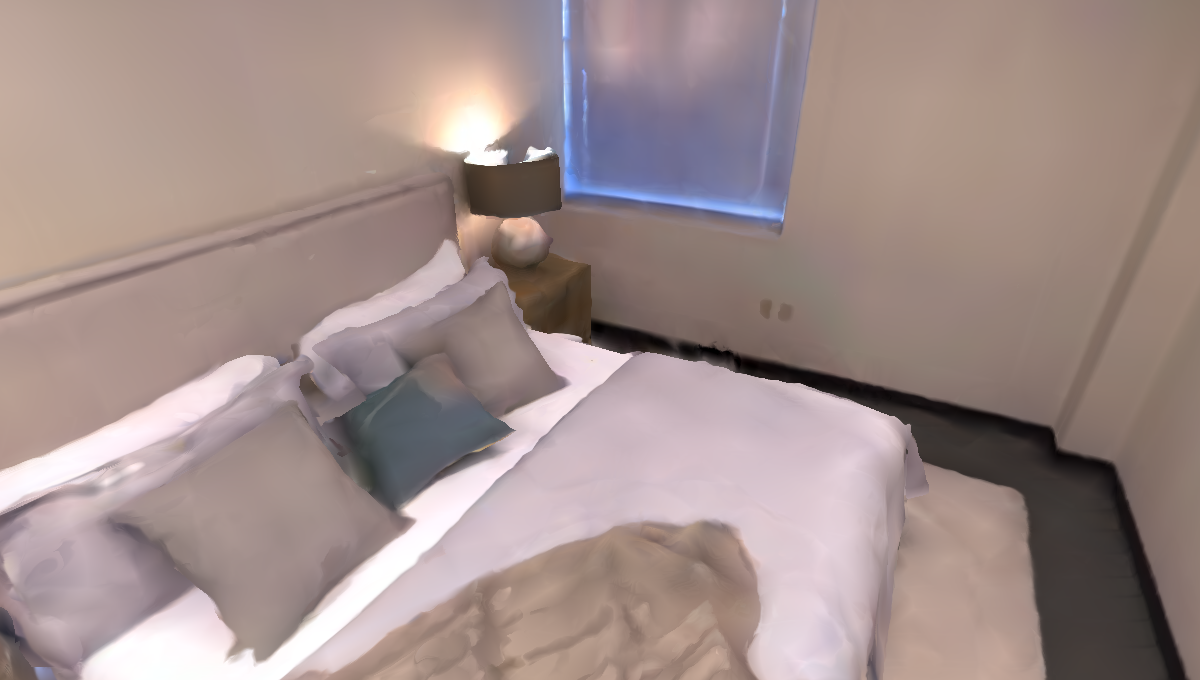}
        \caption{NICE-SLAM PSNR: 32.7}
        \vspace{1em}
        \label{fig:render_nice-r1-rgb}
    \end{subfigure}
    \begin{subfigure}{0.5\textwidth}
        \centering
        \includegraphics[width=0.9\linewidth]{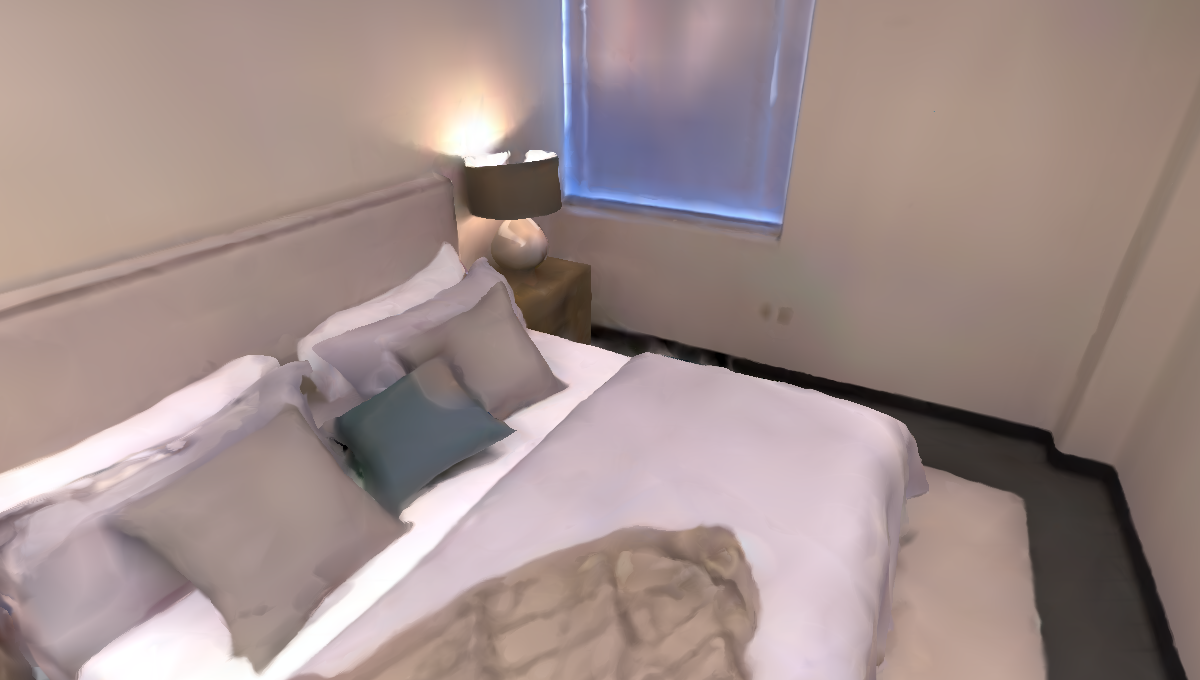}
        \caption{Ours PSNR: 33.3}
        \label{fig:render_our-r1-rgb}
    \end{subfigure}%
    
    \caption{\textbf{Reconstruction and Rendering of Replica Room1.} In this relatively easier scene, we perform slightly better than NICE-SLAM in rendering and reconstruction with less artifacts.}
    \label{fig:Room1_recon}
\end{figure}

\begin{figure}
    \centering
    \begin{subfigure}{0.5\textwidth}
        \centering
        \includegraphics[width=0.9\linewidth]{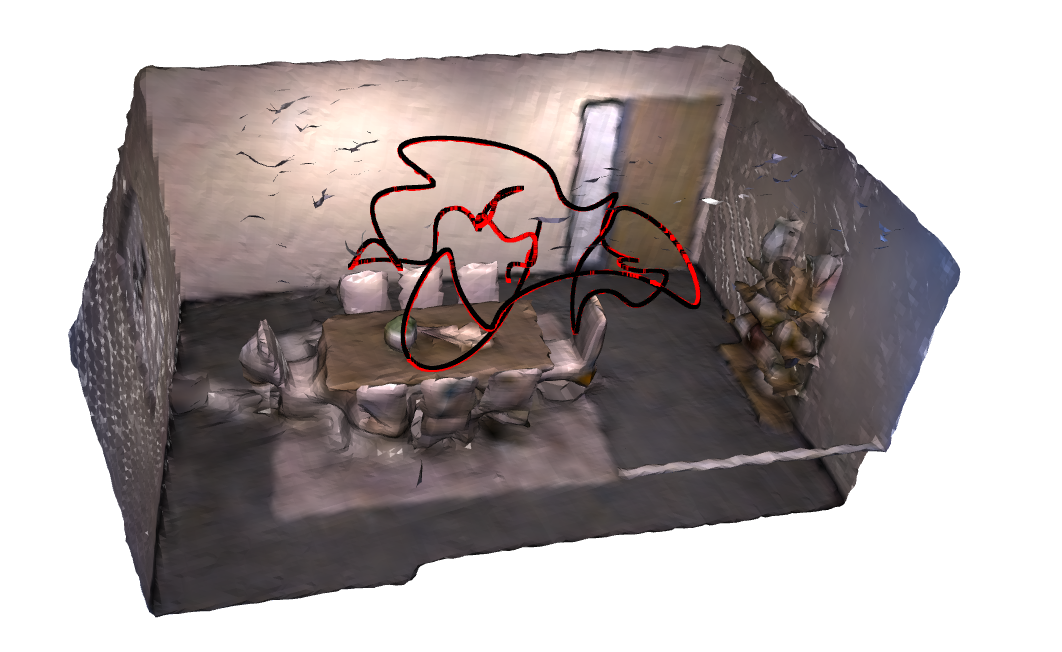}
        \caption{NICE-SLAM}
         \vspace{1em}
        \label{fig:render_nice-r2}
    \end{subfigure}
    \hfill
    \begin{subfigure}{0.5\textwidth}
        \centering
        \includegraphics[width=0.9\linewidth]{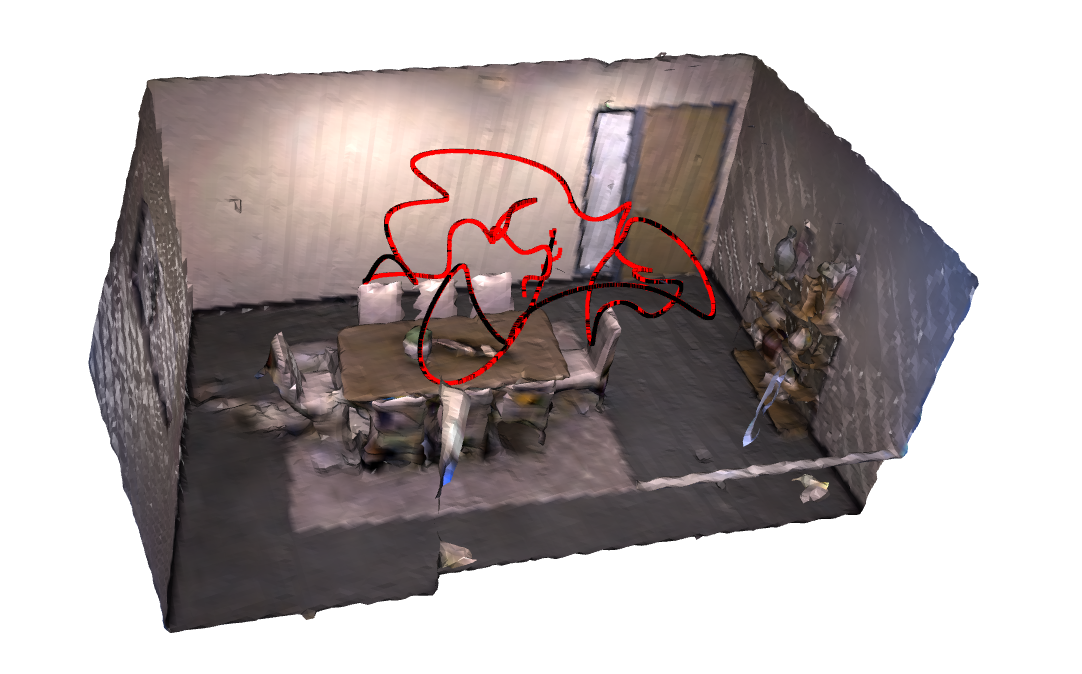}
        \caption{Ours}
        \vspace{1em}
        \label{fig:render_ours-r2}
    \end{subfigure}
    \begin{subfigure}{0.5\textwidth}
        \centering
        \includegraphics[width=0.9\linewidth]{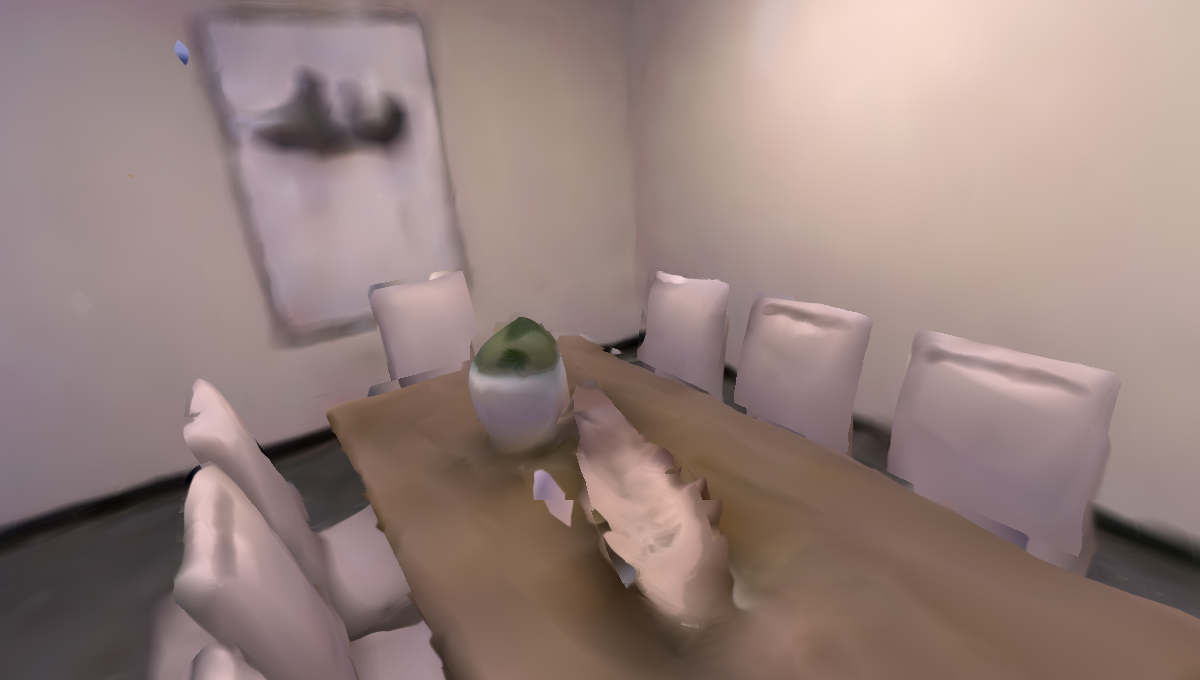}
        \caption{NICE-SLAM PSNR: 33.3}
        \vspace{1em}
        \label{fig:render_nice-r2-rgb}
    \end{subfigure}
    \begin{subfigure}{0.5\textwidth}
        \centering
        \includegraphics[width=0.9\linewidth]{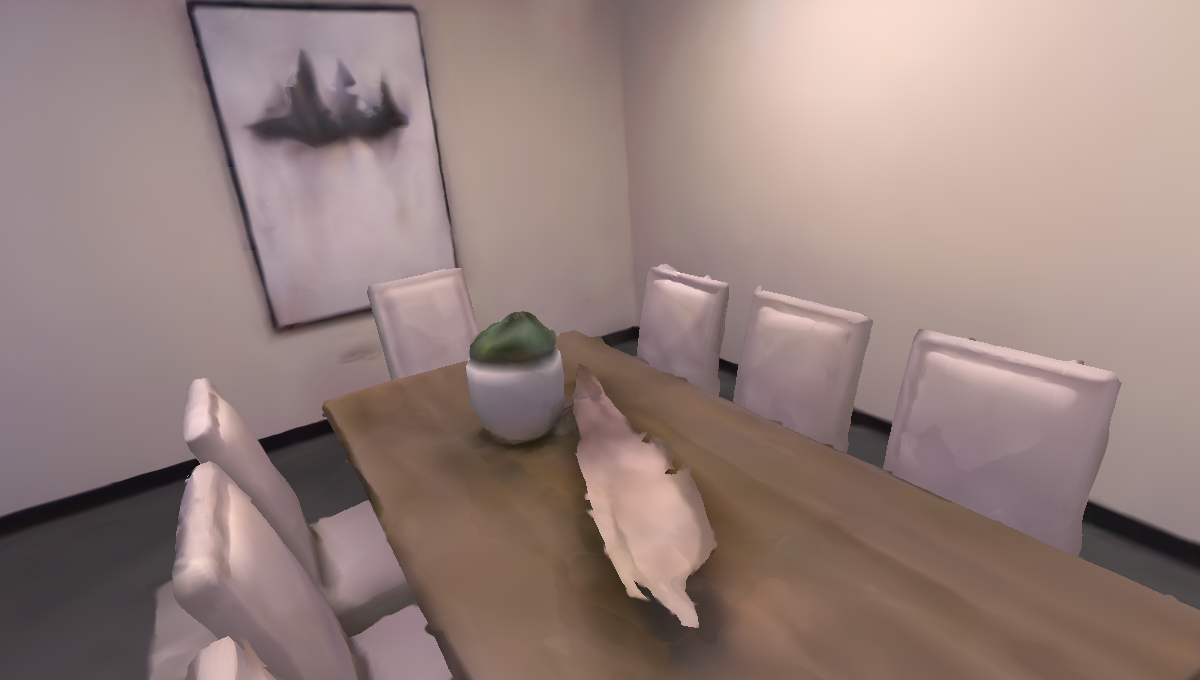}
        \caption{Ours PSNR: 36.8}
        \label{fig:render_our-r2-rgb}
    \end{subfigure}%
    
    \caption{\textbf{Reconstruction and Rendering of Replica Room2.} 
    While the reconstruction demonstrates that the NICE-SLAM trajectory is highly aligned with the ground truth, it adversely affects rendering performance, resulting in lower fidelity. In contrast, our method maintains high-fidelity rendering.}
    \label{fig:Room2_recon}
\end{figure}



\begin{figure}
    \centering
    
    \begin{subfigure}{0.5\textwidth}
        \centering
        \includegraphics[width=0.75\linewidth, trim={0  0.7cm  0  1.4cm},clip]{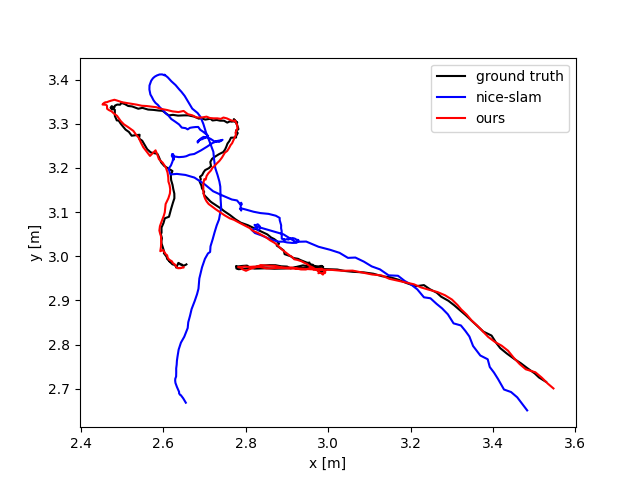}
        \caption{scan/0000}
        \label{fig:scan0000_traj}
    \end{subfigure}%
    \hfill
    \begin{subfigure}{0.5\textwidth}
        \centering
        \includegraphics[width=0.75\linewidth, trim={0  0.7cm  0  1.4cm},clip]{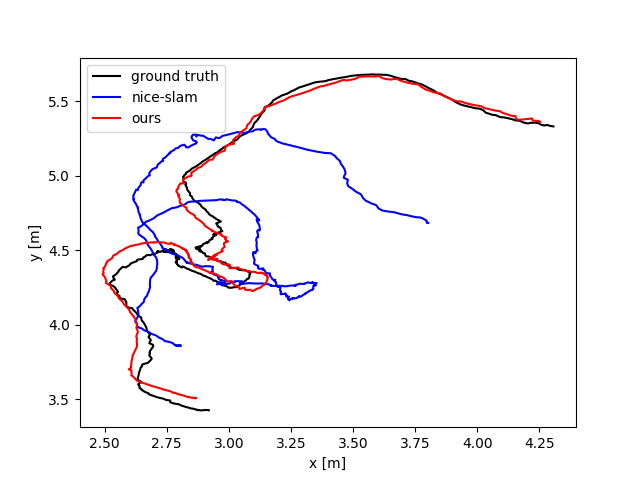}
        \caption{scan/0059}
        \label{fig:scan0059_traj}
    \end{subfigure}%
    \hfill
    \begin{subfigure}{0.5\textwidth}
        \centering
        \includegraphics[width=0.75\linewidth, trim={0  0.7cm  0  1.4cm},clip]{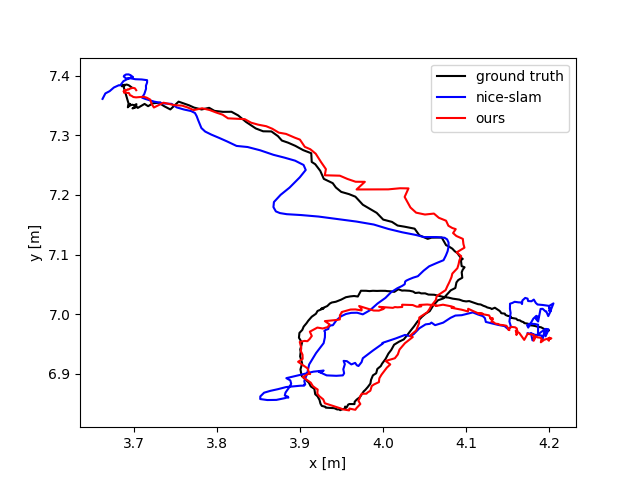}
        \caption{scan/0106}
        \label{fig:scan0106_traj}
    \end{subfigure}%
    \hfill
    \begin{subfigure}{0.5\textwidth}
        \centering
        \includegraphics[width=0.75\linewidth, trim={0  0.7cm  0  1.4cm},clip]{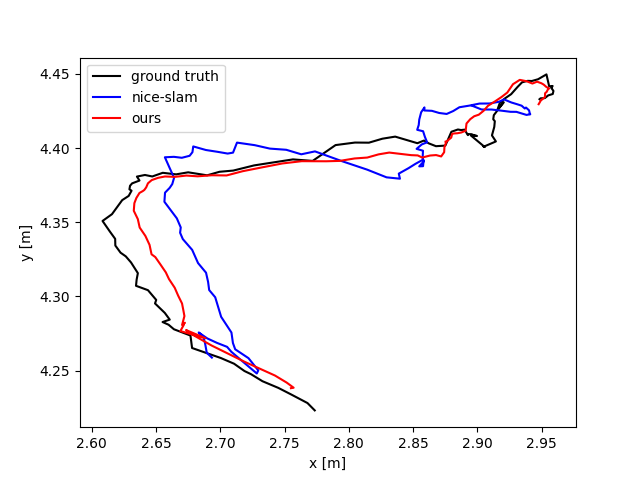}
        \caption{scan/0181}
        \label{fig:scan0181_traj}
    \end{subfigure}%
    \hfill
    \begin{subfigure}{0.5\textwidth}
        \centering
        \includegraphics[width=0.75\linewidth, trim={0  0.7cm  0  1.4cm},clip]{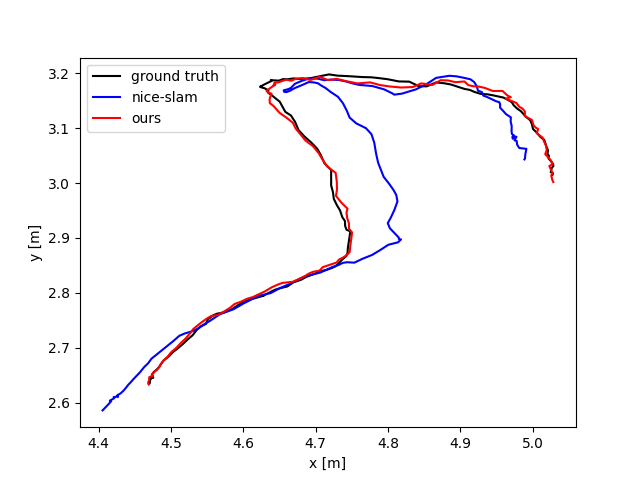}
        \caption{scan/0207}
        \label{fig:scan0207_traj}
    \end{subfigure}%
    
    \caption{\textbf{Qualitative results of tracking on ScanNet\cite{dai2017scannet}}.The initial trajectories diverge in the NICE-SLAM trajectory from the ground truth, while ours align with it.}
    \label{fig:scannet_traj_plot}
\end{figure}

\begin{figure*}
    \centering
    
    \begin{subfigure}{0.5\textwidth}
        \centering
        \includegraphics[width=0.9\linewidth, trim={0  0.7cm  0  1.4cm},clip]{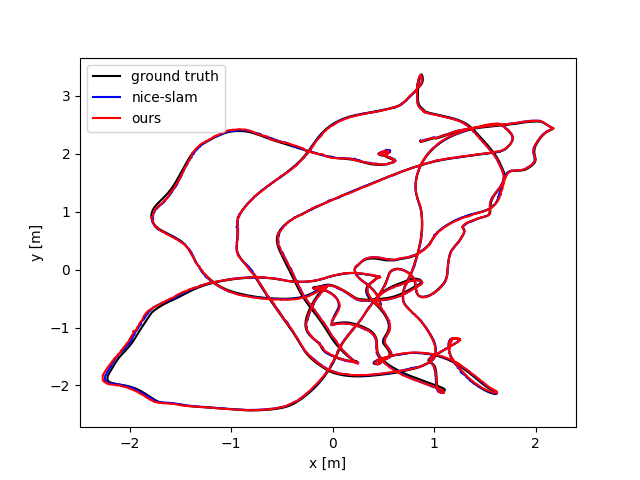}
        \caption{euroc/v101}
        \label{fig:v101_traj}
    \end{subfigure}%
    \hfill
    \begin{subfigure}{0.5\textwidth}
        \centering
        \includegraphics[width=0.9\linewidth, trim={0  0.7cm  0  1.4cm},clip]{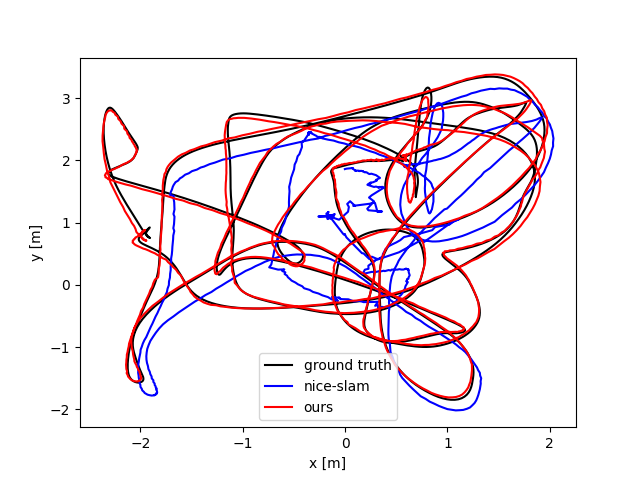}
        \caption{euroc/v102}
        \label{fig:v102_traj}
    \end{subfigure}%
    \hfill
    \begin{subfigure}{0.5\textwidth}
        \centering
        \includegraphics[width=0.9\linewidth, trim={0  0.7cm  0  1.4cm},clip]{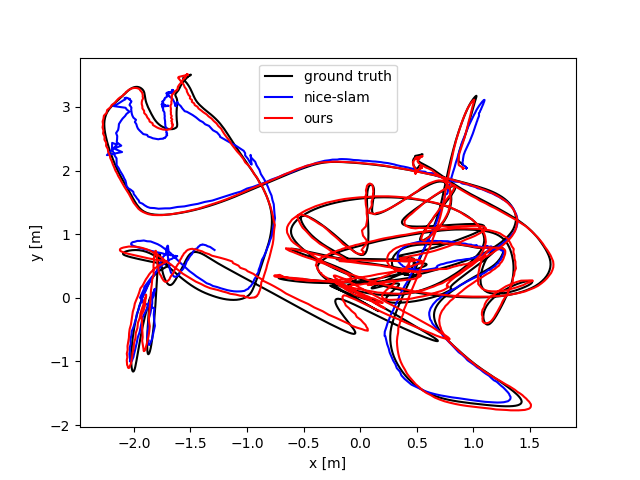}
        \caption{euroc/v103}
        \label{fig:v103_traj}
    \end{subfigure}%
    \hfill
    \begin{subfigure}{0.5\textwidth}
        \centering
        \includegraphics[width=0.9\linewidth, trim={0  0.7cm  0  1.4cm},clip]{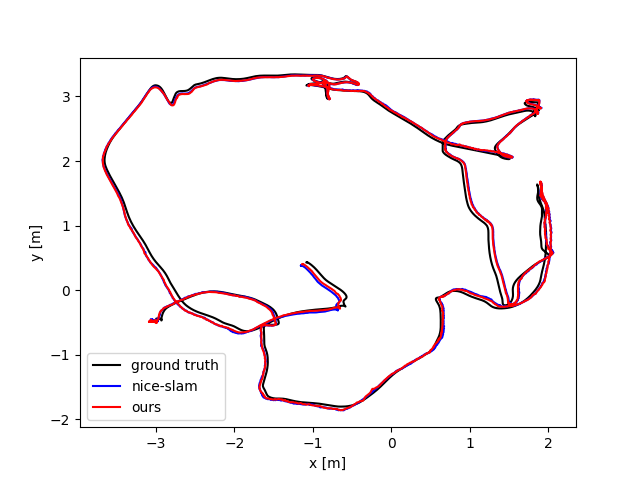}
        \caption{euroc/v201}
        \label{fig:v201_traj}
    \end{subfigure}%
    \hfill
    \begin{subfigure}{0.5\textwidth}
        \centering
        \includegraphics[width=0.9\linewidth, trim={0  0.7cm  0  1.4cm},clip]{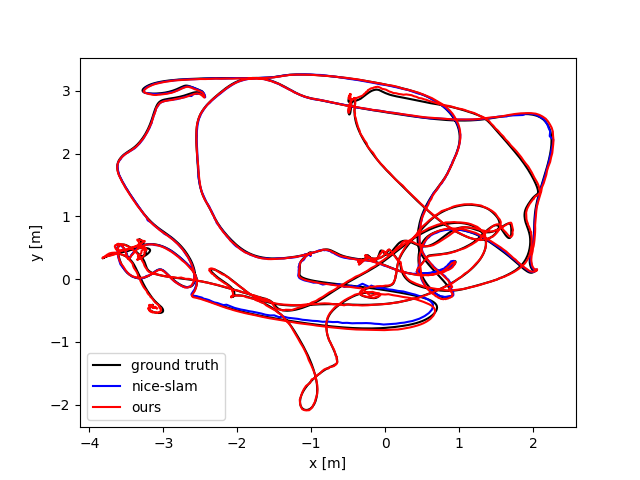}
        \caption{euroc/v202}
        \label{fig:v202_traj}
    \end{subfigure}%
    \hfill
    \begin{subfigure}{0.5\textwidth}
        \centering
        \includegraphics[width=0.9\linewidth, trim={0  0.7cm  0  1.4cm},clip]{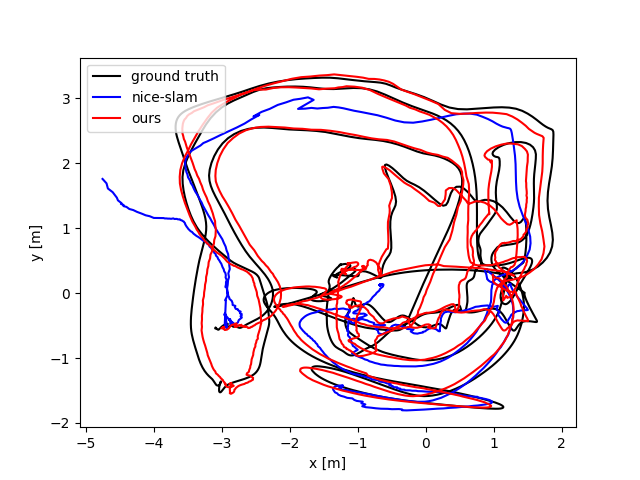}
        \caption{euroc/v203}
        \label{fig:v203_traj}
    \end{subfigure}%
    
    \caption{\textbf{Qualitative results of tracking on EUROC\cite{burri2016euroc}}. We compare the trajectories of our method to NICE-SLAM. Notably, NICE-SLAM encounters failures at v102, v202, and v203, so only part of trajectories are displayed. The results indicate that our method significantly aligns with the ground truth trajectory.}
    \label{fig:euroc_traj_plot}
\end{figure*}

\clearpage
{
    \small
    \bibliographystyle{ieeenat_fullname}
    \bibliography{main}
}


\end{document}